\documentclass[Afour,sageh,times]{sagej}

\usepackage{moreverb,url}
\usepackage{amsmath} 
\usepackage{amssymb}  
\usepackage{multirow}
\usepackage{array,booktabs}
\usepackage{diagbox}
\usepackage{balance}
\usepackage{mathtools}
\usepackage{subfigure}
\usepackage{verbatim}
\usepackage{adjustbox}
\usepackage{algorithm}
\usepackage{algpseudocode}
\usepackage{romannum}
\usepackage{amssymb}
\usepackage{pifont}
\usepackage{hyperref}
\usepackage{enumitem}
\hypersetup{
  colorlinks=true, 
  urlcolor=blue,   
  linkcolor=blue,  
  citecolor=magenta  
}
\usepackage[table]{xcolor}
\usepackage{rpm_SIunits}
\usepackage{rpm_acronyms}
\usepackage{rpm_math}
\usepackage{rpm_misc}
\usepackage{soul,color}
\usepackage{multirow}
\usepackage{soul,color}
\usepackage{lipsum}
\usepackage{tikz} 
\usepackage{MnSymbol}

\newcommand{\cmark}{\ding{51}}
\newcommand{\xmark}{\ding{55}}

\newcolumntype{?}[1]{!{\vrule width #1}}

\newcommand\BibTeX{{\rmfamily B\kern-.05em \textsc{i\kern-.025em b}\kern-.08em
T\kern-.1667em\lower.7ex\hbox{E}\kern-.125emX}}

\def\volumeyear{2024}

\usepackage{bbding,pifont}

\usepackage{xspace}
\newcommand{\eg}{e.g.\xspace}
\newcommand{\ie}{i.e.\xspace}

\begin{document}

\runninghead{Noh \emph{et al.}}

\title{GaRLILEO: Gravity-aligned Radar-Leg-Inertial Enhanced Odometry}

\author{Chiyun Noh\affilnum{1}\affilnum{$^\dagger$}, Sangwoo Jung\affilnum{1}\affilnum{$^\dagger$}, Hanjun Kim\affilnum{1}, Yafei Hu\affilnum{2}, Laura Herlant\affilnum{2}, and Ayoung Kim\affilnum{1}}

\affiliation{\affilnum{1} Dept. of Mechanical Engineering, SNU, Seoul, S. Korea \\
\affilnum{2} Robotics and AI Institute, Cambridge, USA\\
\affilnum{$^\dagger$}The authors contributed equally to this paper}

\corrauth{Ayoung Kim, Dept. of Mechanical Engineering, SNU, Seoul, S. Korea}
\email{ayoungk@snu.ac.kr}

\begin{abstract}

Deployment of legged robots for navigating challenging terrains (\eg, stairs, slopes, and unstructured environments) has gained increasing preference over wheel-based platforms. In such scenarios, accurate odometry estimation is a preliminary requirement for stable locomotion, localization, and mapping. Traditional proprioceptive approaches, which rely on leg kinematics sensor modalities and inertial sensing, suffer from irrepressible vertical drift caused by frequent contact impacts, foot slippage, and vibrations, particularly affected by inaccurate roll and pitch estimation. Existing methods incorporate exteroceptive sensors such as \ac{LiDAR} or cameras. Further enhancement has been introduced by leveraging gravity vector estimation to add additional observations on roll and pitch, thereby increasing the accuracy of vertical pose estimation. However, these approaches tend to degrade in feature-sparse or repetitive scenes and are prone to errors from double-integrated IMU acceleration. To address these challenges, we propose \textbf{GaRLILEO}, a novel gravity-aligned continuous-time radar-leg-inertial odometry framework. GaRLILEO decouples velocity from the IMU by building a continuous-time ego-velocity spline from SoC radar Doppler and leg kinematics information, enabling seamless sensor fusion which mitigates odometry distortion. In addition, GaRLILEO can reliably capture accurate gravity vectors leveraging a novel soft $\mathcal{S}^2$-constrained gravity factor, improving vertical pose accuracy without relying on \ac{LiDAR} or cameras. Evaluated on a self-collected real-world dataset with diverse indoor-outdoor trajectories, GaRLILEO demonstrates state-of-the-art accuracy, particularly in vertical odometry estimation on stairs and slopes. We open-source both our dataset and algorithm to foster further research in legged robot odometry and SLAM. 
\texttt{\url{https://garlileo.github.io/GaRLILEO/}}

\end{abstract}


\maketitle

\section{Introduction}

\begin{figure*}[!t]
    \centering
    \includegraphics[width=\textwidth]{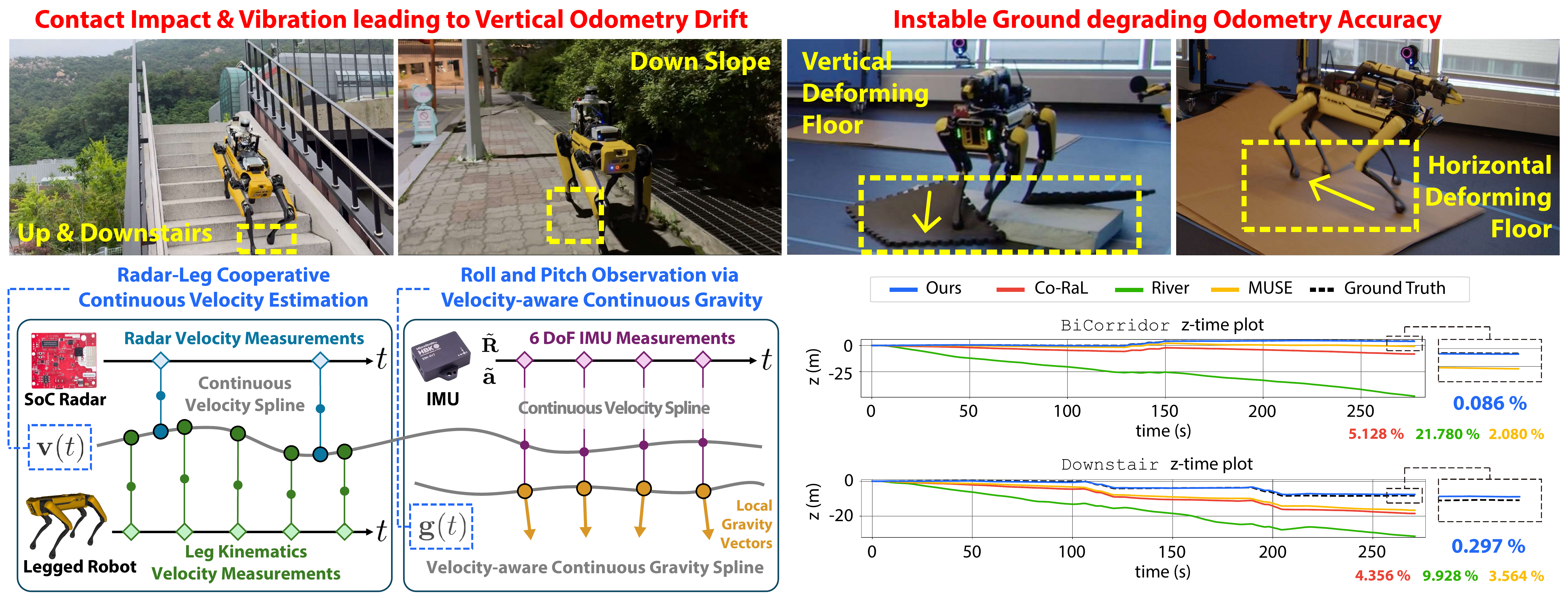}
    \caption{Overall preview of GaRLILEO. The four subfigures in the upper row present the problematic situations that quadrupedal robots may encounter while performing real-world tasks, while the yellow letters specify the situations and the red words explain the substantial issues generated from them. Two boxes in the left part of the lower row summarize the major contribution and method of the GaRLILEO, which significantly reduces odometry error, especially in the vertical direction. Two graphs in the right part of the lower row present the short experimental results, showing the accuracy of GaRLILEO in multiple sequences that include loops, sharp turns, and staircases, where most baselines fail to maintain accuracy in odometry estimation. }
    \label{fig:figure_1}
\end{figure*}

Legged robots have increasingly attracted attention for their robust mobility and adaptability in harsh environments, where traditional wheeled \ac{UGV} systems often struggle. Their ability to traverse stairs, steep slopes, uneven surfaces, and unstructured terrain makes them well-suited for real-world deployment in search-and-rescue, inspection, and exploration tasks \citep{tranzatto2022teamcerberus}. To fully leverage these capabilities in practice, it is essential to ensure accurate odometry estimation, which underpins stable locomotion, localization, and mapping in such challenging scenarios.

A common practice for robust state estimation in legged robots is leveraging proprioceptive sensing, which directly captures internal kinematics of the robot through contact measurements, joint encoders, and inertial sensing \citep{hartley2018legged, hartley2020contact, kim2021legged, yang2023multi, lin2023proprioceptive}. These proprioceptive approaches capitalize on the direct sensing of robot dynamics as they do not depend on external observations, making them inherently robust to visual or geometrical degradation. Nonetheless, frequent contact impacts, foot slippage, and intense vibrations significantly impair the accuracy of proprioceptive odometry, particularly in the vertical direction. 

A natural progression to address this vertical drift is to leverage exteroceptive sensors, such as cameras and \ac{LiDAR}s. Most \ac{LiDAR}-based methods apply ground segmentation and ground constraints to suppress vertical errors \citep{shan2018lego, wei2021ground, seo2022pago, wang2024robust}. However, these strategies are mainly effective in wide, flat, and structured environments, which differ significantly from the cluttered and irregular terrains targeted by legged robots. Camera-based approaches also frequently rely on planar segmentation and Manhattan world assumptions to constrain vertical error \citep{li2020structure, shu2021visual}, yet these assumptions break down in complex natural environments. Moreover, recent studies further indicate that simply introducing planar landmarks may not guarantee direct improvement in managing odometry drift \citep{arndt2023planar}, underscoring the limitations of exteroceptive registration in realistic legged robot scenarios.

Using an \ac{IMU} with cameras and \ac{LiDAR}s can significantly enhance the state estimation. Among its many advantages, gravity estimation provides additional constraint for roll and pitch, enhancing state estimation, a concept that was first introduced in VINS-Mono \citep{qin2018vins}.
While some studies \citep{kubelka2022gravity, D-LIOM, nemiroff2023joint, wildcat, agha2021nebula} report notable improvements, others indicate only marginal performance gains \citep{burnett2025tro}. 
This may be because most existing methods rely on fusing \ac{IMU} acceleration with pose estimates. Such fusion requires double integration, and the gravity estimate inherently depends on the quality of the pose estimation. Consequently, feeding this pose-dependent gravity estimate back into the state may provide only limited improvement.

A promising alternative for overcoming these limitations is the use of radar. By providing direct velocity measurements, radar can fuse its ego-velocity with \ac{IMU} acceleration via single integration for local gravity estimation~\citep{noh2025garlio}. 
While effective, this integration can be limited in legged robot operations, where velocity changes sporadically due to frequent contacts and impacts.
Another more intuitive way to integrate radar into the legged robot system is by utilizing the instantaneous ego-velocity derived from the leg kinematics as in Co-RaL~\citep{jung2024co}.
This approach takes advantage of radar’s ability to operate reliably under environmental degradation while also utilizing the high-frequency, proprioceptive information provided by leg kinematics sensors. 
However, despite these improvements, residual vertical drift persists due to inaccurate roll and pitch estimation. This limitation primarily stems from the exclusive reliance on \ac{IMU} gyroscopes for orientation.

To bridge the critical gap between proprioceptive odometry and robust gravity estimation, we extend our previous works GaRLIO~\citep{noh2025garlio} and Co-RaL~\citep{jung2024co} to complete \textbf{GaRLILEO}, a \underline{\textbf{G}}ravity-\underline{\textbf{a}}ligned \underline{\textbf{R}}adar-\underline{\textbf{L}}eg-\underline{\textbf{I}}nertia\underline{\textbf{L}} \underline{\textbf{E}}nhanced \underline{\textbf{O}}dometry framework. First, we decouple the \ac{IMU} from velocity estimation, relying solely on radar and leg kinematics to compute the ego-velocity. This decoupled scheme is particularly advantageous for legged robot systems where noisy accelerations are often recorded during ground contact. Unlike previous radar-based approaches that represent velocity as a discrete state, our method formulates velocity as a continuous-time variable and enables seamless sensor fusion across modalities with different frame rates. We note that this is uniquely feasible in a leg–radar system, as radar directly measures velocity and leg kinematics can directly compute it, eliminating the need for pose-level measurements that would require double integration.

Beyond seamless fusion between radar and leg kinematics, the fundamental challenge of pose estimation of legged robots lies in the contact impacts and vibrations that persistently corrupt gravity estimation. To suppress these sudden, undesired acceleration measurements from the \ac{IMU}, we employ splines as an inherent filter, bounding the gravity vector within a smoothed, continuous vector space. Furthermore, to naturally constrain the magnitude of the gravity vector during optimization, we introduce $\mathcal{S}^2$ gravity factor. Rather than conventionally restricting the vector to a unit sphere, this factor actively anchors its magnitude to 9.81 while optimizing its direction. This design enables precise gravity estimation, effectively constraining the roll and pitch of a legged robot, even under harsh conditions such as contact impacts, instantaneous slips, and intense vibrations.
Our main contributions are as follows:

\begin{itemize}
    \item The proposed continuous-time proprioceptive state estimation framework effectively overcomes the asynchronous, high-impact nature of radar-Leg-\ac{IMU} systems. By formulating ego-centric velocity through splines and decoupling it from noisy \ac{IMU} accelerations, abrupt motion prevalent in legged locomotions is better handled. This continuous formulation inherently filters out instantaneous leg slips and radar noise, providing a stable, distortion-free foundation for downstream gravity and pose estimation without relying on visual or \ac{LiDAR} features.

    \item Our velocity-aware gravity estimation directly attacks the pervasive issue of vertical drift in legged odometry. By integrating a soft $\mathcal{S}^2$-constrained gravity factor within our continuous-time framework, we continuously anchor the gravity vector's magnitude while optimizing its direction. This ensures that even under the intense vibrations caused by foot contacts, the system maintains a precise lock on the local gravity vector, reducing roll and pitch degradation.
    
    \item We present a comprehensive, real-world radar-Leg-Inertial dataset which features aggressive elevation changes across stairs, slopes, and slippery indoor/outdoor terrain. Validated against high-fidelity \ac{TLS} and motion capture ground truth trajectories, GaRLILEO demonstrates \ac{SOTA} performance, proving exceptionally resilient to z-axis variations where traditional methods fail. To foster further research, both the dataset and framework source code are released to the community. 
\end{itemize}

\section{Related works}

In this section, we review prior work most relevant to our approach. We begin with \ac{SoC} radar odometry methods that leverage ego-velocity estimation.
We then turn to recent developments in odometry for legged robots, covering both exteroceptive and proprioceptive paradigms. Building on this, we highlight advances in gravity estimation within state estimation frameworks. Lastly, we review continuous-time odometry methods that are particularly relevant to our work.

\subsection{SoC Radar Odometry}

Radars have emerged as a critical sensing modality in robotics, offering robust perception in adverse environments \citep{harlow2024new}.
Robotics applications commonly utilize two types of radars: spinning radars and phased-array \ac{SoC} radars \citep{kim2025sherloc}. This paper mainly focuses on \ac{SoC} radars, which employ \ac{FMCW} technology to generate 4D point clouds capturing range, azimuth, elevation, and Doppler radial velocity. While structurally similar to \ac{LiDAR} data, \ac{SoC} radar measurements are typically sparser and exhibit lower point precision, posing unique challenges for odometry estimation.

A primary approach to leveraging \ac{SoC} radar for odometry involves directly estimating ego-velocity using point-wise radial velocity measurements. Early work by \citet{kellner2013instantaneous} introduced instantaneous ego-motion estimation using \ac{RANSAC} for outlier rejection and least-squares optimization for a single \ac{SoC} radar.
Building on this foundation, subsequent studies can be categorized into two main directions: those integrating spatial information into ego-velocity estimation and those fusing it with inertial measurements from \ac{IMU}s.

Several approaches have explored the integration of spatial information from \ac{SoC} radar point clouds. For example, \citet{michalczyk2022tightly} utilized stochastic cloning to associate 3D points between consecutive point clouds, thereby enhancing odometry accuracy. Similarly, 4D-iRIOM~\citep{zhuang20234d} combined \ac{SoC} radar velocity with scan-matching techniques, improving robustness by aligning scan-to-submap registration with Doppler-driven ego-velocity estimates. Recently, \citet{huang2024less} introduced \ac{RCS}-based filtering to refine point correspondences. Despite these advancements, the low precision and sparsity of \ac{SoC} radar point clouds often make registration vulnerable, leading to unbounded drift or failure in odometry estimation under challenging conditions.

In contrast to registration-based methods, other studies have integrated ego-velocity estimation with inertial measurements from \ac{IMU}s. For instance, \citet{doer2020ekf} employed an \ac{EKF}, while \citet{park20213d} utilized factor-graph optimization to achieve 6-\ac{DoF} odometry in visually degraded environments. Specifically, \citet{park20213d} leveraged two perpendicular \ac{SoC} radars for 3D ego-velocity estimation and introduced a radar velocity factor for pose-graph \ac{SLAM} that incorporates \ac{IMU} rotation data. More recently, DRIO~\citep{chen2023drio} estimated ego-velocity and ground points, achieving robust 2D odometry. Also, Co-RaL~\citep{jung2024co} proposed a 4-\ac{DoF} optimization strategy to mitigate vertical drift caused by the limited elevation resolution of \ac{SoC} radars. Similarly, DeRO~\citep{do2024dero} employed dead-reckoning with \ac{SoC} radar ego-velocity and gyroscope data, further refined through an \ac{IEKF} with tilt angle estimation based on accelerometers. Recently, River~\citep{chen2024river} tightly fused \ac{SoC} radar ego-velocity and \ac{IMU} measurements using a B-spline-based framework, presenting precise velocity estimation. 

While prior works have significantly advanced \ac{SoC} radar odometry by leveraging ego-velocity and spatial information, they face persistent challenges in mitigating vertical drift and ensuring robust performance under the dynamic conditions of legged robot systems. In such scenarios, low-precision vertical velocity measurements from \ac{SoC} radars and contact-induced noise in \ac{IMU} acceleration data often contribute to significant vertical drift in odometry estimation. Most existing methods either suffer from unbounded drift due to insufficient vertical constraints or rely on point cloud registration schemes that are highly sensitive to radar noise and sparsity. To overcome these limitations, our proposed method, GaRLILEO, seamlessly integrates radar-derived ego-velocity with high-rate proprioceptive measurements within a continuous-time framework, significantly reducing odometry distortion. Furthermore, our robust proprioceptive-based local gravity estimation scheme effectively mitigates vertical drift, enabling stable and accurate odometry even in dynamic and complex environments.

\subsection{Leg kinematics Odometry}

Compared with traditional wheeled \ac{UGV}, a legged robot provides two unique sensor modalities: (1) contact sensors, which indicate the contact state of each foot, and (2) joint encoders, which measure the orientation of each joint. Using forward kinematics, the relative position of each foot with respect to the robot base can be computed~\citep{roston1992dead}, completing leg-based odometry. The leg odometry can be divided into two categories: methods that leverage exteroceptive sensors such as \ac{LiDAR} or cameras, and those that focus primarily on fusing proprioceptive sensors alone.

\subsubsection{Exteroceptive Approach}

A wide range of odometry frameworks for legged robots have harnessed exteroceptive sensors—such as cameras and \ac{LiDAR}—in combination with leg kinematics, to capitalize on feature-rich geometric information. \citet{fallon2014drift} introduced a pose estimator that fuses inertial and leg kinematics measurements with localization derived from \ac{LiDAR} and pre-built maps. Similarly, \citet{nobili2017heterogeneous} fused inertial, leg kinematics, camera, and \ac{LiDAR} measurements based on the \ac{EKF}, demonstrating robust state estimation across multiple walking gaits. This line of work was further extended by Pronto~\citep{camurri2020pronto}, which focused on managing time-delayed signals from the vision sensors and fusing leg kinematics velocity from each leg by using a weighted average. VILENS~\citep{wisth2022vilens} proposed a tightly-coupled, graph-optimization-based approach by fusing camera, \ac{LiDAR}, leg kinematics sensors, and \ac{IMU}. This work enabled temporal proprioceptive odometry based on the preintegration factor when exteroceptive sensors failed under extreme conditions. STEP~\citep{kim2022step} improved the stability of stereo camera-based odometry by incorporating leg kinematics, achieving robust performance in dynamic scenes. Similarly, Leg-KILO~\citep{ou2024leg} demonstrated stable \ac{LiDAR}-based odometry by leveraging leg kinematics during dynamic movements of legged robots. Cerberus~\citep{yang2023cerberus} introduced online calibration of kinematics parameters and contact outlier rejection to reduce drift in camera-\ac{IMU}-leg kinematics odometry. To address dynamic locomotion behaviors such as jumping and trotting at varying speeds, \citet{dhedin2023visual} fused leg odometry from Pronto~\citep{camurri2020pronto} with \ac{IMU} frequency speed camera-\ac{IMU} odometry. Recently, MUSE~\citep{nistico2025muse} integrated foot-slip detection with camera-\ac{LiDAR} odometry to further enhance robustness. Holistic Fusion~\citep{nubert2025holistic} provided a unified framework in which leg odometry, especially using the contact frame as a landmark feature, could be seamlessly fused with exteroceptive sensors. 

While these exteroceptive sensor-based approaches have shown strong odometry performance—especially when combined with leg kinematics information—they still encounter challenges in environments where perceptual features are sparse, ambiguous, or repetitive. In particular, visual and \ac{LiDAR} odometry can degrade rapidly in poorly illuminated areas, featureless corridors, reflective surfaces, or in the presence of dense smoke and dust, limiting their reliability in many real-world legged robot applications. 

\subsubsection{Proprioceptive Approach}

In contrast to exteroceptive approaches, proprioceptive-based odometry estimates the robot pose without dependence on external features. \citet{bloesch2012state} introduced an \ac{EKF}-based framework that fused leg kinematics measurements and \ac{IMU} data for state estimation, which was later extended to a \ac{UKF} backbone \citep{bloesch2013state}. 

Based on the contact theorem that a contact frame is fixed while a contact sensor is on, \citet{hartley2018legged} proposed a forward kinematics factor and a preintegrated contact factor. This was later generalized to a hybrid contact factor that dynamically switches the contact frame based on the assumption that at least one foot is in contact with the ground \citep{hartley2018hybrid}. The contact kinematics theory was incorporated into an invariant \ac{EKF} framework, yielding a Lie group-based estimator that achieved globally consistent state estimation using IMU, kinematics, and contact measurements~\citep{hartley2020contact}. \citet{fink2020proprioceptive} additionally fused force and torque sensor data to develop a low-level state estimator, calculating both kinematics and dynamics of the robot. Still, the possible foot slip, even with the contact sensor in place, remains a challenge. 

To address foot slip on the contact frame, approaches from various directions have been introduced recently. TSIF~\citep{bloesch2017two} presented a recursive estimation framework minimizing the residual between two consecutive states and demonstrated improved resilience to measurement outliers. \citet{kim2021legged} adopted a fixed-lag smoother state estimation on the SO(3) manifold, paired with slip rejection strategies to mitigate kinematics model failures. DRIFT~\citep{lin2023proprioceptive} combined contact estimation and gyro filtering within an invariant \ac{EKF}, enabling robust odometry on low-cost legged robots. 
\citet{yang2023multi} utilized multiple \ac{IMU}s on each foot to explicitly detect contact and foot slip, overcoming the zero-velocity contact frame assumption prevalent in contact sensor-based proprioceptive odometry studies.

Although proprioceptive odometry is robust to environmental conditions and feature degradation, it suffers from drift over time, particularly in the vertical direction, due to the drastic vibration from contact impact and the lack of an absolute orientation reference. To address this, Co-RaL~\citep{jung2024co} introduced an integration of \ac{SoC} radar-derived ego-velocity with leg kinematics velocity incorporating rolling contact awareness. However, despite these integrated measurements, vertical drift remains due to inaccurate roll and pitch estimation. This stems from not accounting for \ac{IMU} acceleration, which contains critical information, such as the gravity vector, essential for accurate odometry.

GaRLILEO improves upon Co-RaL by integrating \ac{SoC} radar, leg kinematics, and IMU data using a continuous-time B-spline approach and introducing robust continuous velocity-aware gravity estimation.
This enables substantially improved robustness of odometry, especially in the vertical direction, compared to Co-RaL and other purely proprioceptive methods. 
\subsection{Local Gravity Estimation}

Gravity, with its constant magnitude and direction, provides a physically grounded reference for reliable roll and pitch estimation in odometry and \ac{SLAM}. Accurate roll and pitch estimation is crucial for mitigating vertical drift, which is stimulated by erroneous leakage of horizontal movement onto the vertical axis. Accordingly, accurate local gravity estimation has emerged as a critical constraint for suppressing accumulated vertical drift errors in state estimation frameworks.

Early gravity estimation in \ac{LiDAR} odometry typically inferred the local gravity vector from \ac{IMU} acceleration or employed probabilistic filtering over time. Nebula~\citep{agha2021nebula} introduced a gravity factor based on \ac{IMU} acceleration during stationary intervals, constraining roll and pitch in the state estimation. D-LIOM~\citep{D-LIOM} and Wildcat~\citep{wildcat} formulated gravity alignment as an optimization constraint using \ac{IMU} data and exteroceptive-based odometry. \citet{nemiroff2023joint} further extended this by jointly optimizing accelerometer intrinsics and the gravity vector. While these velocity-ignorant models presented the potential for local gravity estimation to mitigate odometry vertical drift, they fundamentally relied on correlating pose changes from exteroceptive odometry with IMU acceleration, which amplified errors due to IMU bias and noise through the process of double integration. This limitation hinders robustness in dynamic environments or those prone to slippage. Furthermore, \citet{burnett2025tro} reports that these methods offer only marginal performance gains; this is likely due to the inherently pose-dependent nature of velocity-ignorant gravity estimation, which limits the benefit of feeding estimation back into the state.

To overcome these issues, GaRLIO~\citep{noh2025garlio} explicitly incorporated direct velocity measurements from radar Doppler data for local gravity estimation. By fusing radar-derived ego-velocity with LiDAR odometry, GaRLIO constructed a velocity-aware gravity constraint that significantly enhanced the accuracy of gravity estimation. However, GaRLIO still depended on an initial gravity guess from \ac{LiDAR} scan registration, making it susceptible in feature-degraded environments or unadaptable to a system without \ac{LiDAR}. Additionally, mismatches in time intervals between the \ac{IMU} preintegration and radar ego velocity updates may introduce estimation inconsistencies due to the lack of continuity in the discrete-time sensor-fusion approach.

In contrast, our GaRLILEO framework integrates a continuous velocity-aware local gravity estimation approach, enabling precise and robust estimation of the gravity vector even in visually degraded or featureless environments. Furthermore, by continuously fusing \ac{SoC} radar velocity and leg kinematics measurements via a time-continuous B-spline optimization framework, GaRLILEO prevents the inconsistency that may arise from discrete sensor fusion across modalities with differing frame rates.

\subsection{Continuous-time State Estimation}

Sensor fusion is essential in robotics to leverage the complementary strengths of heterogeneous sensors. Traditional discrete-time frameworks synchronize each sensor by associating with the nearest available timestamp; however, this introduces motion distortion and overlooks latent state information between sampled states \citep{talbot2024continuous}.

To address this limitation, recent works have adopted B-spline-based continuous-time trajectory representations, enabling smooth and differentiable state estimation that supports querying poses, velocities, and accelerations at arbitrary time instances. \citet{furgale2012continuous, furgale2015continuous} formulated batch estimation of robot trajectories in SE(3) using temporal B-splines, establishing the groundwork for continuous-time sensor fusion. 

\citet{ovren2019trajectory, hug2020hyperslam, lang2022ctrl, hug2022continuous} extended these ideas for vision-inertial fusion, achieving improved robustness to motion distortion and asynchronous measurements. 
Similarly, \citet{droeschel2018efficient} leveraged B-splines for continuous-time \ac{SLAM} with \ac{LiDAR}-inertial system. 
Later, recent works like \citep{lv2023continuous, lang2023coco, lang2024gaussian} addressed multi-modal \ac{SLAM} and odometry between \ac{LiDAR}, camera, and \ac{IMU}. 
\citet{jung2023asynchronous} further addressed asynchronous fusion for multi-\ac{LiDAR}-\ac{IMU} odometry using B-spline approach for continuous-time formulations. For \ac{SoC} radar-\ac{IMU} systems, River~\citep{chen2024river} introduced a B-spline-based radar-inertial velocity estimator that could operate robustly under perception-degraded conditions.

Inspired by these recent advances, GaRLILEO presents the first integration of leg kinematics with \ac{SoC} radar and \ac{IMU} in a continuous-time B-spline framework. By directly incorporating leg kinematics velocities and radar measurements at their respective timestamps without preintegration, GaRLILEO provides more accurate and robust state estimation. Additionally, this continuous-time framework even enhances the gravity estimation accuracy, reducing the vertical drift, especially in challenging legged robot scenarios where slip and dynamic contact events frequently occur. 
\section{Preliminary}
In this section, we summarize the background for the following sections. 

\subsection{Radar Radial Velocity and Ego Velocity}
Leveraging the \ac{FMCW} technique, phased array radar can provide not only the 3D position of the points that they acquire, but also the radial velocity of the point in the sensor coordinate. 
If a single radar point is generated from a stationary object, the radial velocity of the point can be represented with the ego-centric velocity of the sensor. Let the ego-centric velocity of the sensor is $\mathbf{v}$, the radial velocity of the point is $v_j$, and the 3D position of the point is represented with vector $\mathbf{p}_j$; then their relationship is like follows:
\begin{equation}
\begin{aligned}
\label{eq:3Dsin}
    v_j = - \frac{\mathbf{p}_j}{\left\|\mathbf{p}_j\right\|}\mathbf{v}.
\end{aligned}
\end{equation}

Using this radial velocity, we can calculate ego-velocity using \ac{RANSAC} and least-square optimization processes \citep{kellner2013instantaneous}. 
For a simple illustration, let's assume that the radar acquires only 2D data. Then, equation \eqref{eq:3Dsin} can be expressed as follows:
\begin{equation}
\begin{aligned}
\label{eq:2Dsin}
    \theta_j &= \tan^{-1} \frac{p_{j,y}}{p_{j, x}} \\ 
    v_j &= - 
    \begin{bmatrix}
        \cos(\theta_j) ~~~~~ \sin(\theta_j)
    \end{bmatrix}
    \mathbf{v},
\end{aligned}
\end{equation}
where $\mathbf{v} = {[v_x ~~~ v_y]}^\top$. Expanding the \eqref{eq:2Dsin} to all $N$ radar points, the result equation is like follows:
\begin{equation}
\begin{aligned}
\label{eq:2D_full}
    \begin{bmatrix}
        v_1
        \\
        \vdots
        \\
        v_N
    \end{bmatrix}
    =
    \begin{bmatrix}
        \cos(\theta_1) ~~~~~ \sin(\theta_1)
        \\
        \vdots
        \\
        \cos(\theta_N) ~~~~~ \sin(\theta_N)
    \end{bmatrix}
    \mathbf{v},
\end{aligned}
\end{equation}
while this equation can be expanded to 3D points as follows:
\begin{equation}
\begin{aligned}
\label{eq:3D_full}
    \begin{bmatrix}
        v_1
        \\
        \vdots
        \\
        v_N
    \end{bmatrix}
    =
    \begin{bmatrix}
        \frac{p_{1, x}}{p_{1, r}} ~~~~~ \frac{p_{1, y}}{p_{1, r}} ~~~~~ \frac{p_{1, z}}{p_{1, r}}
        \\
        \vdots
        \\
        \frac{p_{N, x}}{p_{N, r}} ~~~~~ \frac{p_{N, y}}{p_{N, r}} ~~~~~ \frac{p_{N, z}}{p_{N, r}}
    \end{bmatrix}
    \mathbf{v},
\end{aligned}
\end{equation}
where the $p_{j,r}=\sqrt{ p_{j,x}^2 + p_{j,y}^2 + p_{j,z}^2 }$ . 

Based on the equation \eqref{eq:2D_full} or \eqref{eq:3D_full}, whether the data is 2D or 3D, if the matrix including the angular information of points is referred to as $\mathbf{M}$, and the vector including the radial velocity of points is referred to as $\mathbf{v}_R$, the ego velocity $\mathbf{v}$ can be calculated by least-square optimization using a pseudo-inverse matrix as $\mathbf{v} = {(\mathbf{M}^\top \mathbf{M})}^{-1} \mathbf{M}^\top \mathbf{v}_R $, or based on \ac{SVD} of matrix $\mathbf{M}$. 


\subsection{Leg Kinematics and Ego Velocity}

Two key sensors for state estimation of the legged robot are the joint encoders and contact sensors. Joint encoders measure the absolute angle of each joint, being a key proprioceptive sensor for estimating the robot's pose. Contact sensors are positioned at every foot, measuring whether the foot is currently in contact or not.

Using the joint encoder measurement and the physical modeling of each node, the relative coordinate transformation between frames $\mathcal{F}^{i}$ and $\mathcal{F}^{i+1}$ at each joint can be calculated as follows:
\begin{equation}
\begin{aligned}
\label{eq:lie_group}
    \mathbf{T}_{i+1}^{i} &=
    \begin{bmatrix}
        \mathbf{R}_i ~~~ \mathbf{t}_i \\
        \mathbf{0}^\top ~~~ 1
    \end{bmatrix}
    \\
    \mathbf{R}_i &= \mathrm{Exp} ( { \alpha_{i} \; \mathbf{u}_i }^{\top} ) ,
\end{aligned}
\end{equation}
where $\alpha_i$ denotes the joint angle measured by the $i$-th joint encoder, and $\mathbf{u}_i$ is the basis vector of the $i$-th joint expressed in its local frame.  Using the $\mathbf{T}$ on every joint, the relative coordinate between the robot base frame and the end-effector (foot) frame can be calculated as follows:

\begin{equation}
\begin{aligned}
    \mathbf{T}_\text{foot}^\text{base} &= \mathbf{T}_\text{hip}^\text{base} \mathbf{T}_\text{knee}^\text{hip} \mathbf{T}_\text{foot}^\text{knee} ,
\end{aligned}
\end{equation}
assuming the hip and knee joints between the base coordinate and the foot coordinate. This chain-style frame calculation (\ie forward kinematics) allows us to calculate the relative pose of each end-effector based on the base frame of the robot. 

Under the no-slip assumption, when a single contact is made on any foot, the contacting frame of the foot should remain static until the contact state is resolved. Using this, during a foot is remaining contact state, the relative velocity of the base frame can be calculated by inverting the forward kinematics function to estimate the pose difference of the base frame during the contact state. 

From this forward kinematics, we will derive the ego-centric velocity and use it for the velocity factor, which will be detailed in Section~\secref{subsec:velocity_factor}.

\subsection{B-Spline Interpolation}
A $k$-order B-spline consists of several polynomial segments of degree $k-1$ with at most $C^{k-2}$ continuity~\citep{patrikalakis2002shape}. This continuous-time spline representation allows evaluating the state at arbitrary timestamps via smooth interpolation, which is particularly useful for fusing asynchronous sensors and reduces the need for explicit sensor-to-sensor time synchronization by enabling direct evaluation at each measurement time. Moreover, for legged robots where gait patterns introduce high-frequency noise, the smoothness of the spline acts as a low-pass filter, alleviating spike artifacts often observed in discrete-time alternatives.

Leveraging these advantages, we parameterize continuous-time trajectories of ego-centric velocity $\mathbf{v}(t) \in \mathbb{R}^{3}$ and rotation $\mathbf{R}(t) \in \mathrm{SO}(3)$ using third-order ($k$=3) uniform cumulative B-splines, each consisting of second-degree polynomial segments. Due to the continuity property, discontinuities in velocity and acceleration are effectively eliminated. This cumulative B-spline formulation offers $O(k)$ computational complexity in temporal derivatives, making it well suited for real-time odometry~\citep{sommer2020efficient}. 

At a given time $t \in [t_i, t_{i+1})$, the velocity $\mathbf{{v}}(t)$ depends exclusively on $k$ control points due to the local support property of B-splines, and its representation in matrix form is as follows:
\begin{equation}
\begin{aligned}
\label{eq:R3_spline}
\mathbf{{v}}(u) &=
    \begin{bmatrix} \mathbf{v}_i& \mathbf{d}_1^i & \cdots & \mathbf{d}_{k-1}^i\end{bmatrix} 
    \cdot \widetilde{M}^{(k)} \cdot {\mathbf{u}}^\top,
\end{aligned}
\end{equation} 
where $\mathbf{v}_i$ denotes the $i$-th control point of $\mathbf{{v}}(t)$, $\mathbf{d}_{j}^i = \mathbf{v}_{i+j} - \mathbf{v}_{i+j-1}$, ${\mathbf{u}} = \begin{bmatrix} 1 & u & \cdots & u^{k-1}\end{bmatrix}$, $u = (t-t_i)/(t_{i+1}-t_i)$ is the normalized time within the knot interval, and $ \widetilde{M}^{(k)}$ is the cumulative spline matrix that depends solely on the B-spline order. Since $u$ is the only term that depends on time, differentiating equation \eqref{eq:R3_spline} with respect to $t$ yields the following expression:
\begin{equation}
\begin{aligned}
\label{eq:R3_spline_derivate}
\mathbf{\dot{v}}(u) &=
    \frac{1}{\Delta t}\begin{bmatrix} \mathbf{v}_i& \mathbf{d}_1^i & \cdots & \mathbf{d}_{k-1}^i\end{bmatrix}
    \cdot \widetilde{M}^{(k)} \cdot {\mathbf{\dot{u}}}^\top.
\end{aligned}
\end{equation} 
At this point, we'd like to note that, in this paper, the velocity spline is constructed from ego-centric velocity; therefore, $\mathbf{\dot{v}}$ does not directly represent ego-centric acceleration. Computing ego-centric acceleration requires additional considerations about the angular velocity.

Analogous to the $\mathbb{R}^{3}$ case, a cumulative B-spline of order $k$ defined on a Lie group $\mathcal{L}$ with control points $\mathbf{R}_{0},\cdots, \mathbf{R}_{N}\in\mathcal{L}$ is expressed as:
\begin{equation}
\begin{aligned}
\label{eq:so3_spline}
     \mathbf {R}(u) &= \mathbf {R}_{i} \cdot \prod _{j=1}^{k-1}{\mathrm{Exp}\left(\lambda _{j}(u)\cdot \mathrm{Log}\left(\mathbf {R}_{i+j-1}^\top\mathbf {R}_{i+j}\right)\right)}\\
     \mathbf {\dot{R}}(u) &= \mathbf {R}_{i} \cdot 
     \prod _{j=1}^{k-1}\Biggl\{\prod _{l=1}^{j-1} A_l(u)\Biggr\} \dot{A}_j(u) \Biggl\{\prod _{l=j+1}^{k-1} A_l(u)\Biggr\}, \\
\end{aligned}
\end{equation} 
where $A_j(u)= {\mathrm{Exp}\left(\lambda _{j}(u)\cdot \mathrm{Log}\left(\mathbf {R}_{i+j-1}^\top\mathbf {R}_{i+j}\right)\right)}$, ${\lambda}(\tau) = \widetilde{M}^{(k)}\cdot {\mathbf{u}}(\tau)$, and $\lambda _{j}$ is $j-$th element of ${\lambda}$. More details can be found on ~\citep{sommer2020efficient}.

\section{Factor Graph Formulation}
\label{sec:factor_formulation}

\subsection{State Definition and Notation}
In this work, we adopt a robot-centric state to embed the local gravity within the system state, which is defined as follows:
\begin{eqnarray}
\label{eq:state}
	 \mathbf{x} &\triangleq& \begin{bmatrix}\mathbf{x_R}& \mathbf{x_v}&
     \mathbf{x_g}&
     \mathbf{x_{b_v}}&
     \mathbf{x_{b_a}}&
     | &
     \mathbf{{g}}^\mathtt{G}& \mathbf{\Bar{R}}^{\mathtt{I_0}}_{\mathtt{G}}
     \end{bmatrix} \text{, where}  \nonumber
     \\
     \mathbf{x_R} &\triangleq&
     \begin{bmatrix}
     \mathbf{{R}}^{\mathtt{G}}_\mathtt{I_0}&
     \cdots&
     \mathbf{{R}}^{\mathtt{G}}_\mathtt{I_{l-1}}
     \end{bmatrix}  \nonumber
     \\
     \mathbf{x_v} &\triangleq&
     \begin{bmatrix}
     \mathbf{{v}}_\mathtt{I_0}^\mathtt{I_0}&
     \cdots&
     \mathbf{{v}}_\mathtt{I_{m-1}}^\mathtt{I_{m-1}}
     \end{bmatrix} \nonumber
     \\
     \mathbf{x_g} &\triangleq&
     \begin{bmatrix}
     \mathbf{{g}}^\mathtt{I_0}&
     \cdots&
     \mathbf{{g}}^\mathtt{I_{n-1}}
     \end{bmatrix} \nonumber
     \\ 
     \mathbf{x_{b_v}} &\triangleq&
     \begin{bmatrix}
     \mathbf{b}_{v_0}&
     \cdots&
     \mathbf{b}_{v_{i-1}}
     \end{bmatrix} \in \mathbb{R}^{2 \cross i} \nonumber 
     \\
     \mathbf{x_{b_a}} &\triangleq&
     \begin{bmatrix}
     \mathbf{b}_{a_0}&
     \cdots&
     \mathbf{b}_{a_{i-1}}
     \end{bmatrix} \in \mathbb{R}^{3 \cross i} \nonumber 
     \\ 
     \mathbf{{g}}^\mathtt{G} &\triangleq& \begin{bmatrix}0 & 0 & 9.81\end{bmatrix}.
     \vspace{-8mm}
\end{eqnarray}
The system state $\mathbf{x}$ consists of B-spline control points and bias terms. $\mathbf{x_R}$, $\mathbf{x_v}$ and $\mathbf{x_g}$ denote the control points of the B-splines for the $\mathrm{SO}(3)$ orientation $\mathbf{R}(t)$, ego-centric velocity $\mathbf{v}(t)$, and local gravity vector $\mathbf{g}(t)$, respectively. For $\mathbf{g}(t)$, we adopt an $\mathbb{R}^3$ spline parameterization for practical implementation and debugging simplicity compared to an $\mathcal{S}^2$ spline. $\mathbf{R}^{\mathtt{G}}_\mathtt{I}$ represents the rotation of the \ac{IMU} frame expressed in the global frame, while $\mathbf{v}_\mathtt{I}^\mathtt{I}$ and $\mathbf{g}_\mathtt{I}$ denote the ego-centric velocity and the local gravity vector, both expressed in the \ac{IMU} frame. The bias terms $\mathbf{b}_v$ and $\mathbf{b}_a$ correspond to the velocity state bias and the \ac{IMU} accelerometer bias, respectively. The global coordinate frame ${\mathcal{F}}^\mathtt{G}$ is defined such that its $z$-axis is aligned with the global gravity vector $\mathbf{{g}}^\mathtt{G}$. The rotation matrix $\mathbf{\Bar{R}}^{\mathtt{I_0}}_\mathtt{G}$ is used to align the estimated global gravity vector to $\mathbf{{g}}^\mathtt{G}$. Further details on $\mathbf{{g}}^\mathtt{G}$ and $\mathbf{\Bar{R}}^{\mathtt{I_0}}_\mathtt{G}$ are included in Section~\secref{subsubsec:Aligning_gravity}.

To achieve sensor fusion with multiple sensor modalities, our algorithm employs an incremental factor graph optimization framework that leverages B-splines for optimizing each control point. The factor graph used in our approach comprises an \ac{IMU} factor, a velocity factor, and a gravity factor, collectively enabling robust and accurate state estimation. 

\subsection{IMU Factor}
We employ two types of \ac{IMU} factors: a gyroscope factor $\mathbf{r_\omega}$, and an accelerometer factor $\mathbf{r_a}$. Let $\tilde{\mathbf{a}}_{m_i}$ and $\tilde{\boldsymbol{\omega}}_{m_i}$ denote the acceleration and angular velocity measurements obtained from the \ac{IMU} accelerometer and gyroscope at time $t_i$, respectively. The IMU measurements are modeled as
\begin{align}
    \boldsymbol{\omega}_{m_i} &= \tilde{\boldsymbol{\omega}}_{m_i} - \mathbf{b}_{\omega_i} - \mathbf{n}_{\omega_i} \\
    \mathbf{a}_{m_i} &= \tilde{\mathbf{a}}_{m_i} - \mathbf{b}_{a_i} - \mathbf{n}_{a_i},
\end{align}
where $\boldsymbol{\omega}_{m_i}$ and $\mathbf{a}_{m_i}$ are the true angular velocity and acceleration at time $t_i$; 
$\mathbf{b}_{\omega_i}$, $\mathbf{b}_{a_i}$ are the gyroscope and accelerometer biases; and 
$\mathbf{n}_{\omega_i}$, $\mathbf{n}_{a_i}$ are zero-mean Gaussian noises. In our setup, the gyroscope bias is weakly observable, so we set $\mathbf{b}_{\omega_i}\!\equiv\!\mathbf{0}$, 
while the accelerometer bias $\mathbf{b}_{a_i}$ is modeled as a Gaussian-driven random walk. Building on these equations, the residual functions for both \ac{IMU} factors are formulated as follows, enabling direct optimization of control points for the continuous-time trajectory on rotation and velocity:
\begin{equation}
\begin{aligned}
  \label{eq:imu_factor}
     \mathbf{r_\omega}(t_i) &= \boldsymbol{\omega}(t) -\tilde{\boldsymbol{\omega}}_{m_i} 
    \\
    \mathbf{r_a}(t_i) &= (\boldsymbol{\omega}(t) \times\mathbf{{v}}(t) + \dot{\mathbf{v}}(t) - \mathbf{R}(t)^\top \mathbf{{g}}^\mathtt{G})+\mathbf{b}_{a_i} - \tilde{\mathbf{a}}_{m_i}.
\end{aligned}
\end{equation}

\subsection{Velocity Factor}
\label{subsec:velocity_factor}
Our system leverages two sources of ego-centric velocity: a radar-derived velocity and a leg kinematics-based velocity. For the radar sensor, assuming that the $j-$th target point of radar measurements $\tilde{\mathbf{p}}^{\mathtt{r}}_j$ is stationary, the Doppler measurement $\tilde{v}^\mathtt{r}_{m,j}$ equals the magnitude of the ego-velocity projected onto the unit line-of-sight vector to the target point and its conversion into the \ac{IMU} coordinate frame is expressed as:
\begin{eqnarray}\
\begin{aligned}
    \tilde{v}^\mathtt{r}_{m_i,j} &= - \frac{(\tilde{\mathbf{p}}^\mathtt{r}_j)^\top{}}{\left\| \tilde{\mathbf{p}}^\mathtt{r}_j\right\|} \left( {\mathbf{R}}^{\mathtt{I}}_\mathtt{r} \right)^{\top}\left({\mathbf{{v}}_i}+\lfloor \boldsymbol{\omega}^\mathtt{I}_i \rfloor_{\times}\mathbf{t}^\mathtt{I}_\mathtt{r}\right),
\end{aligned}
\end{eqnarray}
where $\left[ {\mathbf{R}}^{\mathtt{I}}_\mathtt{r}, \mathbf{t}^{\mathtt{I}}_\mathtt{r} \right]$ is the extrinsic matrix from radar frame $\mathcal{F}^\mathtt{r}$ to \ac{IMU} frame $\mathcal{F}^\mathtt{I}$ and $\mathbf{{v}}_i$,  $\boldsymbol{\omega}^\mathtt{I}_i$ are true ego-centric velocity and angular velocity at time $t_i$. Because static points predominate in most environments, this formulation provides highly robust velocity estimates; however, the limited point resolution of radar introduces noticeable inaccuracies, particularly along the elevation axis, as mentioned in Co-RaL~\citep{jung2024co}.

\begin{figure}[!t]
    \centering
    \subfigure[Body-centric Leg Locomotion]{
        \includegraphics[width=0.45\columnwidth]{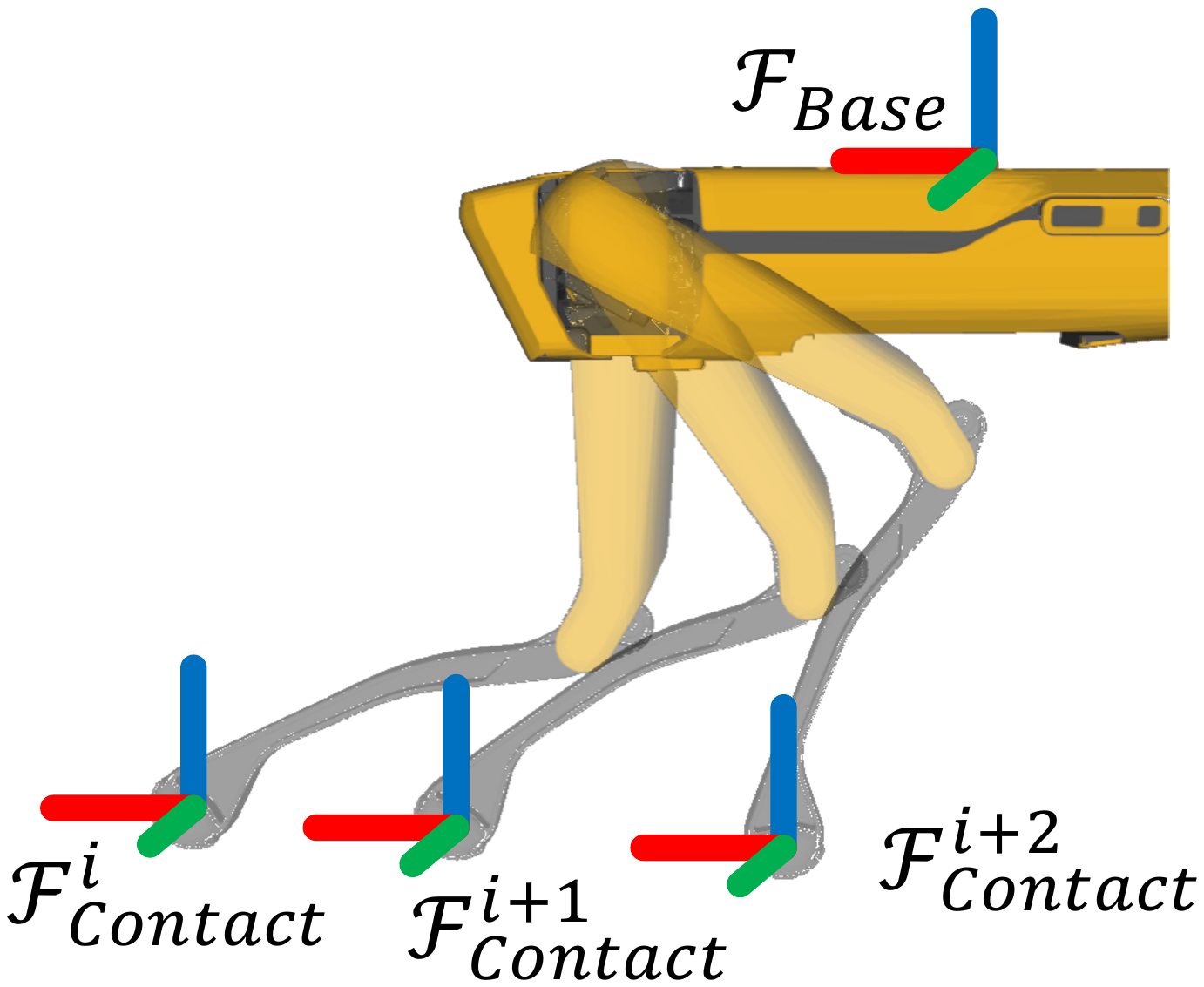}
        \label{fig:body_centric}
    }
    \subfigure[Contact-centric Leg Locomotion]{
        \includegraphics[width=0.48\columnwidth]{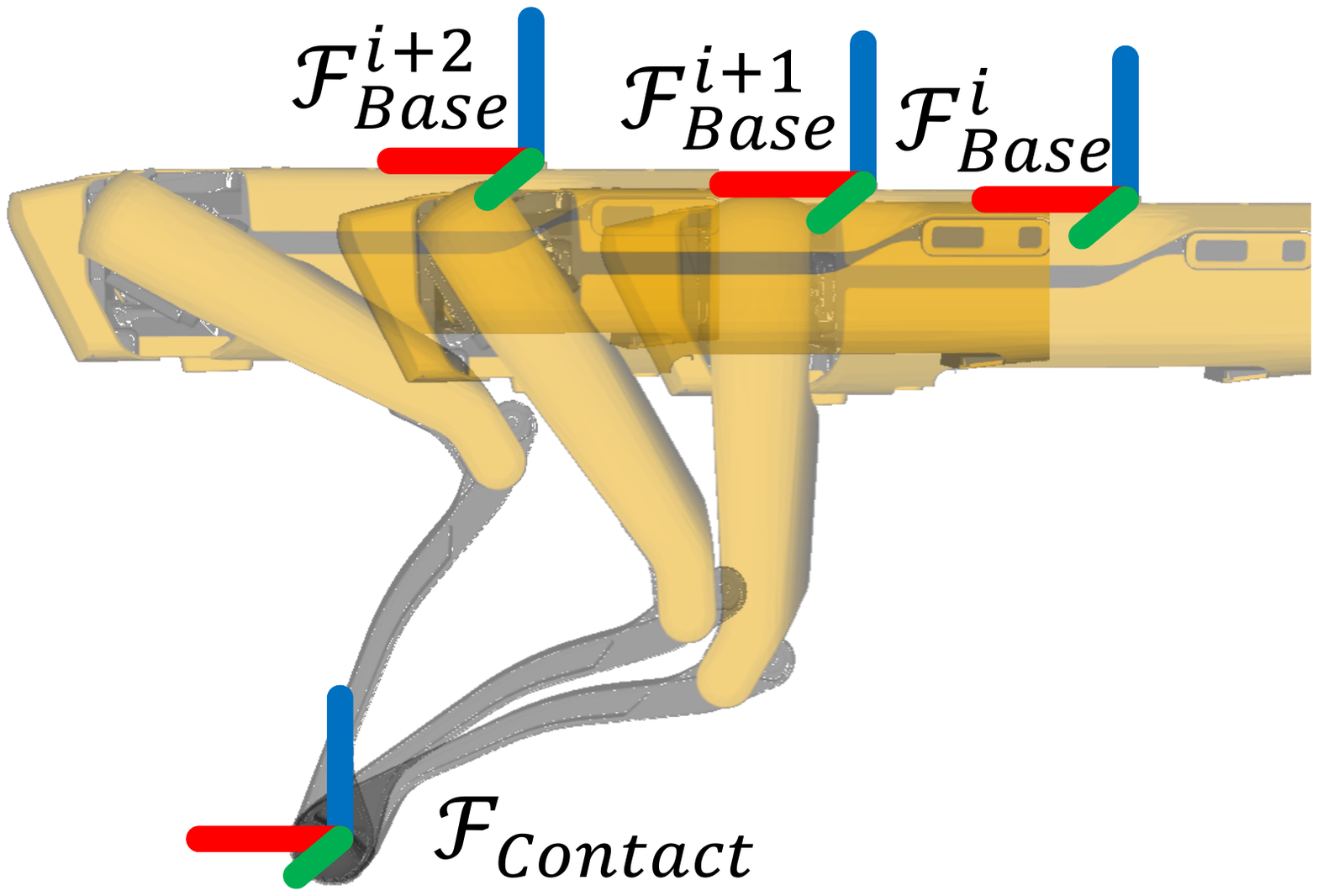}
        \label{fig:contact_centric}
    }
    \caption{Comparison between body-centric and contact-centric leg locomotion of a single leg during the contact state. \subref{fig:body_centric} On body-centric calculation, the end-effector position differs as time passes. Based on the contact information from the contact sensor positioned at every foot, every contact frame should remain static while the contact sensor is on. Therefore, using the forward kinematics, the ego-centric velocity of the robot base can be calculated as in \subref{fig:contact_centric}. }
    \label{fig:leg_vel}
\end{figure}

The translation between the robot’s body frame $\mathcal{F}^\mathtt{B}$ and the contact frame $\mathcal{F}^\mathtt{C}$ at time $t_i$ can be obtained using the following forward-kinematics function $\textbf{f}_{p}$ :
\begin{eqnarray}
    \label{eq:fkt} \mathbf{t}^{\mathtt{B_i}}_{\mathtt{C_i}}&=&\textbf{f}_{p} \left( {\tilde{\boldsymbol{\alpha}}_i} \right),
\end{eqnarray}
where the $\tilde{\boldsymbol{\alpha}}$ is the vector of joint angle measurements from the encoders on every joint at time $t_i$. Joint position and velocity are obtained from the leg-joint encoders, and the driver calculates which foot is in contact with the ground~\footnote{The Spot provides this information by measuring the ground reaction forces applied to each foot and provides this information}. These measurements enable computation of ego-centric velocity in the robot's body frame,
\begin{eqnarray} 
    \mathbf {v}_\mathtt{B_i}^\mathtt{B_i} = -\textbf{J}_{p}({\tilde{\boldsymbol{\alpha}}_i)} {\tilde{\dot{\boldsymbol{\alpha}}}_i}- \boldsymbol{\omega }^\mathtt{B}\times \textbf{f}_{p}({{\tilde{\boldsymbol{\alpha}}_i}}),
\end{eqnarray}
where $\textbf{J}_{p}(\cdot) \in \mathbb{R}^{3 \cross 3}$ is a Jacobian of $\textbf{f}_{p}(\cdot)$~\citep{wisth2022vilens}.
Empirically, the joint velocity reported by the encoders equals the finite difference approximation of the joint positions; however, under highly nonlinear joint trajectories, the velocity obtained in this manner diverges from the actual velocity. Therefore, we computed the ego-centric velocity by differentiating the leg translation function $\mathbf{f}_p({\tilde{\boldsymbol{\alpha}}})$ derived from joint positions only as follows:
\begin{eqnarray} 
    \mathbf {v}_\mathtt{B_i}^\mathtt{B_i} = 
    -\frac{\textbf {f}_{p}({{\tilde{\boldsymbol{\alpha}}_{i+1}}}) -  \textbf {f}_{p}({{\tilde{\boldsymbol{\alpha}}_i}})}
    {t_{i+1} - t_i},
\end{eqnarray}
\textbf{}which is intuitively explained in \figref{fig:leg_vel}. Using the contact-centric leg kinematic locomotion, the ego-centric velocity of the system can be estimated. While this approach provides accurate estimates when the contact frame remains fixed to the ground, practical deployments must account for leg slip. Accordingly, we augment the leg kinematics factor with a velocity bias term to account for slip-induced drift of the leg kinematics-based velocity, particularly on slippery or deformable surfaces. Because slippage occurs primarily along the horizontal $x-$ and $y-$directions, the velocity bias is modeled exclusively in these two dimensions. Taking these considerations into account, the corresponding leg kinematics factor $\mathbf{r_L}$ and the radar factor $\mathbf{r_R}$ are defined as
\begin{equation}
\begin{aligned}
  \label{eq:velocity_factor}
    \mathbf{r_L}(t_i) &=  \left(\mathbf{{R}}^{\mathtt{I}}_\mathtt{B}\right)^{\top} \left(\mathbf {v}(t_i) + \lfloor \boldsymbol{\omega}(t_i) \rfloor_{\times} \mathbf{t}^\mathtt{I}_\mathtt{B} \right) - \mathbf {v}_\mathtt{B_i}^\mathtt{B_i} -
    \begin{bmatrix}\mathbf{b}_{v}& 0\end{bmatrix}
    \\
    \mathbf{r_R}(t_i) &= \sum_{j \in [1, q]} \left[ - \frac{(\tilde{\mathbf{p}}^\mathtt{r}_j)^\top{}}{\left\| \tilde{\mathbf{p}}^\mathtt{r}_j\right\|} \left( {\mathbf{R}}^{\mathtt{I}}_\mathtt{r} \right)^{\top}\left({\mathbf{{v}}(t_i)}+\lfloor \boldsymbol{\omega}(t_i) \rfloor_{\times} \mathbf{t}{}^\mathtt{I}_\mathtt{r}\right) \right]
    \\
    &~+ \sum_{j \in [1, q]} \left[ - \tilde{v}^\mathtt{r}_{m_i,j} \right]    
    ,
\end{aligned}
\end{equation}
\normalsize
where $\left[ {\mathbf{R}}^{\mathtt{I}}_\mathtt{B}, \mathbf{t}^\mathtt{I}_\mathtt{B} \right]$ is the extrinsic matrix from robot base frame to \ac{IMU} frame, and $q$ refers number of radar targets.

\subsection{Gravity Factor}

\begin{figure}[!t]
    \centering
    \includegraphics[width=0.8\columnwidth]{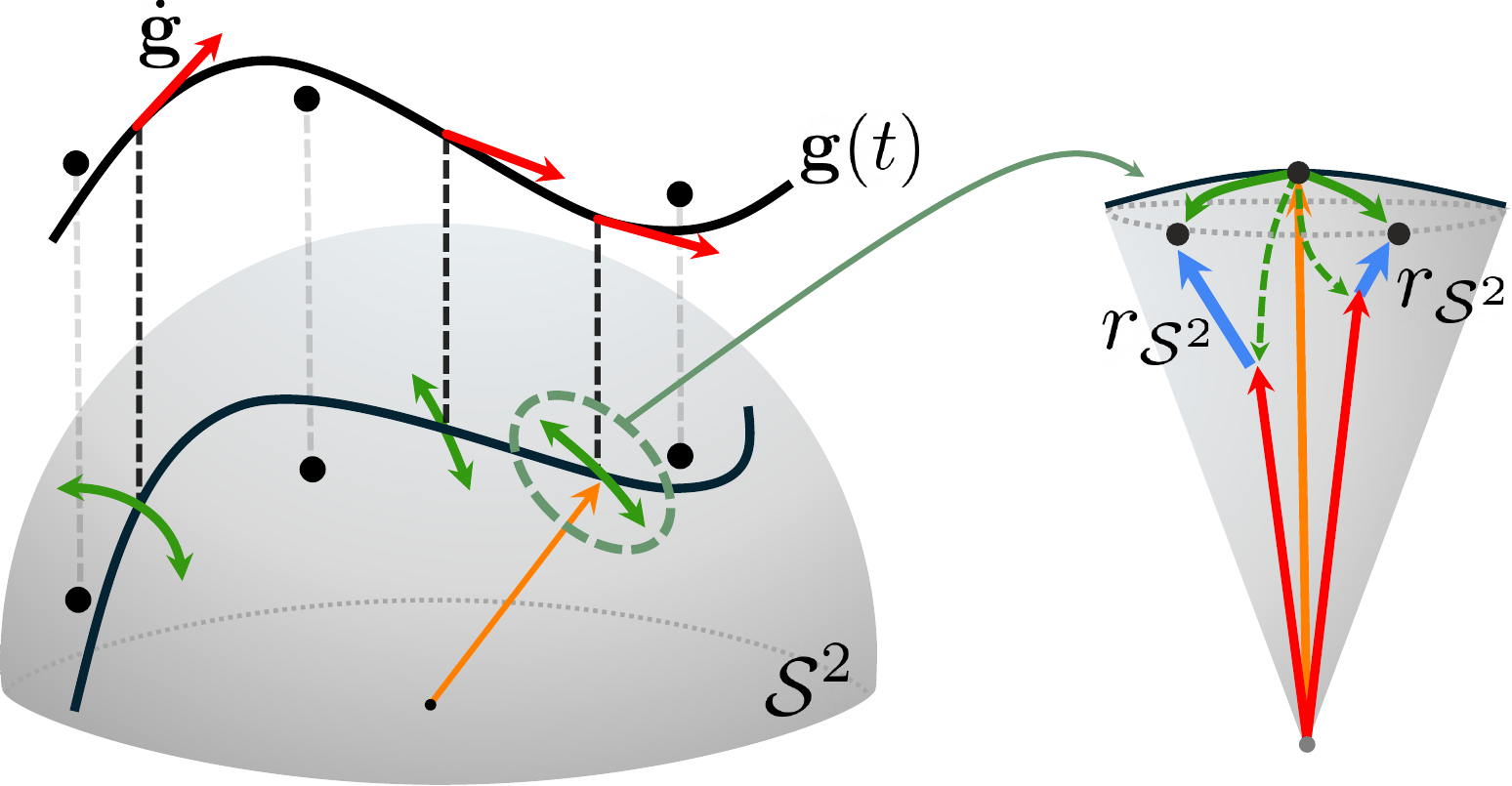}
    \caption{Soft $\mathcal{S}^2$-constrained gravity factor $r_{\mathcal{S}^2}$. Prior to optimization, the orange vectors are initialized through gravity spline extrapolation. During optimization, the factor (blue) constrains vectors that would otherwise drift in the $\mathbb{R}^3$ manifold (red), steering them to lie on the $\mathcal{S}^2$ surface (green).}
    \label{fig:soft_factor}
\end{figure}

To mitigate the \ac{LiDAR}-dependent and discrete-time sensor fusion limitations of GaRLIO~\citep{noh2025garlio} and enable robust gravity estimation for legged robot platforms, we adopt a B-spline-based, velocity-aware gravity factor $\mathbf{r_G}$ to parameterize continuous-time state functions as follows:
\begin{eqnarray}
\begin{aligned}
   \mathbf{r_G}(t_i) = \sum_{i,\;j \in W} \left\| \mathbf{g}(t_i) - 
   \frac{{\tilde{\mathbf{R}}_{m_i}}^\top\tilde{\mathbf{R}}_{m_j}\mathbf{v}(t_{j}) - \mathbf{v}(t_{i}) - \beta_{j}^i}{t_{j}-t_{i}} \right\|^2,
\end{aligned}
\end{eqnarray}
where $t_i$ and $t_j$ denote the timestamps of the $i$-th and $j$-th IMU measurements included in window $W$, respectively. $\tilde{\mathbf{R}}_{m}$ denotes the onboard AHRS orientation measurement obtained from internal filter of IMU by fusing magnetometer with standard inertial measurements, while $\beta_{j}^i=\int_{t\in[{t_i},{t_{j}})}\tilde{\mathbf{R}}_m(t)(\tilde{\mathbf{a}}_t-{\mathbf{b}}_{a_t})dt$.
Furthermore, to mitigate erroneous gravity estimation caused by \ac{IMU} noise, we also introduce a gravity spline equipped with a smoothing constraint $\mathbf{r_{\mathcal{S}^2}}$ (\figref{fig:soft_factor}). The gravity spline control points are subject to a fixed-norm constraint. Still, since we parameterize local gravity using splines on an $\mathbb{R}^3$ manifold rather than an $\mathcal{S}^2$ manifold, the interpolated B-spline segments between control points are not explicitly norm-constrained. To resolve this, a novel first-order derivative factor is defined and applied to the spline as follows: 
\begin{eqnarray}
\label{eq:s2_manifold}
\begin{aligned}
    \mathbf{r_{\mathcal{S}^2}}(t) =  \dot{\mathbf{g}}(t)+\lfloor \boldsymbol{\omega}(t) \rfloor_{\times}\mathbf{g}(t),
\end{aligned}
\end{eqnarray}
which enforces a soft norm constraint throughout the spline trajectory.

\section{Continuous Radar-Leg-IMU Fusion}
\begin{figure}[!t]
    \centering
    \includegraphics[width=\columnwidth]{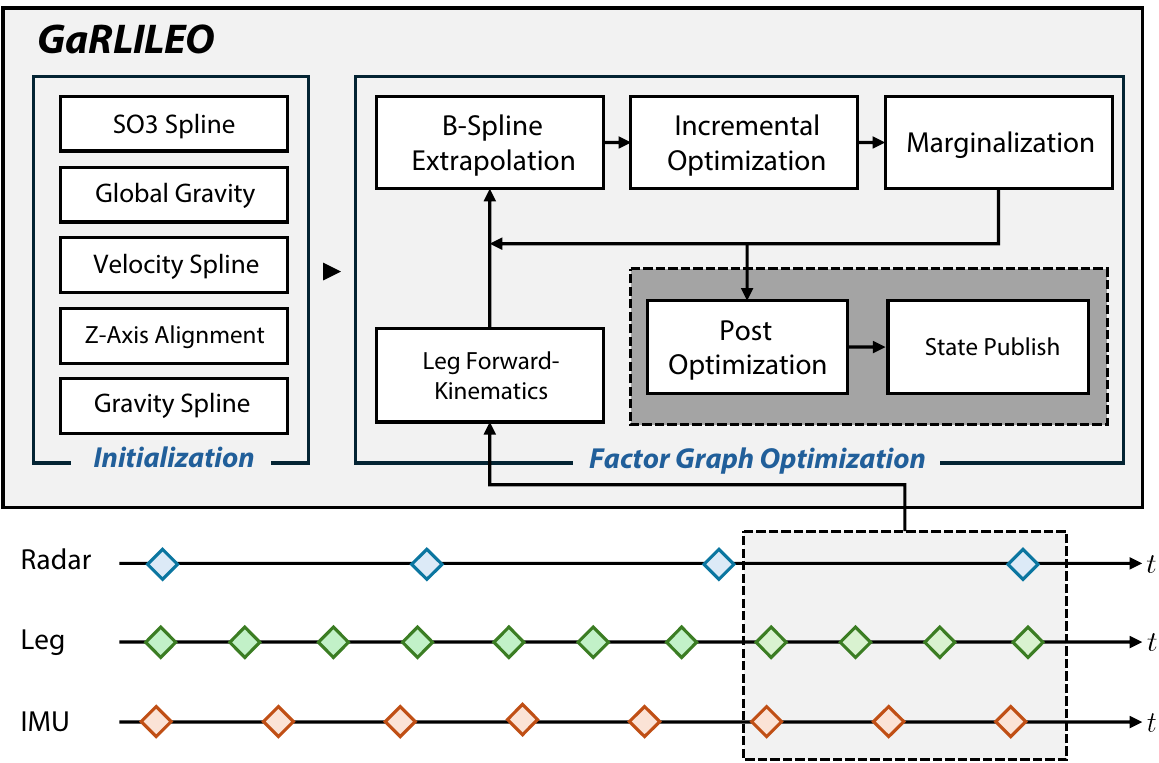}
    \caption{Overview pipeline of GaRLILEO}
    \label{fig:overview}
\end{figure}

\figref{fig:overview} illustrates the overall pipeline of GaRLILEO. The algorithm performs state estimation using three distinct splines representing $\mathrm{SO}(3)$ rotation, ego velocity, and local gravity. In the initialization stage, the three splines and global gravity vector are recovered from the input data, similar to River~\citep{chen2024river}. After transforming the data into a gravity-aligned coordinate frame, a tightly coupled incremental factor graph optimization is executed. Subsequently, a rotation refinement procedure using the estimated local gravity is applied to the marginalized older states, followed by odometry estimation through a dead reckoning approach.

\subsection{Initialization}
\subsubsection{SO(3) Spline Initialization}
The initial $\mathrm{SO}(3)$ spline is constructed by applying the factor $\mathbf{r_\omega}$, using \ac{IMU} gyroscope measurements as constraints. The initial global frame is set as the initial frame of \ac{IMU}.
The initialization is formulated as
\begin{equation}
\begin{aligned}
    \mathop {\min } 
     \limits_{\mathbf{x_R}} \quad &
      {\sum\limits_{k \in W_{init}} {\left\| {{{\mathbf{r_\omega}(t_k)}}} \right\|^2}}.
\end{aligned}
\end{equation}

\subsubsection{Global Gravity Initialization}
Our method refines rotation by estimating local gravity and comparing it with global gravity. Therefore, accurate estimation of the global gravity reference is critical. For global gravity estimation, two different methods are popularly exploited: dynamic initialization and stationary initialization methods.

The first type is \textit{dynamic initialization} introduced in River~\citep{chen2024river}, which estimates global gravity using an initial $\mathrm{SO}(3)$ spline and radar measurements. Here, ego-velocities are computed from radar scans collected during the initialization phase, and then are converted into the global frame using the initial $\mathrm{SO}(3)$ spline. The global gravity is estimated using the following equation:
\begin{equation}
\begin{aligned}
    \mathop {\min } 
     \limits_{\mathbf{g}_\mathtt{I_{0}}} \quad &
      {\sum\limits_{k \in W_{init}} {\left\| {{{\mathbf{r_{dg}}(t_k)}}} \right\|^2}},\\
   \textrm{s.t.} \quad & \left\|\mathbf{g}^\mathtt{I_{0}}\right\| = 9.81 ,\\
   \textrm{where} \quad &\mathbf{r_{dg}}(t_k) = \frac{ {\mathbf{\alpha}_\mathtt{k+1}^\mathtt{k}} - \mathbf{R}(t_{\mathtt{k+1}}) {\mathbf{v}}_\mathtt{I_{k+1}}^\mathtt{I_{k+1}} + \mathbf{R}(t_{\mathtt{k}}) {\mathbf{v}}_\mathtt{I_{k}}^\mathtt{I_{k}}}{t_{k+1}-t_{k}} - {\mathbf{g}^\mathtt{I_{0}}} \\ 
   &\mathbf{\alpha}_\mathtt{k+1}^\mathtt{k} =\int_{t\in[\mathtt{r_k}, \mathtt{r_{k+1}})}\mathbf{R}(t)(\tilde{\mathbf{a}}_t-\widehat{\mathbf{b}}_{a_t})dt.
\end{aligned}
\end{equation}
However, due to the highly nonlinear characteristics of \ac{IMU} acceleration measurements from contact impact during legged robot locomotion, estimating global gravity dynamically at the radar frame rate using only the initial $\mathrm{SO}(3)$ spline results in inaccurate estimation. Therefore, a more stable method for estimating the global gravity is required on the legged robot \ac{UGV} system. 

Instead, we adopted a \textit{stationary initialization} method for precise global gravity estimation. By computing the mean and variance of ego-velocities derived from accumulated radar scans during the initialization phase, the stationary condition can be determined using the following equation:
\begin{eqnarray}
\begin{aligned}
    \text{mean}({\mathbf{v}}_\mathtt{I_{k}}^\mathtt{I_{k}}) < \tau_1 \;\;\; \mathrm{\&}
    \;\;\;
     \mathrm{tr}(\mathrm{Var}({\mathbf{v}}_\mathtt{I_{k}}^\mathtt{I_{k}})) < \tau_2.
\end{aligned}
\end{eqnarray}
Under stationary conditions, the accelerometer measurements directly correspond to gravity. The mean of these measurements $\Bar{\tilde{\mathbf{a}}}$ serves as a prior factor, which is combined with the factor from the dynamic initialization method to calculate global gravity:
\begin{eqnarray}
\begin{aligned}
   \mathbf{r_{sg}}(t_k) =& \; w_1 \cdot(\Bar{\tilde{\mathbf{a}}}- {\mathbf{g}^\mathtt{I_{0}}}) \\+& \; w_2 \cdot(\frac{ {\mathbf{\alpha}_\mathtt{k+1}^\mathtt{k}} - \mathbf{R}(t_{\mathtt{k+1}}) {\mathbf{v}}_\mathtt{I_{k+1}}^\mathtt{I_{k+1}} + \mathbf{R}(t_{\mathtt{k}}) {\mathbf{v}}_\mathtt{I_{k}}^\mathtt{I_{k}}}{t_{k+1}-t_{k}} - {\mathbf{g}^\mathtt{I_{0}}}).
\end{aligned}
\end{eqnarray}
Since this approach yields a more stable and accurate global gravity estimation compared to dynamic initialization on a legged robot, every sequence included in the dataset of this paper begins stationary, thereby enabling the use of static initialization.

\begin{figure}[!t]
    \centering
    \includegraphics[width=\columnwidth]{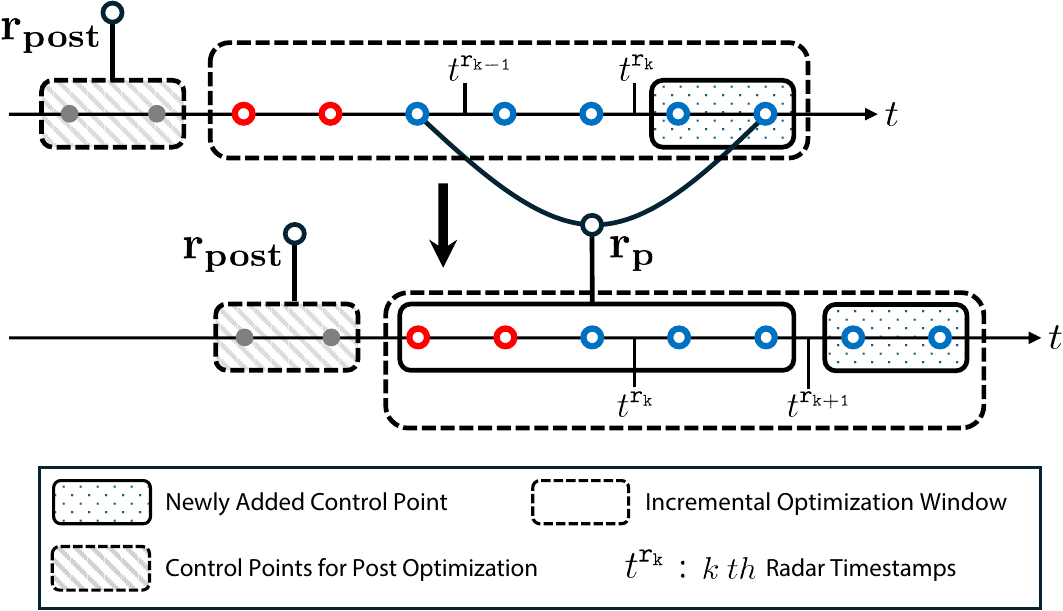}
    \caption{Factor Graph Overview. At each iteration, the number of control points marginalized (grey) equals the number newly added in the preceding window (red); the remaining control points are carried forward as a prior factor for the next solve (blue). The marginalized control points (grey) are then subjected to a post-optimization stage that refines the SO(3) spline.}
    \label{fig:factor_graph}
\end{figure}

\subsubsection{Velocity Spline Initialization}
The radar, \ac{IMU}, and leg kinematics information are leveraged to construct the ego-centric velocity spline. During this process, velocity and acceleration biases are simultaneously initialized by solving the following \ac{LSQ} problem:
\begin{equation}
\begin{aligned}
 \mathop {\min } 
 \limits_{\mathbf{X}} \quad &
 \Biggl\{ {\sum\limits_{k \in I} {{w_\omega}} {\left\| {{{\mathbf{r_\omega}(t_k)}}} \right\|^2} +}  
 {\sum\limits_{k \in I} {{w_a}} {\left\| {{{\mathbf{r_a}(t_k)}}} \right\|^2}} + \\
 & \quad \quad \quad \quad \quad {\sum\limits_{k \in L} {{w_L}} {\left\| {{{\mathbf{r_L}(t_k)}}} \right\|^2}} + 
 {\sum\limits_{k \in R} {{w_R}} {\left\| {{{\mathbf{r_R}(t_k)}}} \right\|^2}}\Biggr\}.
\end{aligned}
\end{equation}
To maintain consistency, the $\mathrm{SO}(3)$ rotation spline is fixed during the initialization process of the velocity spline.

\subsubsection{Z-Axis Alignment with Global Gravity}
\label{subsubsec:Aligning_gravity}
The gravity vector has only 2-\ac{DoF} observations, roll and pitch direction~\citep{kubelka2022gravity}. Thus, to ensure optimization updates are confined to observable axes without explicit constraints, the coordinate frame is transformed by aligning its $z$-axis with the global gravity vector. Consequently, the $\mathrm{SO}(3)$ splines undergo the same coordinate transformation:
\begin{equation}
\begin{aligned}
       \mathbf{x_R} = & \; (\mathbf{\Bar{R}}^\mathtt{I_0}_{\mathtt{G}})^\top \cdot \mathbf{x_R} \cdot \mathbf{\Bar{R}}^\mathtt{I_0}_{\mathtt{G}}, \\
      \textrm{where} \quad \mathbf{{g}}^\mathtt{G} = & \;(\mathbf{\Bar{R}}^\mathtt{I_0}_{\mathtt{G}})^\top \cdot \mathbf{{g}}^\mathtt{I_0} .
\end{aligned}
\end{equation}

\subsubsection{Gravity Spline Initialization}
The final phase of initialization is on gravity spline. We initialize the gravity spline by recovering it through control points computed as 
\begin{equation}
    \mathbf{{g}}^\mathtt{I_k} = (\mathbf{{R}}^\mathtt{G}_\mathtt{I_k})^\top\mathbf{{g}}^\mathtt{G},
\end{equation}
using the previously initialized $\mathrm{SO}(3)$ splines.

\subsection{Factor Graph Optimization}
\subsubsection{Incremental Optimization}
Following initialization and spline recovery, incremental factor graph optimization is performed on the B-spline control points, tightly coupling \ac{IMU}, leg kinematics, and radar measurements.  When new control points are introduced, an equal number of the oldest points in the window are marginalized, and the information of the remaining non-marginalized control points is condensed into a prior factor and propagated to the next optimization window (\figref{fig:factor_graph}). The end time of the window is determined by the timestamp of the incoming radar topic, and optimization begins after the three B-splines are linearly extended and their control points appended. The factor graph comprises both gravity factors $\mathbf{r_G}, \mathbf{r_{\mathcal{S}^2}}$; an IMU gyroscope factor $\mathbf{r_\omega}$; radar and leg kinematics velocity factors $\mathbf{r_R}, \mathbf{r_L}$; a bias-prior factor $\mathbf{r_b}$; a marginalization-prior factor $\mathbf{r_p}$; and an end-tail factor $\mathbf{r_e}$. The optimization problem associated with the proposed system is defined as follows:
\begin{equation}
\begin{aligned}
\label{eq:incremental_optim}
 \mathop {\min } 
 \limits_{\mathbf{X}}  \sum\limits_{k \in W_i} \Bigl\{ &{ {{w_g}} \;{ {{{\mathbf{r_G}(t_k)}}} } + {{w_{\mathcal{S}^2}}} {\left\| {{{\mathbf{r_{\mathcal{S}^2}}(t_k)}}} \right\|^2}} + {{{w_\omega}} {\left\| {{{\mathbf{r_\omega}(t_k)}}} \right\|^2}} + \\&{{{w_L}} {\left\| {{{\mathbf{r_L}(t_k)}}} \right\|^2}} + 
 {{{w_R}}\cdot \rho _{r}({\left\| {{{\mathbf{r_R}(t_k)}}} \right\|^2}})\Bigr\} \\ &+{{{w_b}} {\left\| {{{\mathbf{r_b}_i}}} \right\|^2}}+{{{w_p}} {\left\| {{{\mathbf{r_p}_i}}} \right\|^2}} + 
 {{{w_e}} {\left\| {{{\mathbf{r_e}_i}}} \right\|^2}},
 \\
 \textrm{s.t.} \quad & \left\|\mathbf{{g}}^\mathtt{I_i}\right\| = 9.81,
\end{aligned}
\end{equation}
\normalsize
where $\rho_r$ is the Cauchy loss function, employed to mitigate the influence of non-static object-oriented radar points. The bias-prior factor and the end-tail factor, which are not addressed in Section~\secref{sec:factor_formulation}, are defined as follows:
\begin{equation}
\begin{aligned}
  \label{eq:bias_end_factor}
    \mathbf{r_b}_i &=  \begin{bmatrix} \mathbf{b}_{a_{i}} - \mathbf{b}_{a_{i-1}} \\ \mathbf{b}_{v_{i}} - \mathbf{b}_{v_{i-1}} \end{bmatrix}
    \\
    \mathbf{r_e}_i &= {\begin{bmatrix}{{\mathbf{{v}}_\mathtt{I_{m-1}}^\mathtt{I_{m-1}}}}-2\cdot {{\mathbf{{v}}_\mathtt{I_{m-2}}^\mathtt{I_{m-2}}}}+{{\mathbf{{v}}_\mathtt{I_{m-3}}^\mathtt{I_{m-3}}}} \\ \mathrm{Log}\left(\left(\mathbf{{R}}^\mathtt{G}_\mathtt{I_{l-2}}\right) ^\top \mathbf{{R}}^\mathtt{G}_\mathtt{I_{l-3}} \left(\mathbf{{R}}^\mathtt{G}_\mathtt{I_{l-2}}\right) ^\top \mathbf{{R}}^\mathtt{G}_\mathtt{I_{l-1}}\right) \end{bmatrix}}.
\end{aligned}
\end{equation}

\begin{figure}[!t]
    \centering
    \includegraphics[width=\columnwidth]{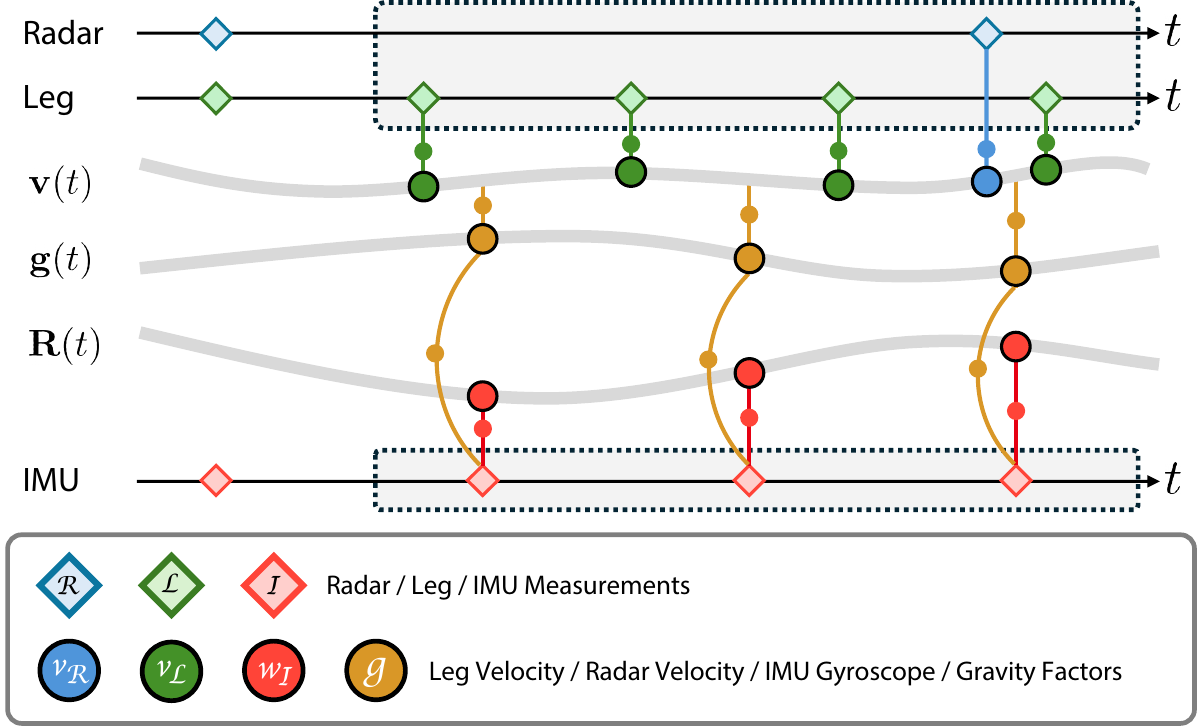}
    \caption{Relationship between sensor data and splines during incremental optimization. Gray-shaded boxes indicate sensor measurements active within the sliding optimization window; colored lines denote inter-spline and measurement-spline dependencies. In GaRLILEO, velocity is decoupled from the IMU, constructing a continuous-time ego-velocity spline from radar and leg-kinematics measurements. This decoupling is particularly advantageous for legged robots, where ground contact induces noisy accelerations on \ac{IMU}. }
    \label{fig:increment_opti}
\end{figure}

Dependencies between sensor measurements and each spline during the optimization of \eqref{eq:incremental_optim} are illustrated in \figref{fig:increment_opti}. To mitigate the contact-induced noise of \ac{IMU} acceleration degrading the velocity spline, we decoupled the velocity spline from \ac{IMU}. Instead, we leverage continuous-time ego-centric velocity spline generated from \ac{SoC}-radar and leg kinematics measurements, improving the robustness of local gravity estimation compared with prior methods. Within each window $W_i$, the biases $b_{v_{i}}, b_{a_{i}}$ are assumed to remain constant, and an end-tail factor minimizes endpoint instability of the splines~\citep{chen2024river}. 

\subsubsection{Marginalization}
\label{subsec:marginalization}
At each optimization step, we marginalize out the states and measurements that moved outside of the sliding window, summarizing them into a single prior factor. By reusing historical constraints in this condensed form, the estimator retains observability of active variables while bounding the graph size, achieving computational efficiency for real-time performance. By linearizing all factors about the current best estimate of the $i$-th window, we obtain an equation as follows:
\begin{equation}
\begin{aligned}
    \begin{bmatrix}\mathbf {H}_{\alpha \alpha } & \mathbf {H}_{\alpha \beta } \\ \mathbf {H}_{\beta \alpha } & \mathbf {H}_{\beta \beta } \end{bmatrix} \begin{bmatrix}\mathbf {x}_\alpha \\ \mathbf {x}_\beta \end{bmatrix} = \begin{bmatrix}\mathbf {b}_\alpha \\ \mathbf {b}_\beta \end{bmatrix},
\end{aligned}
\end{equation}
where $\mathbf{x}_\beta$ represents the states which are marginalized out, while $\mathbf{x}_\alpha$ indicates the states retained in the optimization. By applying the Schur complement~\citep{sibley2010sliding}, we reformulate the equation as follows:
\begin{equation}
\begin{aligned}
\label{eq:schur}
    (\mathbf {H}_{\alpha \alpha }-\mathbf {H}_{\alpha \beta }\mathbf {H}_{\beta \beta }^{-1}\mathbf {H}_{\beta \alpha }) \cdot \mathbf {x}_\alpha = \mathbf {b}_\alpha-\mathbf {H}_{\alpha \beta }\mathbf {H}_{\beta \beta }^{-1}\mathbf {b}_\beta,
\end{aligned}
\end{equation}
leading to the marginalization-prior factor $\mathbf{r_p}$ as:
\begin{equation}
\begin{aligned}
    \mathbf{r_p}_i = (\mathbf {H}_{\alpha \alpha }-\mathbf {H}_{\alpha \beta }\mathbf {H}_{\beta \beta }^{-1}\mathbf {H}_{\beta \alpha }) \cdot (\mathbf {x}_{\alpha,i} -\mathbf {x}_{\alpha,i-1})\\  - 
    (\mathbf {b}_\alpha-\mathbf {H}_{\alpha \beta }\mathbf {H}_{\beta \beta }^{-1}\mathbf {b}_\beta).
\end{aligned}
\end{equation}

\subsubsection{Post Optimization}
After incorporating the optimized local gravity state from \eqref{eq:incremental_optim}, we refine the rotation states of the relevant static control points. This procedure is applied to the control points marginalized in the previous optimization window, as shown in \figref{fig:factor_graph}. In this step, we update only the rotation states while keeping others fixed:
\begin{equation}
\begin{aligned}
    \mathop {\min } 
     \limits_{\mathbf{x_R}} \quad &
      {\sum {\left\| {{{\mathbf{r_{post}}(k)}}} \right\|^2}}, \\
       \textrm{where}
       \quad &\mathbf{r_{post}}(k) = \mathbf{R}^\mathtt{G}_{\mathtt{I_k}} {\mathbf{g}_\mathtt{I_{k}}} - {\mathbf{g}_\mathtt{G}}.
\end{aligned}
\end{equation}
Since the global $z$-axis is aligned with gravity, the above equation provides information only about roll and pitch but not yaw; rotation about the gravity axis remains unobservable. After the post-optimization step, the robot's pose is estimated using dead reckoning based on the optimized states. The ego-centric velocity is first transformed into the global frame, and subsequently, the robot position is computed using the following equation:
\begin{equation}
\begin{aligned}
    \mathbf{p}(t_k) = \mathbf{p}(t_{k-1})+\int_{t\in[{t_{k-1}},{t_{k}})}\mathbf{R}(t) \cdot \mathbf{v}(t) dt.
\end{aligned}
\end{equation}

\begin{figure*}[t!]
    \centering
    \includegraphics[width=0.8\textwidth]{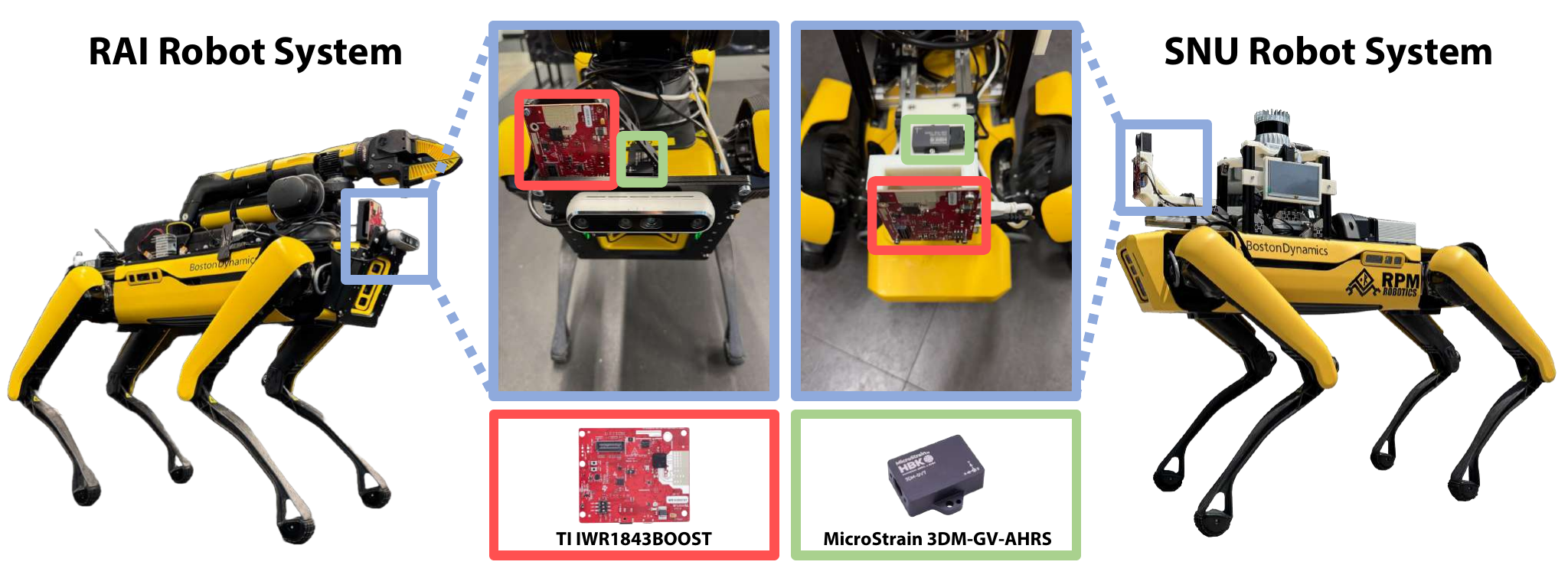}
    \caption{SNU and RAI sensor system deployment. Both systems include the same TI-mmWave radar, Microstrain IMU, and Boston Dynamics Spot quadrupedal robot, but are attached with different extrinsics. }
    \label{fig:cali}
\end{figure*}

\begin{table*}[t]
\small\sf\centering
\caption{The Sensor Specifications.}
\label{tab:sensor}

\resizebox{0.85\textwidth}{!}{
\begin{tabular}{llcccc}
\toprule
\hline
 
Sensor &
Manufacture &
Model &
Topic name &
Frequency&
Description\\

 \hline
Legged robot&
Boston Dynamics &
Spot&
{\begin{tabular}[c]{@{}c@{}}/joint\_states\\/spot/status/feet \end{tabular}}
&
{\begin{tabular}[c]{@{}c@{}} \unit{150}{Hz} \\ \unit{150}{Hz} \end{tabular}}&
\multirow{3}{*}{Sensor Measurements}
\\

Radar&
Texas Instruments &
IWR1843BOOST &
/ti\_mmwave/radar\_scan\_pcl\_0&
\unit{20}{Hz}&
\\

IMU &
MicroStrain &
3DM-GV7-AHRS &
/imu & 
\unit{100}{Hz}&
\\

\hline

LiDAR&
Ouster&
OS1-32&
/ouster/points& 
\unit{10}{Hz}&
\multirow{2}{*}{Ground-truth Reference} \\

Laser scanner&
Leica&
RTC360&
-&
-&
\\

\hline
\bottomrule
\end{tabular}

}
\end{table*}

\begin{table*}[!t]
\small\sf\centering
\caption{The Description for Each Sequence.}
\label{tab:sequence}

\resizebox{0.9\textwidth}{!}{
\begin{tabular}{l|ccc|cc|ccc}
\toprule
 \hline
Sequence & Path Length (\m) & Elevation Change (\m) & Duration (\s) & Outdoor & Indoor & Stair & Slope & \# of Loop \\ \hline

\texttt{Atrium} & 109.93 & - & 124.50 & \xmark & \cmark & \xmark & \xmark & 1\\
\texttt{BridgeLoop} & 161.17 & 1.72 & 187.20 & \xmark & \cmark & \cmark & \cmark & 3 \\
\texttt{CorriLoop} & 208.68 & - & 229.40 & \xmark & \cmark & \xmark & \xmark & 2 \\
\texttt{BiCorridor} & 240.82 & 4.72 & 277.29 & \xmark & \cmark & \cmark & \xmark & 1 \\
\texttt{Downstair} & 233.75 & 8.81 & 270.90 & \xmark & \cmark & \cmark & \xmark & 2 \\
\texttt{Upstair} & 197.22 & 9.37 & 227.89 & \xmark & \cmark & \cmark & \cmark & 1 \\ \hline
\texttt{SlopeStair} & 273.37 & 10.06 & 307.49 & \cmark & \cmark & \cmark & \cmark & 1\\
\texttt{Overpass} & 169.17 & 7.23 & 213.49 & \cmark & \xmark & \cmark & \xmark & 1 \\
\texttt{Tunnel}& 247.94 & - & 277.00 & \cmark & \cmark & \xmark & \xmark & 1 \\
\texttt{Quad} & 447.83  & 10.72 & 503.69 & \cmark & \xmark & \cmark  & \cmark & 1 \\ \hline
\texttt{MoCap-E} & 44.91 & 0.57 & 139.10 & \xmark & \cmark & \cmark & \cmark & 2 \\
\texttt{MoCap-H} & 42.48 & 0.60 & 79.46 & \xmark & \cmark & \cmark & \cmark & 2 \\

\hline
\bottomrule
\end{tabular}
}
\end{table*}

\begin{figure*}[!t]
    \centering
    \subfigure[\texttt{Upstair}]{
        \includegraphics[width=0.19\textwidth]{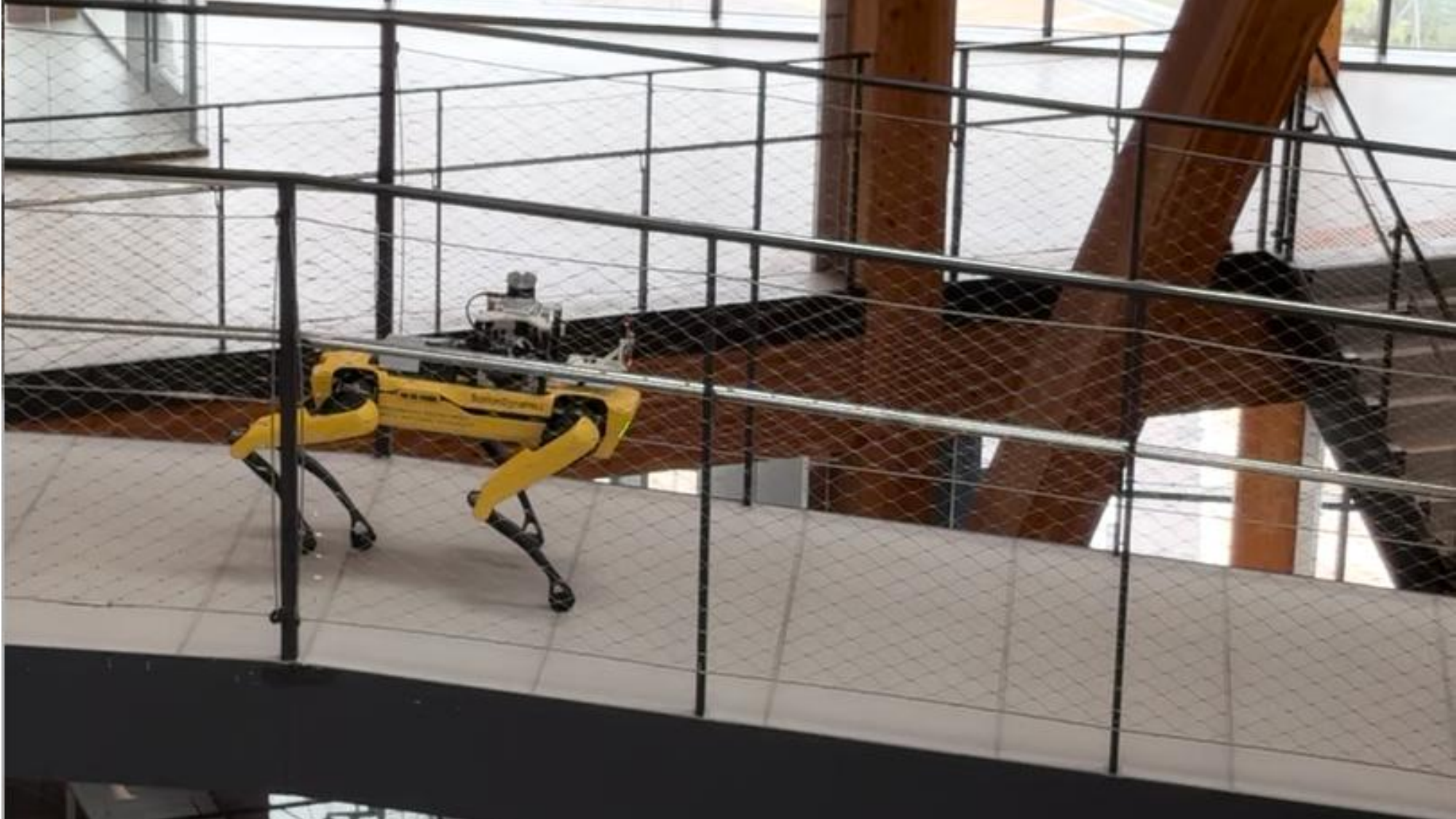}
        \label{fig:env_tower}
    }%
    \subfigure[\texttt{Quad}]{
        \includegraphics[width=0.19\textwidth]{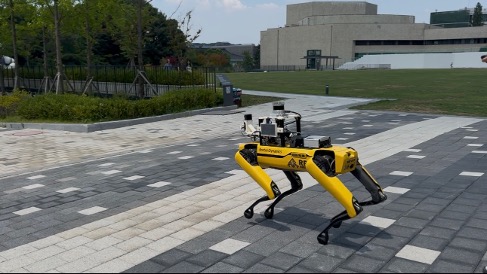}
        \label{fig:env_grass}
    }%
    \subfigure[\texttt{Overpass}]{
        \includegraphics[width=0.19\textwidth]{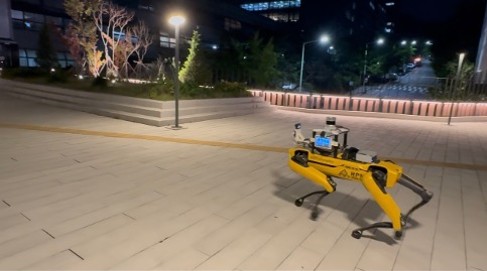}
        \label{fig:env_bridge}
    }%
    \subfigure[\texttt{Tunnel}]{
        \includegraphics[width=0.19\textwidth]{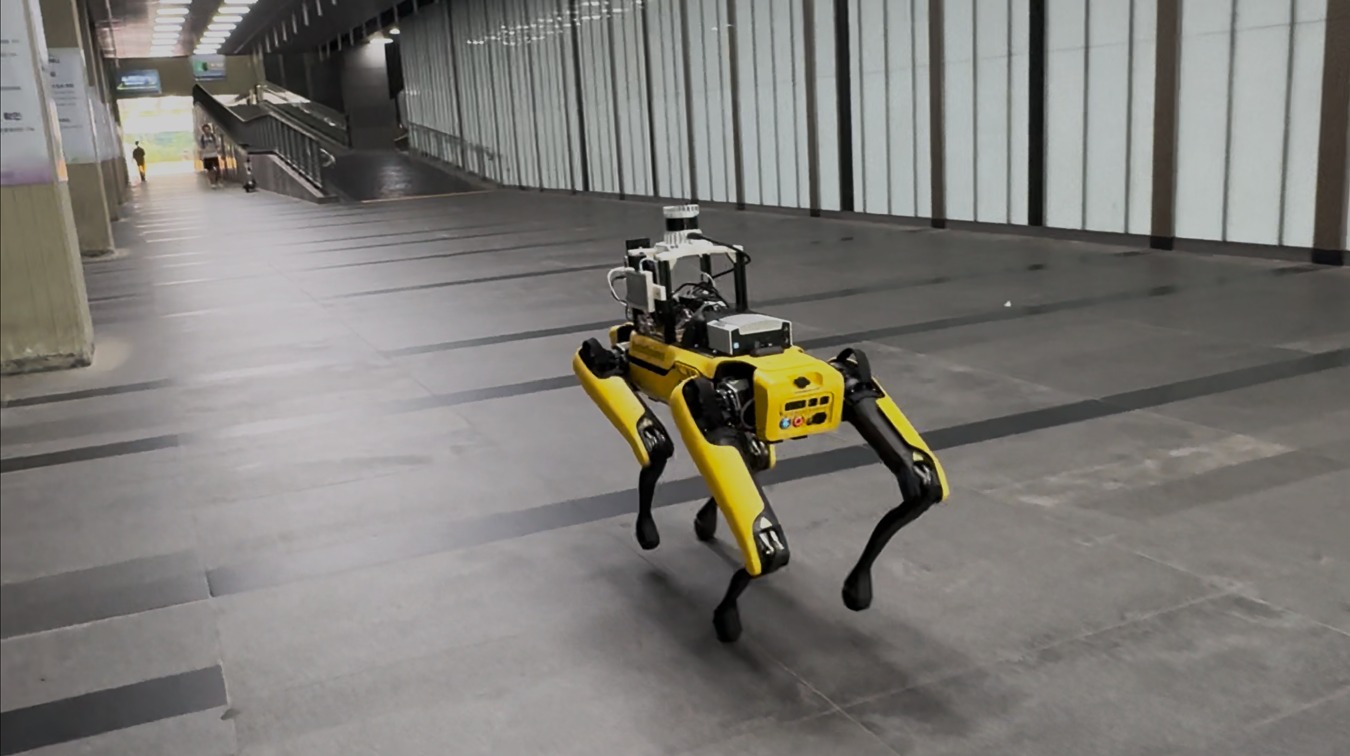}
        \label{fig:env_tunnel}
    }%
    \subfigure[\texttt{MoCap}]{
        \includegraphics[width=0.19\textwidth]{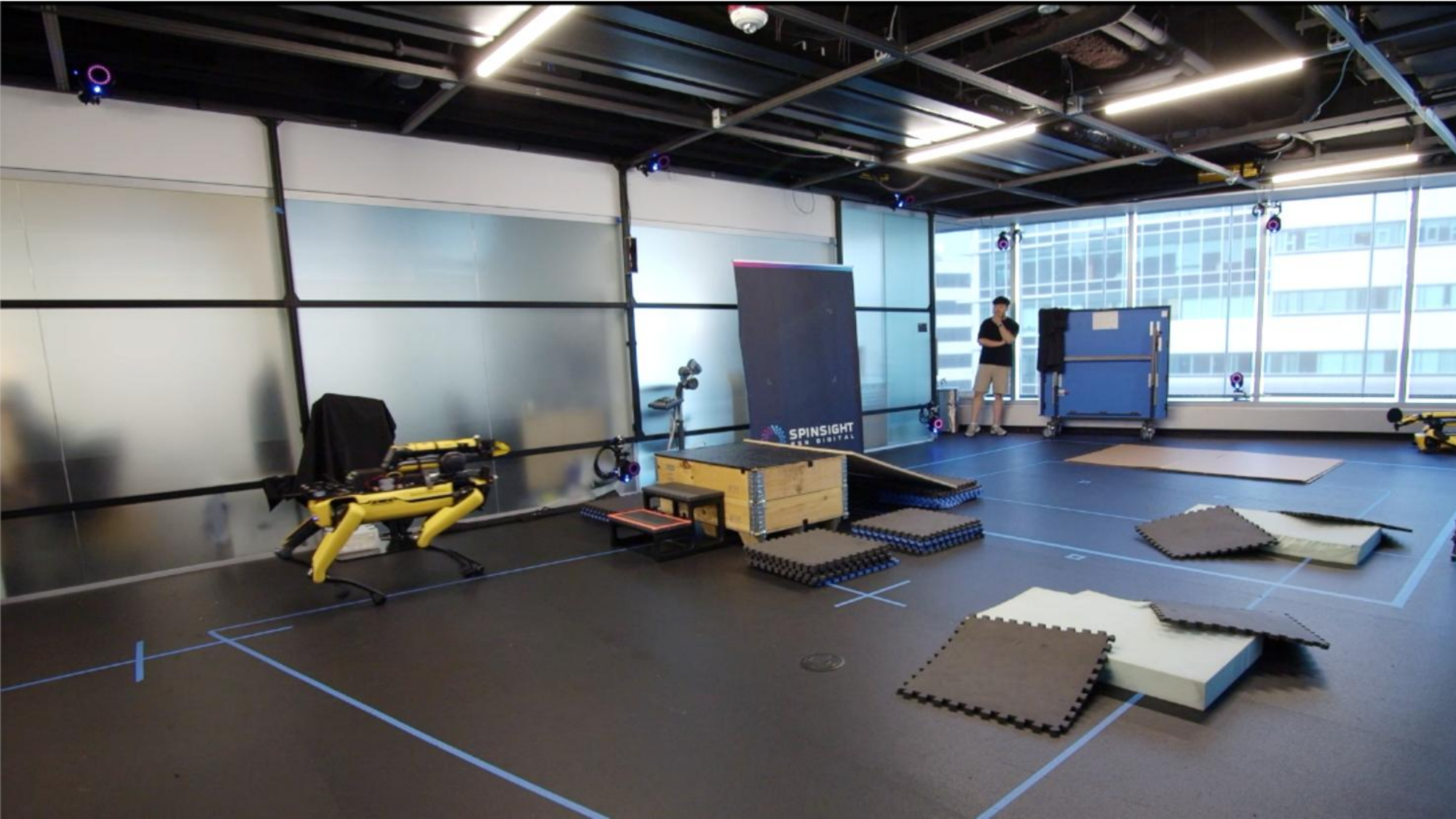}
        \label{fig:env_mocap}
    }    
    \caption{Environmental examples of the acquired sequences. Diverse environments are included in each sequence to consider various situations that the quadrupedal robot may encounter in a real-world mission. }
    \label{fig:sequence_env}
\end{figure*}

\section{Radar-Leg-IMU Dataset}

\subsection{System Configuration}
\label{sec:data.system}

The overall sensor configuration and the coordinate frames of each sensor are illustrated in \figref{fig:cali}, while detailed specifications for each sensor are provided in \tabref{tab:sensor}. Two different data acquisition setups are employed in this work, both built on Spot, a quadrupedal robot from Boston Dynamics, which provides joint encoder and contact sensor data at \unit{150}{Hz}. An IWR1843BOOST mmWave radar module captures 4D radar point clouds with a maximum range of \unit{11}{m} and a range resolution of \unit{4.8}{cm}, while a 3DM-GV7-AHRS IMU from Microstrain operates at \unit{100}{Hz} to provide inertial measurements including onboard AHRS orientation measurement. Both systems share the same radar and IMU sensor models.

\paragraph{SNU System}
The SNU system, used for experiments at Seoul National University, collects data in diverse conditions, including both indoor and outdoor environments. To obtain a baseline trajectory, an OS1-32 LiDAR (maximum range \unit{150}{m}) is mounted, and a \ac{TLS}-generated map is used as ground-truth reference (see Section~\secref{sec:GT_Traj}). Data acquisition and logging are performed onboard using the Spot CORE, equipped with an Intel 8th Gen i5 processor and 16 GB of DDR4 RAM.

\paragraph{RAI System}
The RAI system, used for experiments at the Robotics and AI Institute, operates entirely within an indoor motion-capture environment, providing a controlled experimental setup. While sharing the radar and IMU configurations of the SNU system, the \ac{LiDAR} is omitted, and ground truth is obtained from the motion capture system. Data is acquired and logged onboard with an NVIDIA Jetson AGX Orin, including a 12-core Arm Cortex-v8.2 CPU and 64 GB of LPDDR5 RAM.

\subsection{Details of Each Sequence}

An overview of the twelve sequences is given in \tabref{tab:sequence}, and their environments are illustrated in \figref{fig:sequence_env}; details of each sequence are as follows:

\begin{itemize}
    \item Atrium: A large flat floor indoor atrium with floor-to-ceiling glass and a high ceiling, where the expansive open floor and long glass facades induce strong specular reflections, multipath, and low parallax on radar sensor. \vspace{-2mm}
    \item BridgeLoop: An indoor sequence with three repetitions that traverses pedestrian bridges and short stair segments around an open atrium, stressing robustness in wide, low-parallax spaces. \vspace{-2mm}
    \item CorriLoop: A narrow, rectangular indoor corridor traversed twice; long straight segments with repeating doors/walls and a glossy floor create perceptual aliasing, while four sharp 90° turns stress turn handling and loop consistency. \vspace{-2mm}
    \item BiCorridor: A two-level corridor running in the same building as \texttt{CorriLoop}. One loop on the first floor, then a stair ascent and a second loop on the adjacent floor in the reverse direction. The narrow rectangular hallways contain long, low-parallax segments with repeating doors/walls that induce geometric degeneracy and perceptual aliasing, while the stairs introduce vertical motion and contact disturbances. This sequence stresses robustness to aliasing, direction reversal, and floor-to-floor consistency in constrained indoor spaces. \vspace{-2mm}
    \item Downstair: A multi-floor indoor sequence that starts on the second level with a long rectangular loop, descends one flight to traverse an extended straight corridor, then descends again to finish with a smaller rectangular loop. Wide, glossy hallways and two prolonged downward staircases stress perception during extended descent, floor-to-floor transitions, and low-parallax segments.\vspace{-2mm}
    \item Upstair: An indoor ascent sequence in the same building as \texttt{BridgeLoop}, climbing three floors by alternately traversing pedestrian bridges and short stair flights. The sequence includes upward motion—yaw changes occur between the bridge and stair segments—stressing vertical translation, stair negotiation, and transitions across landings in a wide, low-parallax space.\vspace{-2mm}
    \item SlopeStair: A mixed indoor–outdoor traverse with a long downhill ramp, multiple upstair flights, and doorway/corridor transitions, stressing robustness to large elevation changes, lighting shifts, and abrupt structural/surface variations.\vspace{-2mm}
    \item Overpass: This sequence takes place on outdoor stairs and a pedestrian overpass, with lamps and reflective paving. It stresses robustness to open-air transitions and repeated elevation changes across steps, ramps, and long sidewalk segments.\vspace{-2mm}
    \item Tunnel: This sequence traverses a long, semi-open tunnel lined with glass façades and repetitive concrete pillars. The path includes gentle ramps and a tight U-turn, stressing robustness under feature-degenerate geometry and illumination changes.\vspace{-2mm}
    \item Quad: This sequence traverses an outdoor campus quad with broad paved plazas, tiled walkways, curbs, and stairs. It stresses robustness to feature-sparse open areas and repetitive textures while handling stair climbing, ramps, and outdoor slopes.\vspace{-2mm}
    \item MoCap-E: This sequence has two loops in the indoor \ac{MoCap} room, each including a short stair–slope vertical motion zone and a cushion zone with two soft cushions that disturb leg odometry. In the second loop, a folded box at the slippery zone is deliberately dragged, breaking the assumption of a stationary floor and causing a marked discrepancy between leg kinematics and other sensors.\vspace{-2mm}
    \item MoCap-H: This sequence shares the same layout as \texttt{MoCap-E} but is run at a higher speed. In the slippery zone, the robot even moves forward while the floor shifts backward due to dragging, further violating the stationary floor assumption and amplifying the leg kinematics disagreement.\vspace{-2mm}
\end{itemize}

\subsection{Extrinsic Calibration of Sensor Systems}

Extrinsic calibration between Spot, radar, and \ac{IMU} is required for accurate odometry estimation. On both sensor systems, we leveraged the CAD model of each system to acquire the exact extrinsic parameters between the sensors. As included in \figref{fig:cali}, the radar attached to the RAI system is slightly biased to the right side of the robot, and the \ac{IMU} is attached perpendicularly compared with the SNU system. For more detailed information about the extrinsic calibration parameter, please refer to the project homepage.

\begin{figure}[!t]
    \centering
    \subfigure[Indoor Sequences. Left: \texttt{BiCorridor}, Right: \texttt{Upstair}]{
        \includegraphics[trim= 1cm 0cm 1cm 10cm, clip, width=0.95\columnwidth]{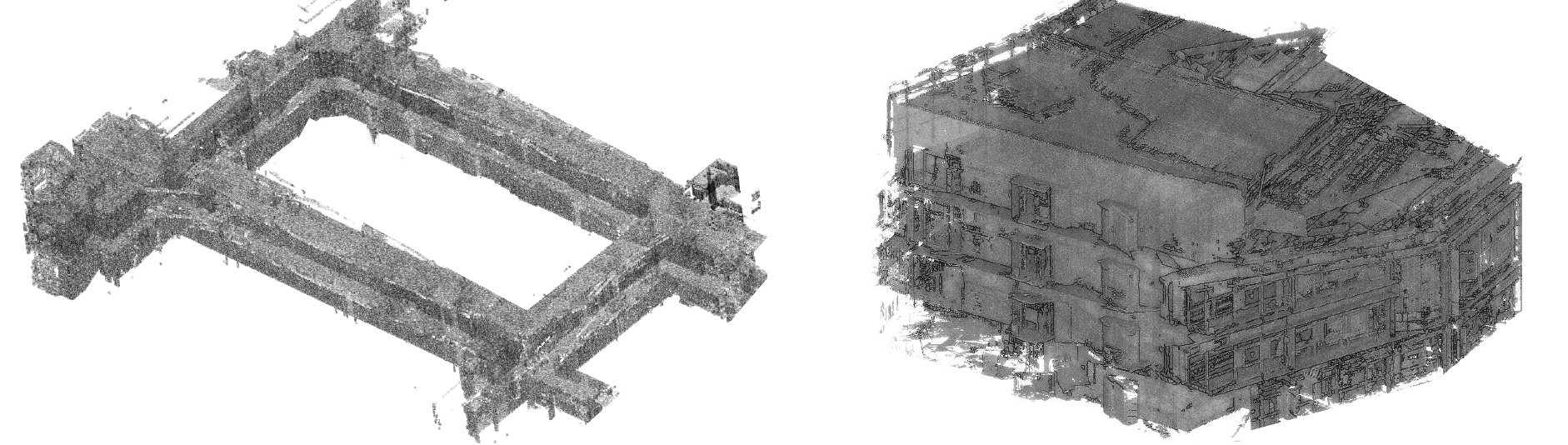}
        \label{fig:map1}
    }
    \subfigure[Outdoor Sequences. Left: \texttt{Quad}, Right: \texttt{Overpass}]{
        \includegraphics[trim= 1cm 10cm 1cm 0cm, clip, width=0.95\columnwidth]{sections/figs/map.pdf}
        \label{fig:map2}
    }  
    \caption{Examples of ground truth \ac{TLS} map on SNU sequences. Leveraged for generating ground truth trajectory. }
    \label{fig:map}
\end{figure}

\subsection{Ground Truth Trajectory Generation}
\label{sec:GT_Traj}
\subsubsection{SNU Sequences}
Accurate 6-\ac{DoF} ground truth poses are essential for evaluating various robotic tasks, including state estimation and \ac{SLAM}. Unlike previous studies, our deployments span both indoor and outdoor environments over several hundred meters, requiring millimeter-level precision. These stringent requirements render traditional ground truth sources, such as \ac{LiDAR}-\ac{IMU}-based references~\citep{jung2024co} and RTK-GNSS systems~\citep{geiger2013vision, barnes2020oxford, kim2025hercules}, unsuitable. While \ac{MoCap} systems can provide high-frequency and high-precision pose estimates, their limited workspace makes them impractical for large-scale deployments~\citep{doer2021yaw}. Some prior works~\citep{tranzatto2022cerberus} utilize survey-grade maps as ground truth references by performing scan-to-map matching of deskewed \ac{LiDAR} points using synchronized inertial and ranging sensors. However, for legged robots, continuous dynamic motion during data acquisition significantly degrades \ac{LiDAR} deskewing and motion estimation accuracy.

To address this, we adopt the approach proposed in~\citep{hu2024paloc}, which combines FAST-LIO2~\citep{xu2022fast} odometry and loop closure factors with a degeneration-aware map factor derived from dense prior maps. As illustrated in \figref{fig:map}, prior maps are collected using a Leica RTC360. This graph-based formulation enables accurate pose estimation even in degenerate and stationary conditions, thereby providing reliable ground truth for our evaluation.

Because this ground truth framework leverages FAST-LIO2~\citep{xu2022fast}, a tightly coupled \ac{LiDAR}-\ac{IMU} \ac{SLAM} as an odometry front-end, precise extrinsic calibration between those two sensors $\mathbf{T}^L_I$ is essential. We perform this extrinsic calibration using the robust method of \citet{zhu2022robust}, which applies diverse rotational motions across all three axes to accumulate sufficient motion data, allowing rapid and accurate estimation without initial parameter guesses. The method directly computes spatial and temporal offsets from unsynchronized data, yielding high-precision \ac{LiDAR}-\ac{IMU} extrinsics required for our ground-truth estimation.

\subsubsection{RAI Sequences}
For \ac{MoCap} sequences acquired using the RAI sensor system, we utilized the \ac{MoCap} system to obtain a highly precise ground truth trajectory. As these sequences were collected exclusively in a controlled indoor environment, the \ac{MoCap} odometry could be reliably used as reference. The experiment was conducted in a \unit{13.5}{m}$~\times~$\unit{5}{m}$~\times~$\unit{3}{m} motion-capture room equipped with 20 Vicon Valkyrie VK 16 cameras, ensuring complete coverage of the space. The system streamed motion capture data at \unit{120}{Hz} using Vicon Tracker 4.3 software.

\section{Experiment Results}

\begin{table*}[tbp]
\small\sf\centering
\caption{Evaluation on radar-based methods. \textbf{Bold} numbers indicate the smallest error, and \underline{underlined} numbers denote the second smallest error within each metric.}
\label{tab:radar_indoor}

\resizebox{\textwidth}{!}{
\begin{tabular}{l|c|r|rrrr?{0.5mm}l|c|r|rrrr}
\toprule
 \hline
\multicolumn{1}{l|}{\multirow{2}{*}{Sequence}} & \multicolumn{1}{c|}{\multirow{2}{*}{RMSE}} & \multicolumn{1}{c|}{\multirow{2}{*}{\textbf{Ours}}} & \multicolumn{4}{c?{0.5mm}}{\textbf{Radar based method}} &
\multicolumn{1}{l|}{\multirow{2}{*}{Sequence}} & \multicolumn{1}{c|}{\multirow{2}{*}{RMSE}} & \multicolumn{1}{c|}{\multirow{2}{*}{\textbf{Ours}}} & \multicolumn{4}{c}{\textbf{Radar based method}} 
\\ \cline{4-7} \cline{11-14}
\multicolumn{1}{l|}{} & \multicolumn{1}{l|}{} & \multicolumn{1}{l|}{} & \multicolumn{1}{c}{\textbf{River}} & \multicolumn{1}{c}{\textbf{Co-RaL}} & \multicolumn{1}{c}{\textbf{EKF-RIO}} & \multicolumn{1}{c?{0.5mm}}{\textbf{DeRO}} 
&
\multicolumn{1}{l|}{} & \multicolumn{1}{l|}{} & \multicolumn{1}{l|}{} & \multicolumn{1}{c}{\textbf{River}} & \multicolumn{1}{c}{\textbf{Co-RaL}} & \multicolumn{1}{c}{\textbf{EKF-RIO}} & \multicolumn{1}{c}{\textbf{DeRO}}
\\ \hline

\multirow{2}{*}{Atrium} & APE$_t$ & \cellcolor[HTML]{def3e6}\textbf{0.816} & 15.473 & \underline{2.606} & 6.679 & 15.127 &  \multirow{2}{*}{SlopeStair} & APE$_t$ & \cellcolor[HTML]{def3e6}\textbf{2.359} & 38.438 & \underline{8.222} & - & - \\
 & APE$_r$ & \cellcolor[HTML]{def3e6}\textbf{1.715} & 25.154 & \underline{4.393} & 12.496 & 9.602 &  & APE$_r$ & \cellcolor[HTML]{def3e6}\textbf{2.805} & 49.587 & \underline{9.954} & - & - \\ 
 & RPE$_t$ & \cellcolor[HTML]{def3e6}\textbf{0.055} & 0.213 & \underline{0.132} & 0.162 & 0.203 &  & RPE$_t$ & \cellcolor[HTML]{def3e6}\textbf{0.061} & 0.248 & \underline{0.203} & - & - \\ 
 & RPE$_r$ & \cellcolor[HTML]{def3e6}\textbf{0.554} & 0.942 & {0.780} & \underline{0.722} & 1.572 &  & RPE$_r$ & 1.051 & \underline{1.049} & \textbf{0.920} & - & -\\ 
 & APE$_z$ & \cellcolor[HTML]{def3e6}\textbf{0.106} & 12.508 & \underline{2.225} & 4.908 & 15.072 &  & APE$_z$ & \cellcolor[HTML]{def3e6}\textbf{0.665} & 24.701 & \underline{2.789} & - & - \\ 
 \hline

\multirow{2}{*}{BridgeLoop} & APE$_t$ & \cellcolor[HTML]{def3e6}\textbf{1.193} & 16.930 & \underline{4.562} & 7.125 & 19.793 & \multirow{2}{*}{Overpass} & APE$_t$ & \cellcolor[HTML]{def3e6}\textbf{1.526} & 11.210 & \underline{8.486} & - & - \\ 
 & APE$_r$ & \cellcolor[HTML]{def3e6}\textbf{2.719} & 108.904 & \underline{7.959} & 30.850 & 14.468 &  & APE$_r$ & \cellcolor[HTML]{def3e6}\textbf{4.043} & 25.574 & \underline{17.416} & - & - \\ 
 & RPE$_t$ & \cellcolor[HTML]{def3e6}\textbf{0.080} & 0.183 & 0.149 & \underline{0.132} & 0.169 &  & RPE$_t$ & \cellcolor[HTML]{def3e6}\textbf{0.091} & 0.281 & \underline{0.230} & - & - \\ 
 & RPE$_r$ & 1.058 & 2.159 & \underline{1.022} & \textbf{0.936} & 1.694 &  & RPE$_r$ & 1.227 & \underline{1.136} & \textbf{0.952} & - & - \\ 
 & APE$_z$ & \cellcolor[HTML]{def3e6}\textbf{0.117} & 13.422 & \underline{4.131} & 5.302 & 19.728 &  & APE$_z$ & \cellcolor[HTML]{def3e6}\textbf{1.074} & 8.034 & \underline{7.728} & - & - \\ 
 \hline
 
\multirow{2}{*}{CorriLoop} & APE$_t$ & \cellcolor[HTML]{def3e6}\textbf{1.627} & 28.885 & \underline{6.576} & 9.283 & 18.049 & \multirow{2}{*}{Tunnel} & APE$_t$ & \cellcolor[HTML]{def3e6}\textbf{3.523} & 29.515 & \underline{7.939} & 26.800 & 40.231 \\ 
 & APE$_r$ & \cellcolor[HTML]{def3e6}\textbf{5.676} & 45.570 & \underline{6.245} & 28.555 & 15.863 &  & APE$_r$ & \cellcolor[HTML]{def3e6}\textbf{2.849} & 31.844 & \underline{8.362} & 33.427 & 31.507 \\
 & RPE$_t$ & \cellcolor[HTML]{def3e6}\textbf{0.066} & 0.245 & \underline{0.129} & 0.180 & 0.178 &  & RPE$_t$ & \cellcolor[HTML]{def3e6}\textbf{0.083} & \underline{0.196} & 0.262 & 0.316 & \underline{0.196} \\ 
 & RPE$_r$ & \cellcolor[HTML]{def3e6}\textbf{0.738} & 1.002 & 0.944 & \underline{0.884} & {0.902} &  & RPE$_r$ & \cellcolor[HTML]{def3e6}\textbf{0.440} & 0.752 & {0.638} & \underline{0.541} & 0.683  \\ 
 & APE$_z$ & \cellcolor[HTML]{def3e6}\textbf{0.168} & 25.794 & 5.846 & \underline{4.001} & 17.595 &  & APE$_z$ & \cellcolor[HTML]{def3e6}\textbf{0.548} & 18.991 & \underline{2.872} & 6.139 & 36.107 \\ 
 \hline
 
\multirow{2}{*}{BiCorridor} & APE$_t$ & \cellcolor[HTML]{def3e6}\textbf{1.425} & 55.161 & \underline{7.631} & 10.293 & 23.500 &  \multirow{2}{*}{Quad} & APE$_t$ & \cellcolor[HTML]{def3e6}\textbf{7.347} & - & \underline{14.395} & - & - \\ 
 & APE$_r$ & \cellcolor[HTML]{def3e6}\textbf{5.519} & 97.208 & \underline{7.232} & 27.697 & 13.141 &  & APE$_r$ & \cellcolor[HTML]{def3e6}\textbf{3.356} & - & \underline{11.615} & - & - \\ 
 & RPE$_t$ & \cellcolor[HTML]{def3e6}\textbf{0.063} & 0.252 & \underline{0.114} & 0.185 & 0.220 &  & RPE$_t$ & \cellcolor[HTML]{def3e6}\textbf{0.080} & - & \underline{0.298} & - & - \\
 & RPE$_r$ & 0.885 & 2.371 & \textbf{0.799} & \underline{0.838} & 0.840 &  & RPE$_r$ & \underline{0.838} & - & \textbf{0.532} & - & - \\ 
 & APE$_z$ & \cellcolor[HTML]{def3e6}\textbf{0.226} & 30.646 & 7.087 & \underline{5.497} & 23.003 &  & APE$_z$ & \cellcolor[HTML]{def3e6}\textbf{2.143} & - & \underline{3.478} & - & - \\ 
 \hline

\multirow{2}{*}{Downstair} & APE$_t$ & \cellcolor[HTML]{def3e6}\textbf{3.916} & 66.838 & \underline{8.250} & 30.835 & 29.710 & \multirow{2}{*}{MoCap-E} & APE$_t$ & \cellcolor[HTML]{def3e6}\textbf{0.852} & \underline{0.916} & - & 2.156 & 3.305 \\ 
 & APE$_r$ & \cellcolor[HTML]{def3e6}\textbf{3.415} & 111.422 & \underline{8.029} & 36.447 & 30.487 &  & APE$_r$ & \cellcolor[HTML]{def3e6}\textbf{3.373} & \underline{5.738} & - & 6.146 & 33.163 \\ 
 & RPE$_t$ & \cellcolor[HTML]{def3e6}\textbf{0.099} & 0.180 & \underline{0.111} & 0.229 & 0.166 &  & RPE$_t$ & \cellcolor[HTML]{def3e6}\textbf{0.184} & \underline{0.239} & - & 0.328 & 0.373 \\ 
 & RPE$_r$ & 1.080 & 1.304 & \underline{0.940} & \textbf{0.875} & 1.619 &  & RPE$_r$ & \cellcolor[HTML]{def3e6}\textbf{1.508} & \underline{1.922} & - & {2.492} & 11.822 \\ 
 & APE$_z$ & \cellcolor[HTML]{def3e6}\textbf{0.530} & 14.934 & 6.096 & \underline{4.251} & 25.466 &  & APE$_z$ & \cellcolor[HTML]{def3e6}\textbf{0.103} & \underline{0.236} & - & 1.776 & 2.555 \\ 
 \hline

\multirow{2}{*}{Upstair} & APE$_t$ & \cellcolor[HTML]{def3e6}\textbf{1.496} & 20.198 & \underline{7.887} & 10.044 & 31.236 & \multirow{2}{*}{MoCap-H} & APE$_t$ & \cellcolor[HTML]{def3e6}\textbf{0.880}  & \underline{1.145} & 2.889 & 2.121 & 6.626  \\ 
 & APE$_r$ & \cellcolor[HTML]{def3e6}\textbf{4.048} & 14.445 & \underline{7.626} & 28.159 & 13.299 &  & APE$_r$ & \underline{1.981} & 8.814 & \textbf{1.501} & 7.416 & 66.105 \\
 & RPE$_t$ & \cellcolor[HTML]{def3e6}\textbf{0.071} & 0.198 & \underline{0.143} & {0.143} & 0.195 &  & RPE$_t$ &  \cellcolor[HTML]{def3e6}\textbf{0.204} & \underline{0.250} & 0.318 & 0.325 & 0.513 \\
 & RPE$_r$ & 0.933 & \underline{0.906} & 1.015 & \textbf{0.808} & 2.119 &  & RPE$_r$ & \underline{1.665}  & 2.219 & \textbf{0.686} & {1.995} & 27.176 \\ 
 & APE$_z$ & \cellcolor[HTML]{def3e6}\textbf{0.514} & 19.964 & \underline{7.317} & 8.871 & 31.169 &  & APE$_z$ & \cellcolor[HTML]{def3e6}\textbf{0.118} & \underline{0.450} & 2.404 & 1.646 & 4.782 \\ 
 \hline

\bottomrule
\end{tabular}
}

\end{table*}


In this section, we evaluate the performance of GaRLILEO against \ac{SOTA} odometry algorithms that utilize \ac{SoC} radar, \ac{IMU}, and leg kinematics. The experiments are conducted on a self-collected real-world dataset. 

Odometry accuracy is assessed using the \ac{RMSE} of \ac{APE} and \ac{RPE}, each decomposed into translational and rotational components. The units are as follows: \ac{APE}$_t$ (\meter), \ac{APE}$_r$ (\degree), \ac{RPE}$_t$ (\meter/\meter), and \ac{RPE}$_r$ (\degree/\meter). To specifically evaluate vertical drift in odometry, we report the z-axis \ac{APE} (\ac{APE}$_z$). All evaluations are performed using the Evo Trajectory Evaluator~\citep{grupp2017evo}, a widely adopted open-source toolkit for odometry benchmarking in robotics.

The comparative analysis is organized by the baselines' sensor configurations, while GaRLILEO is evaluated in its full configuration to report system-level performance, i.e., the benefit of fusing leg kinematics, radar, and IMU. First, we compare GaRLILEO against recent \ac{SoC} radar-\ac{IMU} odometry methods, including Co-RaL~\citep{jung2024co}, which additionally integrates the leg kinematics velocity factor. Next, we also evaluate GaRLILEO against open-source leg kinematics-\ac{IMU} fusion odometry methods. Every parameter is adopted from the official implementation, except that the robot-specific parameters of legged robots are modified based on the official \ac{URDF} file of Boston Dynamics SPOT. Detailed descriptions of the baseline algorithms and comprehensive evaluation results are provided in the following subsections.

After comparing GaRLILEO directly with the baselines, we conducted detailed ablation studies on the modules of GaRLILEO. Specifically, we analyzed the complementary effect between radar and leg kinematics, the contribution of gravity factors to both local gravity vector estimation and odometry, and the impact of the velocity bias term.

\subsection{Radar-IMU Odometry Comparison}
\label{sec:result.radar}

In this subsection, we compare the performance of GaRLILEO with that of five recent \ac{SoC} \ac{RIO} methods. The baseline methods are listed as follows:

\begin{itemize}
\item \textbf{River}~\citep{chen2024river}: A B-spline-based continuous velocity estimator that fuses \ac{SoC} radar and \ac{IMU}, employing dead reckoning to compute full odometry.
\item \textbf{Co-RaL}~\citep{jung2024co}: A cooperative odometry algorithm integrating \ac{SoC} radar, \ac{IMU}, and leg kinematics velocity, designed to operate robustly across diverse environments.
\item \textbf{EKF-RIO}~\citep{doer2020ekf}: An \ac{EKF}-based odometry method that fuses \ac{SoC} radar and \ac{IMU} data.
\item \textbf{DeRO}~\citep{do2024dero}: A dead reckoning based \ac{RIO} method that combines \ac{SoC} radar ego-velocity and gyroscope data using an \ac{IEKF}, with accelerometer-based tilt angle estimation.
\end{itemize}

\begin{figure*}[!t]
    \centering
    \includegraphics[width=\textwidth]{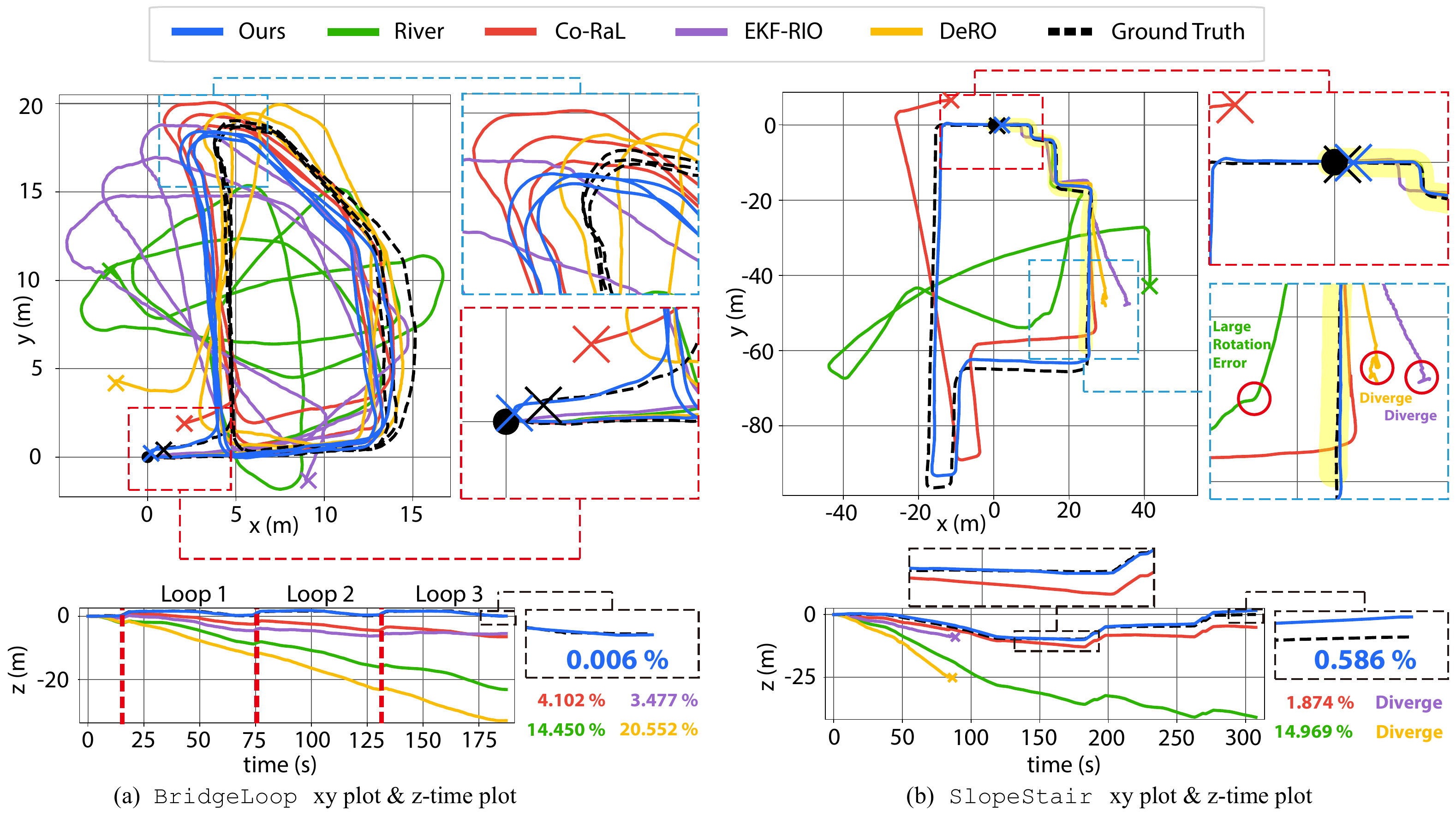}
    \caption{Radar–Inertial baseline odometry results on the (a) \texttt{BridgeLoop} and (b) \texttt{SlopeStair} sequences. Red dotted lines in (a, bottom) indicate the starting points of each loop in \texttt{BridgeLoop} sequence, while the faint yellow region in (b, top) denotes the outdoor segment of the \texttt{SlopeStair} sequence. The red zoomed-in views highlight that GaRLILEO and Co-RaL—both integrating SoC radar and leg kinematics—converge more closely to the ground-truth final position. The colored numbers on the right-hand side of the bottom plots show relative start-to-end vertical drift with respect to the full path length, where GaRLILEO achieves substantially lower drift. The cyan zoomed view in (a) illustrates that GaRLILEO (blue) follows the most consistent turning trajectory across repeated loops. In (b), the red circles indicate the failure points of baseline odometries during outdoor-to-indoor transitions, whereas radar–leg fused odometries, GaRLILEO, and Co-RaL maintain robust estimation. Both subfigures qualitatively illustrate the odometry accuracy of GaRLILEO, particularly in suppressing vertical drift. 
    }
    \label{fig:radar_odom}
\end{figure*}

We evaluate the performance of GaRLILEO and baseline methods, with a focus on odometry accuracy in diverse environments. \tabref{tab:radar_indoor} presents quantitative results for \ac{APE} and \ac{RPE} metrics. Below, we elaborate on the results for each sequence. Some sample trajectories and qualitative evaluation are illustrated in \figref{fig:radar_odom}. For both quantitative and qualitative analysis, GaRLILEO achieves superior odometry accuracy across most metrics, particularly excelling in vertical accuracy (\ac{APE}$_z$).

\noindent\textit{1) Atrium:}
We first evaluate methods on the \texttt{Atrium} sequence, a flat environment with a single loop, minimal vertical motion, where contact-induced drift is negligible. River and DeRO, which rely on dead reckoning, show large \ac{APE}$_t$ and \ac{APE}$_z$ due to insufficient drift mitigation from the high vibration of the legged robot \ac{UGV}. EKF-RIO achieves improved accuracy on most metrics by tightly fusing \ac{SoC} radar–derived ego-velocity with \ac{IMU} measurements via an \ac{EKF}-based approach. However, it exhibits a higher \ac{APE}$_r$ than DeRO. Co-RaL, which integrates the leg kinematics velocity preintegration factor, achieves the second-lowest \ac{APE}$_z$ among the methods. GaRLILEO, leveraging accurate local gravity estimation, achieves an \ac{APE}$_z$ only about 5\% of Co-RaL’s while also notably reducing errors in every other metric, making a big gap with baselines. 

\noindent\textit{2) BridgeLoop:}
We next assess the \texttt{BridgeLoop}, where the robot completes triple loops in a consistent rotational direction while going over gentle slope bridges and short stairs multiple times. In \texttt{BridgeLoop}, noise in roll and pitch estimation from contact uncertainty leads to substantial vertical drift, as reflected in the elevated \ac{APE}$_z$ of River and DeRO. EKF-RIO suffers from significantly high \ac{APE}$_r$ due to the limited orientation observability of \ac{IMU}-only fusion, which becomes more severe in multi-loop trajectories. In contrast, Co-RaL achieves the second lowest \ac{APE}$_r$ via the 4-\ac{DoF} radar factor, enhancing both translational and rotational accuracy. GaRLILEO provides the most robust pose estimates, presenting the lowest errors in every metric except \ac{RPE}$_r$, while \ac{RPE}$_r$ is at a similar level to the second-best baseline, Co-RaL. Notably, the vertical drift \ac{APE}$_z$ is almost identical to the flat sequence \texttt{Atrium}, expressing the vertical robustness of GaRLILEO on multiple turns. 

\noindent\textit{3) CorriLoop:}
The \texttt{CorriLoop} sequence, consisting of two loops on flat indoor terrain, shows similar trends. While River and EKF-RIO suffer from significant rotational errors, GaRLILEO effectively suppresses them through accurate roll and pitch estimation. Although Co-RaL achieves comparable \ac{APE}$_r$ due to the radar factor, their \ac{APE}$_z$ remains about 35$\times$ higher than GaRLILEO’s, reflecting limited roll and pitch observability.

\noindent\textit{4) BiCorridor:}
The \texttt{BiCorridor} sequence combines a loop with stair ascents, traversing two floors with opposite rotational directions. Despite environmental similarities to \texttt{CorriLoop}, the sharp rotations and stair climbs substantially degrade odometry accuracy, particularly along the vertical axis. This results in high \ac{APE}$_t$ and \ac{APE}$_z$ for simple dead-reckoning methods, such as River and DeRO. Interestingly, EKF-RIO achieves lower \ac{APE}$_t$ than Co-RaL, but suffers from much higher \ac{APE}$_r$. In contrast, GaRLILEO effectively addresses these challenges by continuously fusing leg kinematics, \ac{IMU}, and \ac{SoC} radar, maintaining sub-meter \ac{APE}$_z$.

\noindent\textit{5) Downstair:}
In the \texttt{Downstair} sequence, only GaRLILEO and Co-RaL converge successfully, while River, EKF-RIO, and DeRO return erroneous odometry results due to noisy \ac{IMU} measurements caused by rapid and repetitive impacts during stair descent. Among the convergent methods, GaRLILEO and Co-RaL achieve the lowest \ac{APE}$_t$, demonstrating stable and precise estimation. With its local gravity factor, GaRLILEO achieves sub-meter \ac{APE}$_z$, which is lower than 10\% of the second best result from Co-RaL, significantly outperforming all prior radar–\ac{IMU} methods.

\noindent\textit{6) Upstair:}
The \texttt{Upstair} sequence, which involves multiple stair ascents and a sloped bridge in an open space, exhibits similar trends. Methods without leg kinematics input, such as River and EKF-RIO, suffer from pronounced vertical drift due to the lack of leg kinematics constraints. Because this sequence consists of the same directional looped trajectories with stair climbing in each loop, EKF-RIO behaves similarly to the \texttt{BridgeLoop} and \texttt{CorriLoop} sequences. The second-best result, Co-RaL, shows vertical errors nearly 20$\times$ larger than GaRLILEO, underscoring the effectiveness of our continuous-time local gravity estimation for precise roll and pitch correction.

\noindent\textit{7) SlopeStair:}
As the system moves from indoors to outdoors and back in during this sequence, severe drift at the corner before re-entry causes EKF-RIO and DeRO to diverge. This highlights the limitations of \ac{SoC} radar–\ac{IMU}-only systems during indoor/outdoor transitions. Although River converges, their vertical errors (\ac{APE}$_z$) remain significantly higher than those of methods incorporating leg kinematics, highlighting the importance of kinematics sensing for stable and accurate odometry. Co-RaL, fusing leg kinematics with radar–\ac{IMU} data, achieves the second-best performance across most metrics, demonstrating the benefits of multimodal integration. 
Finally, GaRLILEO delivers the most accurate odometry overall, achieving sub-meter-level vertical accuracy.

\noindent\textit{8) Overpass:}
In the \texttt{Overpass} sequence, staircases appear before and after the overpass; when the robot is on the stairs, fewer radar returns are observed because the sensor is angled skyward. Similar to the \texttt{SlopeStair} sequence, EKF-RIO and DeRO fail to converge, underscoring their limitations in handling sharp turns in open environments. River achieves reduced vertical drift (\ac{APE}$_z$) compared to the \texttt{SlopeStair} sequence. This improvement, however, is largely attributed to the shorter trajectory and smaller elevation change of the \texttt{Overpass} sequence itself. Co-RaL demonstrates slightly better accuracy than River due to the additional fusion of leg kinematics, though the improvement is marginal. In contrast, GaRLILEO robustly manages outdoor contact-induced drift, reducing vertical error to lower than 15\% and achieving substantial improvement in overall \ac{APE} compared to all baselines.

\noindent\textit{9) Tunnel:}
In the \texttt{Tunnel} sequence, which features a long passage with a sharp U-turn, GaRLILEO maintains sub-meter \ac{APE}$_z$ and achieves the lowest error on every metric. Co-RaL, benefiting from radar–leg kinematics fusion, ranks second. River and DeRO exhibit severe vertical drift due to contact impacts and limited observability. EKF-RIO shows less drift than these methods, but still larger vertical errors than leg kinematics-fused methods.

\noindent\textit{10) Quad:}
The \texttt{Quad} sequence, the longest and most challenging dataset, includes stairs, slopes, and substantial elevation changes. Frequent contact drift, multiple dynamic objects, outdoor stairs and slopes, and a reduced number of radar points make accurate pose estimation particularly difficult. River, EKF-RIO, and DeRO fail to converge due to sparse radar returns. Due to its factor graph framework that adaptively relies on the more reliable sensor modality, Co-RaL succeeds in converging and even outperforms GaRLILEO on \ac{RPE}$_r$. Nevertheless, GaRLILEO achieves a notably lower error on all other metrics through precise local gravity estimation, while exhibiting slightly higher but competitive \ac{RPE}$_r$ compared to Co-RaL. These results highlight that GaRLILEO consistently provides robust and accurate pose estimation across diverse environments, owing to its local gravity model.

\noindent\textit{11) MoCap-E:}
Both \texttt{MoCap} sequences include two challenging test scenarios: (\textit{i}) a slippery zone with a backward-moving floor in the second loop and (\textit{ii}) a cushion zone included in both loops. These environments are designed to degrade leg odometry. In the slippery zone, as the contact frame continuously moves, leg kinematics produce erroneous horizontal velocity estimates. In the cushion zone, where the contact frame moves downward while the contact sensor remains active, leg kinematics yield incorrect vertical velocity estimates.

\begin{table*}[tbp]
\small\sf\centering
\caption{Evaluation on Leg Kinematic based methods}
\label{tab:leg_indoor}

\resizebox{\textwidth}{!}{
\begin{tabular}{l|c|r|rrrr?{0.5mm}l|c|r|rrrr}
\toprule
 \hline
\multicolumn{1}{l|}{\multirow{2}{*}{Sequence}} & \multicolumn{1}{c|}{\multirow{2}{*}{RMSE}} & \multicolumn{1}{c|}{\multirow{2}{*}{\textbf{Ours}}} & \multicolumn{4}{c?{0.5mm}}{\textbf{Leg Kinematic based method}} & \multicolumn{1}{l|}{\multirow{2}{*}{Sequence}} & \multicolumn{1}{c|}{\multirow{2}{*}{RMSE}} & \multicolumn{1}{c|}{\multirow{2}{*}{\textbf{Ours}}} & \multicolumn{4}{c}{\textbf{Leg Kinematic based method}} \\ \cline{4-7} \cline{11-14}
\multicolumn{1}{l|}{} & \multicolumn{1}{l|}{} & \multicolumn{1}{l|}{} & \multicolumn{1}{c}{\textbf{Pronto}} & \multicolumn{1}{c}{\textbf{MUSE}} & \multicolumn{1}{c}{\textbf{DRIFT}} & \multicolumn{1}{c?{0.5mm}}{\textbf{Holistic}} & \multicolumn{1}{l|}{} & \multicolumn{1}{l|}{} & \multicolumn{1}{l|}{} & \multicolumn{1}{c}{\textbf{Pronto}} & \multicolumn{1}{c}{\textbf{MUSE}} & \multicolumn{1}{c}{\textbf{DRIFT}} & \multicolumn{1}{c}{\textbf{Holistic}}  \\ \hline

\multirow{2}{*}{Atrium} & APE$_t$ & \cellcolor[HTML]{def3e6}\textbf{0.816} & 7.849 & \underline{3.297} & 7.228 & 7.407 & \multirow{2}{*}{SlopeStair} & APE$_t$ & \cellcolor[HTML]{def3e6}\textbf{2.359} & - & 35.519 & \underline{20.004} & 30.420  \\ 
 & APE$_r$ & \cellcolor[HTML]{def3e6}\textbf{1.715} & 21.448 & 13.597 & \underline{5.225} & 17.631 &  & APE$_r$ & \cellcolor[HTML]{def3e6}\textbf{2.805} & - & 54.144 & \underline{12.821} & 40.790 \\ 
 & RPE$_t$ & \cellcolor[HTML]{def3e6}\textbf{0.055} & 0.293 & \underline{0.090} &  0.153 & 0.151 &  & RPE$_t$ & \cellcolor[HTML]{def3e6}\textbf{0.061} & - & \underline{0.151} & 0.180 & 0.205  \\ 
 & RPE$_r$ & \cellcolor[HTML]{def3e6}\textbf{0.554} & \underline{0.700} & 0.925 & {0.846} & 1.209 &  & RPE$_r$ & \cellcolor[HTML]{def3e6}\textbf{1.051}  & - & \underline{1.063} & 1.064 & 1.274  \\ 
 & APE$_z$ & \cellcolor[HTML]{def3e6}\textbf{0.106} & \underline{0.592} & 0.685 & 6.460 & 5.303 &  & APE$_z$ & \cellcolor[HTML]{def3e6}\textbf{0.665} & - & \underline{0.967} & 15.480 & 8.427  \\ 
 \hline

\multirow{2}{*}{BridgeLoop} & APE$_t$ & \cellcolor[HTML]{def3e6}\textbf{1.193}  & 6.684 & \underline{4.195} & 9.350 & 5.037 & \multirow{2}{*}{Overpass} & APE$_t$ & \cellcolor[HTML]{def3e6}\textbf{1.526} & \underline{4.039} &  6.507 & 9.785 & 5.920 \\ 
 & APE$_r$ & \cellcolor[HTML]{def3e6}\textbf{2.719} & 16.739 & 26.513 & \underline{5.175} & 25.208 &  & APE$_r$ & \cellcolor[HTML]{def3e6}\textbf{4.043} & \underline{5.367} & 10.373 &  8.268 &  16.034 \\ 
 & RPE$_t$ & \cellcolor[HTML]{def3e6}\textbf{0.080} & 0.256 &  0.200 & 0.212 & \underline{0.162} &  & RPE$_t$ & \cellcolor[HTML]{def3e6}\textbf{0.091} & 0.139 & 0.191 & 0.173 & \underline{0.125}  \\  
 & RPE$_r$ & 1.058  & \textbf{0.950} & \underline{0.967} & 1.180 & 1.823  &  & RPE$_r$ & 1.227 & \textbf{0.582} &  \underline{0.972} & 1.126 & 1.594  \\ 
 & APE$_z$ & \cellcolor[HTML]{def3e6}\textbf{0.117} & 5.160 & \underline{2.250} & 9.206 & 3.870 &  & APE$_z$ & \cellcolor[HTML]{def3e6}\textbf{1.074} & \underline{1.227} & 2.442 & 8.326 & 2.424  \\ 
 \hline

\multirow{2}{*}{CorriLoop} & APE$_t$ & \cellcolor[HTML]{def3e6}\textbf{1.627} & 8.988 & \underline{5.811} & 12.379 & 14.780 & \multirow{2}{*}{Tunnel} & APE$_t$ & \cellcolor[HTML]{def3e6}\textbf{3.523} & \underline{6.171} & 10.194 & 18.074 & 28.576 \\  
 & APE$_r$ & \cellcolor[HTML]{def3e6}\textbf{5.676} & 25.691 & 18.871 & \underline{10.887} & 25.942 &  & APE$_r$ & \cellcolor[HTML]{def3e6}\textbf{2.849} & 13.369 & 11.199 & \underline{7.302} & 41.034 \\ 
 & RPE$_t$ & \cellcolor[HTML]{def3e6}\textbf{0.066} & 0.200 & \underline{0.127} & 0.154 & 0.173 &  & RPE$_t$ & \cellcolor[HTML]{def3e6}\textbf{0.083} & 0.158 & \underline{0.105} & 0.159 & 0.199   \\  
 & RPE$_r$ & \underline{0.738} & \textbf{0.711} & 1.028 & {0.851} & 1.358 &  & RPE$_r$ & \cellcolor[HTML]{def3e6}\textbf{0.440} & \underline{0.482} & {0.500} & 0.744 & 0.903  \\ 
 & APE$_z$ & \cellcolor[HTML]{def3e6}\textbf{0.168} & \underline{1.519} & 2.514 & 11.352 &  12.668 &  & APE$_z$ & \cellcolor[HTML]{def3e6}\textbf{0.548} & \underline{4.161} & 5.425 & 15.080 & 9.741  \\ 
 \hline

\multirow{2}{*}{BiCorridor} & APE$_t$ & \cellcolor[HTML]{def3e6}\textbf{1.425} & 15.078 & 15.035 & \underline{11.867} & 13.990 & \multirow{2}{*}{Quad} & APE$_t$ & \cellcolor[HTML]{def3e6}\textbf{7.347} & \underline{11.851} & - & 36.969 & 122.381  \\ 
 & APE$_r$ & \cellcolor[HTML]{def3e6}\textbf{5.519}  & 44.116 & 51.048 & \underline{7.055} & 38.014 &  & APE$_r$ & \cellcolor[HTML]{def3e6}\textbf{3.356} & \textbf{6.218} & - & 15.488 & 79.279  \\ 
 & RPE$_t$ & \cellcolor[HTML]{def3e6}\textbf{0.063}  & 0.296 & \underline{0.152} & 0.164 &  0.216 &  & RPE$_t$ & \cellcolor[HTML]{def3e6}\textbf{0.080} & 0.157 & - & 0.155 & \underline{0.152} \\ 
 & RPE$_r$ & 0.885  & \textbf{0.840} & 0.918 & \underline{0.851} &  1.419  &  & RPE$_r$ & 0.838 & \textbf{0.543} & - & \underline{0.784} & 0.969 \\ 
 & APE$_z$ & \cellcolor[HTML]{def3e6}\textbf{0.226}  & 3.236 & \underline{2.880} & 11.389 & 7.868 &  & APE$_z$ & \cellcolor[HTML]{def3e6}\textbf{2.143} & \underline{2.924} & - & 21.621 & 4.012 \\ 
 \hline

\multirow{2}{*}{Downstair} & APE$_t$ & \cellcolor[HTML]{def3e6}\textbf{3.916} & 11.613 & \underline{8.323} & 13.448 & 30.346 & \multirow{2}{*}{MoCap-E} & APE$_t$ & \cellcolor[HTML]{def3e6}\textbf{0.852} & 1.416 & \underline{1.313} & 4.588 & 2.335 \\ 
 & APE$_r$ & \cellcolor[HTML]{def3e6}\textbf{3.415} & 10.791 & 8.595 & \underline{8.825} & 43.444 &  & APE$_r$ & \cellcolor[HTML]{def3e6}\textbf{3.373} & 5.441 & \underline{4.984} & 7.150 & 30.734 \\ 
 & RPE$_t$ & \cellcolor[HTML]{def3e6}\textbf{0.099} & 0.187 & 0.182 & \underline{0.159} & 0.174 &  & RPE$_t$ & \cellcolor[HTML]{def3e6}\textbf{0.184} & 0.246 & 0.201 & 0.211 & \underline{0.188} \\ 
 & RPE$_r$ & 1.080 & \textbf{0.658} & 1.095 & \underline{0.977} & 1.637 &  & RPE$_r$ & \cellcolor[HTML]{def3e6}\textbf{1.508} & \underline{1.844} & {1.921} & 2.166 & 3.534 \\ 
 & APE$_z$ & \cellcolor[HTML]{def3e6}\textbf{0.530} & \underline{2.490} & 4.843 & 11.141 & 5.601 &  & APE$_z$ & \cellcolor[HTML]{def3e6}\textbf{0.103} & 1.010 & 0.818 & 4.190 & \underline{0.501}  \\ 
 \hline

\multirow{2}{*}{Upstair} & APE$_t$ & \cellcolor[HTML]{def3e6}\textbf{1.496} & 6.238 & \underline{5.307} &  10.832 & 7.238 & \multirow{2}{*}{MoCap-H} & APE$_t$ & \cellcolor[HTML]{def3e6}\textbf{0.880} & 1.670 & \underline{0.900} & 4.004 & 2.798 \\ 
 & APE$_r$ & \cellcolor[HTML]{def3e6}\textbf{4.048} & 21.379 & 27.242 & \underline{4.352} & 30.934  &  & APE$_r$ & \cellcolor[HTML]{def3e6}\textbf{1.981} & 5.237 & 6.727 & \underline{4.744} & 31.101  \\ 
 & RPE$_t$ & \cellcolor[HTML]{def3e6}\textbf{0.071} & \underline{0.139} & 0.179 &  0.208 & 0.141 &  & RPE$_t$ & 0.204 & \underline{0.177} & 0.195 & 0.193 & \textbf{0.169} \\ 
 & RPE$_r$ & \underline{0.933} & \textbf{0.647} & 0.979 &  1.121 & 1.643 &  & RPE$_r$ & \cellcolor[HTML]{def3e6}\textbf{1.665} & {1.945} & \underline{1.917} & 2.225 & 5.109 \\ 
 & APE$_z$ & \cellcolor[HTML]{def3e6}\textbf{0.514} & \underline{4.485} & 3.206 & 10.676 & 5.574 &  & APE$_z$ & \cellcolor[HTML]{def3e6}\textbf{0.118} & 1.404 & \underline{0.283} & 3.704 & 0.711  \\ 

 \hline
\bottomrule
\end{tabular}
}
\end{table*}


In \texttt{MoCap-E}, Co-RaL fails to converge due to discrepancies between leg kinematics and radar. EKF-RIO and DeRO achieve similar \ac{APE}$_t$ and \ac{APE}$_z$, though DeRO exhibits higher \ac{APE}$_r$ and \ac{RPE}$_r$ since it relies solely on \ac{IMU}-based orientation estimation. River achieves the second-best performance, showing its potential in controlled indoor environments. Still, GaRLILEO delivers the most accurate odometry, effectively handling leg kinematics failures in specific zones and enhancing overall odometry estimation, especially in the vertical direction, by precisely estimating the gravity vector. This robustness stems from its B-spline-based continuous odometry scheme, which ensures stable trajectory estimation, even in the presence of discrepancies between sensor modalities.

\noindent\textit{12) MoCap-H:}
In \texttt{MoCap-H}, the results follow a similar trend. Because the slippery zone that caused Co-RaL to diverge in \texttt{MoCap-E} is shorter, Co-RaL successfully converges over the whole trajectory. However, due to persistent velocity discrepancies between radar and leg kinematics, Co-RaL, which integrates both modalities, performs slightly worse than EKF-RIO and River in terms of \ac{APE}$_t$ and \ac{APE}$_z$. Despite such discrepancies, GaRLILEO again produces the most accurate odometry by leveraging both sensor modalities while robustly handling modality failures that impair cooperative estimation.

\subsection{Leg Kinematics Odometry Comparison}
\label{sec:result.leg}

In this subsection, we compare GaRLILEO against four proprioceptive odometry methods that fuse \ac{IMU} and leg kinematics sensors, including leg joint encoders and contact sensors. The baseline methods are described as follows:

\begin{itemize}
    \item \textbf{Pronto}~\citep{camurri2020pronto}: A proprioceptive-only version of Pronto, enabling real-time odometry estimation at high frequency, compatible with control loop.
    \item \textbf{MUSE}~\citep{nistico2025muse}: A proprioceptive-only version of MUSE, a recent leg kinematics fused odometry that leverages a foot-slip detection algorithm for enhanced robustness.
    \item \textbf{Drift}~\citep{lin2023proprioceptive}: An invariant \ac{EKF}-based leg odometry algorithm that fuses contact estimation and gyroscope filtering, designed for low-cost legged robots.
    \item \textbf{Holistic}~\citep{nubert2025holistic}: A proprioceptive-only version of Holistic Fusion, leveraging foot contact points as landmark measurements. Our leg kinematics velocity estimation result is attached for the front-end. 
\end{itemize}

This subsection evaluates the odometry accuracy of proprioceptive methods, with a focus on their performance in diverse environments. \tabref{tab:leg_indoor} presents quantitative results for \ac{APE} and \ac{RPE} metrics. GaRLILEO consistently achieves the most accurate odometry estimates across most sequences, primarily due to its integration of \ac{SoC} radar-derived ego-velocity, which effectively mitigates the challenges posed by leg contact drift. 

\noindent\textit{1) Atrium:}
Pronto, Drift, and Holistic achieve similar \ac{APE}$_t$, while Pronto, through weighted averaging of leg kinematics velocities, attains the most accurate \ac{APE}$_z$, demonstrating the reliability of leg kinematics in this environment. MUSE, which incorporates a slip detection algorithm, achieves a comparable \ac{APE}$_z$ to Pronto and even lower \ac{APE}$_t$, indicating the presence of slip or drift in the contact frame even in low-dynamic indoor settings. GaRLILEO achieves the most accurate overall odometry, presenting the lowest error on every metric.

\begin{figure*}[!t]
    \centering
    \includegraphics[width=\textwidth]{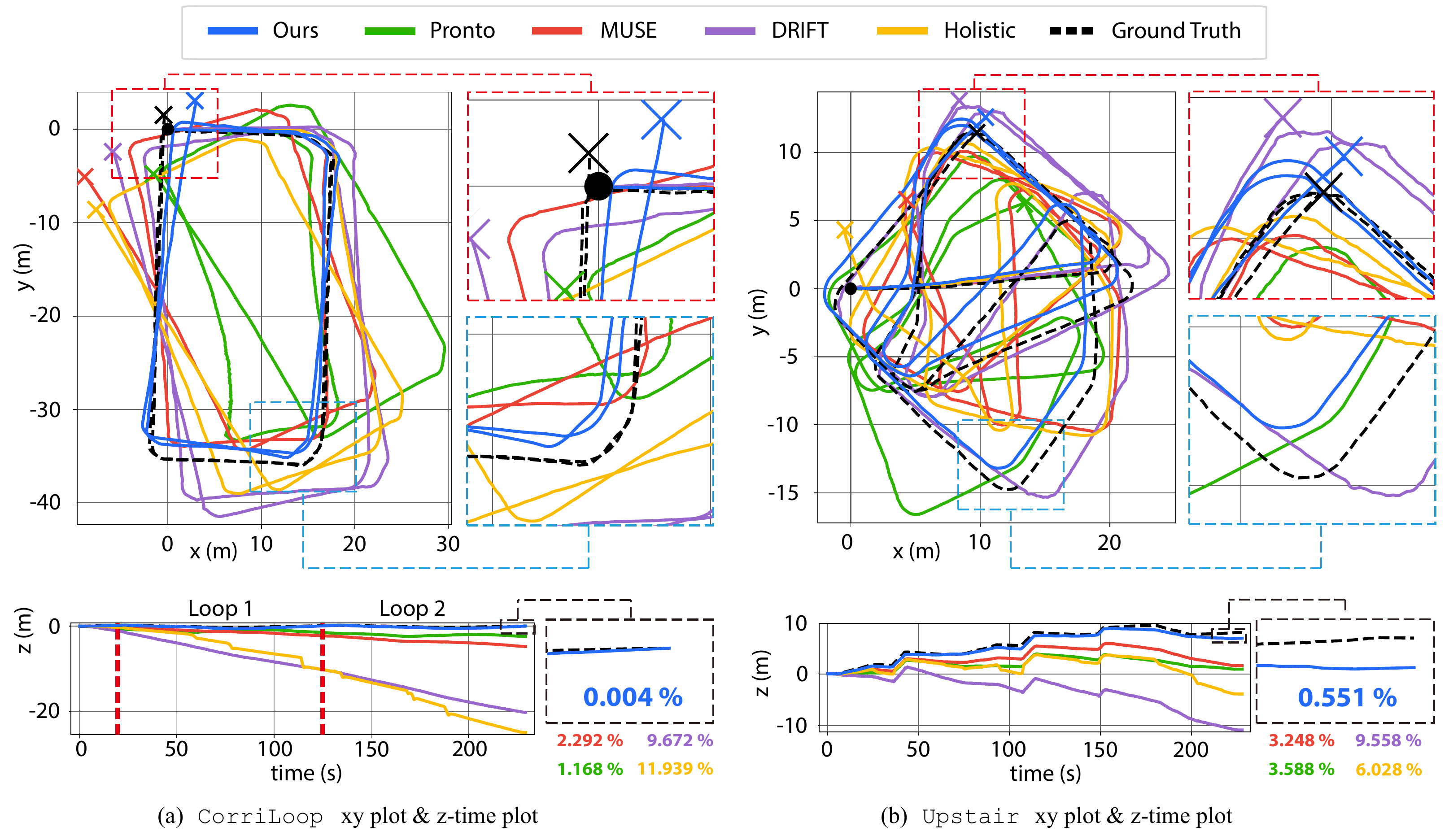}
    \label{fig:leg_odom}
    \vspace{-5mm}
    \caption{Leg-Kinematics baseline odometry results on the (a) \texttt{CorriLoop} and (b) \texttt{Upstair} sequences. Red dotted lines in (a, bottom) indicate the starting points of each loop in \texttt{CorriLoop} sequence. Both red zoomed-in views in (a) and (b) highlight that GaRLILEO integrating SoC radar and leg kinematics converge more closely to the ground-truth final position. The colored numbers on the right-hand side of the bottom plots show relative end-to-end vertical drift with respect to the full path length, where GaRLILEO achieves substantially lower drift compared with other baselines. Cyan zoomed view in (a) illustrates that GaRLILEO (blue) follows the most consistent turning trajectory across repeated loops. Similarly, in (b), the cyan zoomed view indicates that GaRLILEO (blue) follows the most accurate corner odometry during the upstairs with repeated turns. 
    Both subfigures qualitatively show the odometry accuracy of GaRLILEO, particularly in terms of vertical drift suppression. 
    }
\end{figure*}

\noindent\textit{2) BridgeLoop and CorriLoop:}
The \texttt{BridgeLoop} and \texttt{CorriLoop} sequences emphasize the importance of accurate roll and pitch estimation, particularly in mitigating vertical drift. As shown in \tabref{tab:leg_indoor}, MUSE is more robust than Pronto in terms of \ac{APE}$_t$. This highlights the advantage of a physical contact sensor that remains reliable indoors, including stairways and slopes. Drift and Holistic yield less accurate odometry, as they either assume a constant contact frame to correct \ac{IMU} drift or treat contact frames as landmark features. In contrast, GaRLILEO robustly mitigates contact slip effects through radar velocity estimation and a velocity bias term, achieving the most accurate odometry overall, particularly in the vertical direction, owing to precise roll and pitch observation and an accurate gravity estimation scheme.

\noindent\textit{3) BiCorridor:}
In the \texttt{BiCorridor} sequence, although the environment is similar to \texttt{CorriLoop}, the results differ due to the inclusion of an upstairs section with a narrow turn. Pronto and MUSE produce relatively inaccurate odometry, as indicated by large \ac{APE}$_r$ from diverging rotational estimates. A similar trend is observed in Holistic, although its errors are more biased toward the vertical direction. Interestingly, Drift achieves lower \ac{APE}$_r$ than the other baselines due to its gyro filter scheme, but still suffers from substantial vertical drift. In contrast, GaRLILEO delivers the lowest error except \ac{RPE}$_r$, owing to precise gravity estimation that effectively suppresses divergence, particularly in the vertical axis.

\noindent\textit{4) Downstair and Upstair:}
The \texttt{Downstair} and \texttt{Upstair} sequences enable a comparison between heavy stair descent and ascent. Among the leg kinematics-only baselines, Drift exhibits the largest \ac{APE}$_t$ and \ac{APE}$_z$, but achieves relatively low \ac{APE}$_r$. Its gyro filter effectively mitigates horizontal errors from contact impacts on stairs; however, its vulnerability in roll and pitch estimation leads to substantial vertical divergence. Due to the repetitive rotation during stair ascent, Pronto and MUSE show much higher \ac{APE}$_r$ in \texttt{Upstair} compared with \texttt{Downstair}, as they rely on \ac{IMU}-based rotational velocity for orientation estimation. All comparison methods record higher \ac{APE}$_t$ in the downstairs sequence, indicating that stair descent induces stronger contact drift. Despite these challenges, GaRLILEO achieves accurate vertical pose estimation with \ac{APE}$_z$ around 50\cm, delivering the most accurate and robust results through radar-based velocity estimation, particularly in stair environments.

\noindent\textit{5) SlopeStair Sequence:}
In the \texttt{SlopeStair} sequence, leg-odometry performance degrades severely on staircases and outdoor-slope segments, especially just before re-entry indoors, at which point Pronto fails to converge. Drift and Holistic succeed in converging by relying on the contact sensor, but their trajectories diverge significantly. Interestingly, MUSE yields even worse \ac{APE}$_t$, yet maintains sub-meter \ac{APE}$_z$. This reflects the effectiveness of MUSE’s slip detection module in handling vertical drift caused by contact slips, while horizontal drift—especially in yaw—remains difficult to suppress. In contrast, consistent with the radar–\ac{IMU} experiments, GaRLILEO achieves the lowest error on every metrics in this sequence.

\noindent\textit{6) Overpass:}
In the \texttt{Overpass} sequence, the most severe contact slips in leg odometry occur on the stair sections positioned before and after the overpass. Drift achieves competitive \ac{APE}$_r$, but suffers from large vertical drift, resulting in the highest \ac{APE}$_t$. MUSE and Holistic provide similar accuracy, particularly in \ac{APE}$_t$ and \ac{APE}$_z$. Pronto, although excluding its contact estimator, produces comparably robust odometry, especially in \ac{APE}$_r$ and \ac{APE}$_z$. In contrast, GaRLILEO achieves the lowest \ac{APE}$_t$ and \ac{APE}$_z$, demonstrating robustness to contact impacts during stair ascent and descent through its local gravity–based roll and pitch estimation.

\noindent\textit{7) Tunnel:}
In the \texttt{Tunnel} sequence, a single sharp U-turn largely determines the overall \ac{APE} level, as it is the only rotational motion in the entire sequence. Drift achieves competitive horizontal rotation accuracy, recording the second-lowest \ac{APE}$_r$. However, it fails to maintain robust horizontal orientation, resulting in the highest \ac{APE}$_z$. Holistic yields the highest \ac{APE}$_r$, and this single divergence in orientation estimation also leads to the largest \ac{APE}$_t$, despite a moderate \ac{APE}$_z$. Pronto and MUSE achieve similar \ac{APE}$_z$, incorporating joint-encoder-based contact estimation and slip detection, respectively. In contrast, GaRLILEO delivers the most robust odometry across all metrics, owing to precise roll and pitch estimation before and after the sharp U-turn. This is evident in its lowest \ac{APE}$_z$, which is approximately 13\% of the second-best method.

\noindent\textit{8) Quad:}
The \texttt{Quad} sequence involves substantial contact drift, as it covers the longest path among all datasets and includes outdoor stairs and a long slope. Holistic shows low vertical APE (\ac{APE}$_z$); however, orientation divergence causes overall odometry failure, as indicated by the highest \ac{APE}$_t$ and \ac{APE}$_r$. Drift produces more precise orientation estimates thanks to its gyro filtering scheme, yet still suffers from large \ac{APE}$_t$ due to significant vertical drift. MUSE fails to converge over the full trajectory, unable to handle divergence in orientation estimation. Interestingly, Pronto achieves the best performance in both \ac{APE}$_t$ and \ac{APE}$_r$, demonstrating the potential of joint-encoder-based contact estimation, particularly in long outdoor environments. Even on this long outdoor sequence, GaRLILEO delivers the most accurate vertical estimation with its local-gravity model.

\noindent\textit{9) MoCap-E and MoCap-H:}
In the \texttt{MoCap} sequences, discrepancies in leg odometry arise in both horizontal and vertical directions, as also observed in the radar-based experiments. Drift records the highest \ac{APE}$_t$ and \ac{APE}$_z$, caused by vertical divergence in the cushion zone and horizontal errors in the sliding zone. Notably, all leg kinematics–\ac{IMU} baselines exhibit limited accuracy in the sliding zone, since they lack a sensor modality capable of capturing the robot’s state in a dynamic floor. Compared with Holistic, both Pronto and MUSE achieve lower \ac{APE}$_t$, showing greater robustness to contact failure through joint encoder–based contact estimation or slip detection. Among these, MUSE achieves the lowest \ac{APE}$_t$ and \ac{APE}$_z$, as its slip detection module effectively mitigates contact failure, leading to more robust odometry estimation. Across the two \texttt{MoCap} sequences, GaRLILEO is most accurate, thanks to radar-derived ego-velocity fed into the velocity spline, especially in slippery sections where leg-kinematics-based estimation fails.

\subsection{Complementary Effects of Radar and Leg Kinematics}
\label{sec:result.fusion}

To analyze the complementary roles of radar and leg kinematics, we compare three GaRLILEO configurations: the full system \textbf{R+L+I}, a radar-IMU-only variant \textbf{R+I} in which the leg kinematics factor $\mathbf{r}_L$ is disabled, and a leg-IMU-only variant \textbf{L+I} in which the radar factor $\mathbf{r}_R$ is disabled. We additionally report EKF-RIO and MUSE for the SNU sequences, and River and MUSE for the RAI sequences, as they are the best-performing representative radar-based and leg-based comparisons of each sensor configuration.


\begin{table}[t]
\small\sf\centering
\caption{Complementary effects of radar and leg kinematics on SNU sequences. \textbf{L+I} refers to the leg kinematics-only and \textbf{R+I} refers to the radar-only version of the full GaRLILEO (\textbf{R+L+I}).}
\label{tab:seq_result_full}
\resizebox{\columnwidth}{!}{
\begin{tabular}{l|c|ccc|cc}
\toprule  \hline
& & \textbf{R+L+I} & \textbf{R+I} & \textbf{L+I} & \textbf{EKF-RIO} & \textbf{MUSE} \\
\hline

\multirow{4}{*}{Atrium}
& APE$_t$    & \cellcolor[HTML]{def3e6}\textbf{0.816} & 9.793  & \underline{1.423} & 6.679 & 3.297 \\
& RPE$_t$    & \cellcolor[HTML]{def3e6}\textbf{0.055} & 0.191  & \underline{0.085} & 0.162 & 0.090 \\
& APE$_z$    & \cellcolor[HTML]{def3e6}\textbf{0.106} & 9.771  & \underline{0.192} & 4.908 & 0.685 \\
& APE$_{xy}$ & \underline{0.809} & \cellcolor[HTML]{def3e6}\textbf{0.661} & 1.410 & 4.530 & 3.225 \\
\hline

\multirow{4}{*}{BridgeLoop}
& APE$_t$    & \underline{1.193} & 11.802 & \cellcolor[HTML]{def3e6}\textbf{1.051} & 7.125 & 4.195 \\
& RPE$_t$    & \cellcolor[HTML]{def3e6}\textbf{0.080} & 0.167  & 0.143 & \underline{0.132} & 0.200 \\
& APE$_z$    & \cellcolor[HTML]{def3e6}\textbf{0.117} & 11.773 & \underline{0.187} & 5.302 & 2.250 \\
& APE$_{xy}$ & 1.187 & \cellcolor[HTML]{def3e6}\textbf{0.827} & \underline{1.034} & 4.760 & 3.541 \\
\hline

\multirow{4}{*}{CorriLoop}
& APE$_t$    & \underline{1.627} & 17.758 & \cellcolor[HTML]{def3e6}\textbf{1.539} & 9.283 & 5.811 \\
& RPE$_t$    & \cellcolor[HTML]{def3e6}\textbf{0.066} & 0.202  & \underline{0.093} & 0.180 & 0.127 \\
& APE$_z$    & \underline{0.168} & 17.667 & \cellcolor[HTML]{def3e6}\textbf{0.096} & 4.001 & 2.514 \\
& APE$_{xy}$ & \underline{1.619} & 1.801  & \cellcolor[HTML]{def3e6}\textbf{1.536} & 8.377 & 5.238 \\
\hline

\multirow{4}{*}{BiCorridor}
& APE$_t$    & \underline{1.425} & 23.785 & \cellcolor[HTML]{def3e6}\textbf{1.277} & 10.293 & 15.035 \\
& RPE$_t$    & \cellcolor[HTML]{def3e6}\textbf{0.063} & 0.220  & \underline{0.117} & 0.185  & 0.152 \\
& APE$_z$    & 0.226 & 23.740 & \cellcolor[HTML]{def3e6}\textbf{0.144} & 5.497  & \underline{2.880} \\
& APE$_{xy}$ & \underline{1.407} & 1.472  & \cellcolor[HTML]{def3e6}\textbf{1.269} & 8.702  & 14.757 \\
\hline

\multirow{4}{*}{Downstair}
& APE$_t$    & \underline{3.961} & 10.315 & \cellcolor[HTML]{def3e6}\textbf{3.933} & 30.835 & 8.323 \\
& RPE$_t$    & \cellcolor[HTML]{def3e6}\textbf{0.099} & 0.169  & \underline{0.123} & 0.229  & 0.182 \\
& APE$_z$    & \cellcolor[HTML]{def3e6}\textbf{0.530} & 10.262 & \underline{0.738} & 4.251  & 4.843 \\
& APE$_{xy}$ & 3.926 & \cellcolor[HTML]{def3e6}\textbf{1.049} & \underline{3.863} & 30.541 & 6.768 \\
\hline

\multirow{4}{*}{Upstair}
& APE$_t$    & \cellcolor[HTML]{def3e6}\textbf{1.496} & 17.446 & \underline{1.633} & 10.044 & 5.307 \\
& RPE$_t$    & \cellcolor[HTML]{def3e6}\textbf{0.071} & 0.185  & \underline{0.135} & 0.143  & 0.179 \\
& APE$_z$    & \cellcolor[HTML]{def3e6}\textbf{0.514} & 17.339 & \underline{0.575} & 8.871  & 3.206 \\
& APE$_{xy}$ & \cellcolor[HTML]{def3e6}\textbf{1.405} & 1.932  & \underline{1.529} & 4.711  & 4.230 \\
\hline

\multirow{4}{*}{SlopeStair}
& APE$_t$    & \cellcolor[HTML]{def3e6}\textbf{2.359} & 22.781 & \underline{3.452} & \multicolumn{1}{c}{-} & 35.519 \\
& RPE$_t$    & \cellcolor[HTML]{def3e6}\textbf{0.061} & 0.237  & \underline{0.120} & \multicolumn{1}{c}{-} & 0.151 \\
& APE$_z$    & \cellcolor[HTML]{def3e6}\textbf{0.665} & 21.894 & 1.142 & \multicolumn{1}{c}{-} & \underline{0.967} \\
& APE$_{xy}$ & \cellcolor[HTML]{def3e6}\textbf{2.264} & 6.295  & \underline{3.257} & \multicolumn{1}{c}{-} & 35.506 \\
\hline

\multirow{4}{*}{Overpass}
& APE$_t$    & \cellcolor[HTML]{def3e6}\textbf{1.526} & 6.531  & \underline{2.048} & \multicolumn{1}{c}{-} & 6.507 \\
& RPE$_t$    & \cellcolor[HTML]{def3e6}\textbf{0.091} & 0.329  & \underline{0.135} & \multicolumn{1}{c}{-} & 0.191 \\
& APE$_z$    & 1.074 & 6.248  & \cellcolor[HTML]{def3e6}\textbf{0.410} & \multicolumn{1}{c}{-} & \underline{2.442} \\
& APE$_{xy}$ & \cellcolor[HTML]{def3e6}\textbf{1.084} & \underline{1.902} & 2.007 & \multicolumn{1}{c}{-} & 6.031 \\
\hline

\multirow{4}{*}{Tunnel}
& APE$_t$    & \cellcolor[HTML]{def3e6}\textbf{3.523} & 17.355 & \underline{3.950} & 26.800 & 10.194 \\
& RPE$_t$    & \underline{0.083} & 0.185  & \cellcolor[HTML]{def3e6}\textbf{0.079} & 0.316  & 0.105 \\
& APE$_z$    & \cellcolor[HTML]{def3e6}\textbf{0.548} & 4.836  & \underline{0.941} & 6.139  & 5.425 \\
& APE$_{xy}$ & \cellcolor[HTML]{def3e6}\textbf{3.481} & 16.668 & \underline{3.836} & 26.088 & 8.631 \\
\hline

\multirow{4}{*}{Quad}
& APE$_t$    & \cellcolor[HTML]{def3e6}\textbf{7.347} & 53.857 & \underline{9.662} & \multicolumn{1}{c}{-} & \multicolumn{1}{c}{-} \\
& RPE$_t$    & \cellcolor[HTML]{def3e6}\textbf{0.080} & 0.290  & \underline{0.084} & \multicolumn{1}{c}{-} & \multicolumn{1}{c}{-} \\
& APE$_z$    & \cellcolor[HTML]{def3e6}\textbf{2.143} & 37.264 & \underline{4.701} & \multicolumn{1}{c}{-} & \multicolumn{1}{c}{-} \\
& APE$_{xy}$ & \cellcolor[HTML]{def3e6}\textbf{7.028} & 38.883 & \underline{8.442} & \multicolumn{1}{c}{-} & \multicolumn{1}{c}{-} \\

\bottomrule \hline
\end{tabular}
}
\end{table}

\begin{figure*}[!t]
    \centering
    \includegraphics[width=\textwidth]{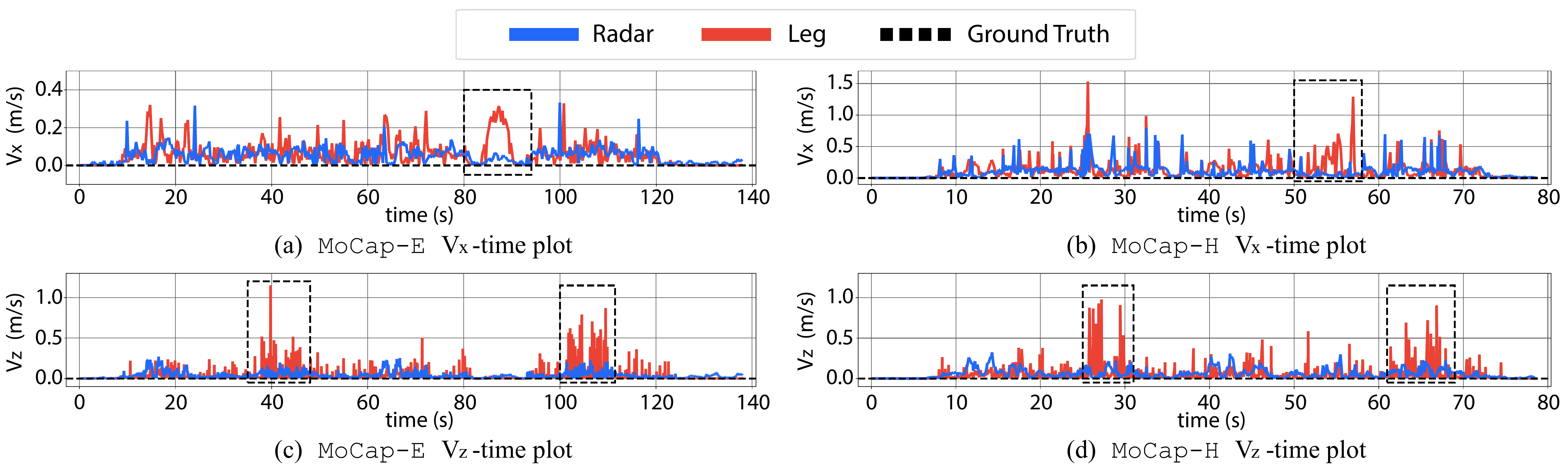}
    \caption{$v_x$-time and $v_z$-time Velocity Error (difference from ground truth) in the \texttt{MoCap-E} and \texttt{MoCap-H} sequences. A dotted square on the (a) and (b) includes a slippery zone where the floor moves backwards, while two dotted squares on the (c) and (d) include cushion zones where the leg kinematics velocity presents a high impact on the vertical direction. }
    \label{fig:mocap_velocity}
\end{figure*}

\noindent\textit{1) SNU Sequences:}
\tabref{tab:seq_result_full} summarizes the results on the SNU sequences. Overall, the sensor-subset variants \textbf{R+I} and \textbf{L+I} exhibit complementary strengths. When radar returns are sufficiently informative, as is often the case in indoor environments, \textbf{R+I} can provide competitive horizontal constraints in terms of APE$_{xy}$ on several sequences, whereas \textbf{L+I} substantially improves vertical robustness in terms of APE$_z$ by leveraging high-rate proprioceptive velocity information. By fusing both modalities, \textbf{R+L+I} attains the best accuracy on most sequences, reflecting cooperative fusion that emphasizes the more reliable modality when the other becomes less informative.

Regarding the modality-specific comparisons, \textbf{L+I} is consistently competitive with, and often outperforms MUSE on SNU sequences, suggesting that our continuous-time formulation and factor design provide strong proprioceptive odometry. In practice, the radar input is available at \unit{20}{\Hz}, and using radar alone is not sufficient to fully realize the benefits of our proposed modules; as a result, \textbf{R+I} tends to perform at a similar level to dedicated radar-IMU pipeline, EKF-RIO on several sequences.

\noindent\textit{2) RAI Sequences:}
To further evaluate complementarity under conditions where leg-kinematics-based odometry degrades, we conduct additional analysis on the \texttt{MoCap-E} and \texttt{MoCap-H} sequences, which include intentional perturbations in leg odometry, as visualized in \figref{fig:mocap_velocity}. Quantitative results are reported in \tabref{tab:mocap_result}.

\begin{figure*}[!t]
    \centering
    \includegraphics[width=\textwidth]{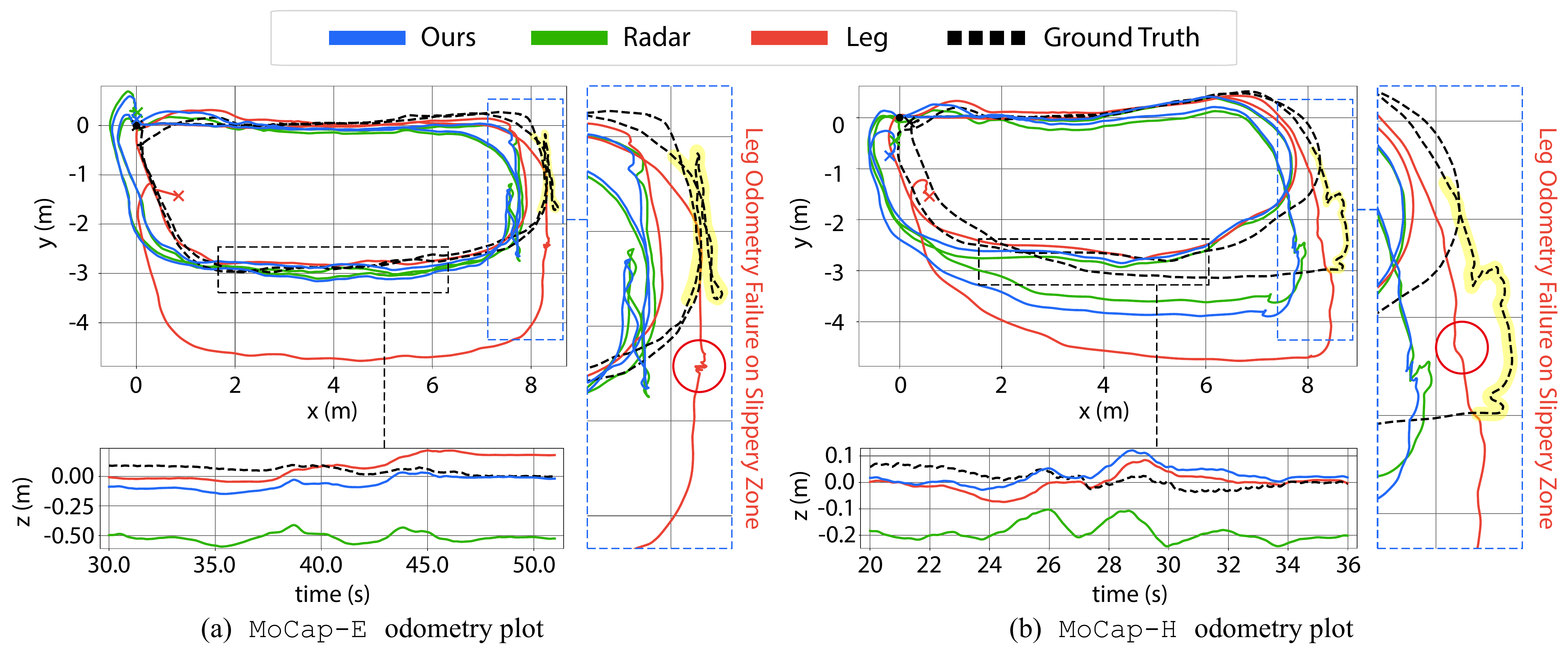}
    \caption{xy plot and z-time plot of \texttt{MoCap} sequences. (a) and (b) are xy and z-time odometry plot of \texttt{MoCap-E} and \texttt{MoCap-H} sequences, respectively. Blue dotted box: zoomed view of slippery zone where leg odometry fails temporarily, especially on the xy plane. The yellow-highlighted area is the exact slippery zone. Black dotted box: zoomed side view of cushion zone where leg odometry fails temporarily, especially on the vertical direction. Comparing the radar-only, leg kinematics-only, and full module versions of GaRLILEO to check the cooperative effect of radar and leg sensors in the slippery zone to check the cooperative work between radar and leg kinematics. }
    \label{fig:mocap_plot}
\end{figure*}

On \texttt{MoCap-E} sequence, where the robot stays stop on the slippery zone during the floor is moving to backward, significant difference between radar-aided methods (\ie \textbf{R+L+I}, \textbf{R+I}, and River) presents comparably more accurate \ac{APE}$_{xy}$ then leg kinematic only methods (\ie \textbf{L+I} and MUSE), proving the radar sensor is working cooperatively with leg kinematic sensors when the stationary plane assumption breaks. In \ac{APE}$_z$, River presents lower error compared with MUSE, as the cushion zone ruins the leg kinematic estimation. Similarly, as can be found from \figref{fig:mocap_plot}~(a), only \textbf{R+I} keeps stable odometry during the cushion zone, while leg kinematic aided ones goes slightly upper direction. However, due to the local gravity spline which is more benefited by the high frequency egocentric-velocity measurement of leg kinematics, \ac{APE}$_z$ of \textbf{R+I} is higher than other versions of GaRLILEO. Still, the combination of both sensors, full GaRLILEO, presents the most accurate odometry estimation. It can be concluded that radar and leg kinematics can work in complementary ways to estimate the robot's odometry in a slippery environment accurately. 

On \texttt{MoCap-H} sequence, though the existence of cushion zone and slippery zone is similar, details like maintaining walking on slippery zone and higher overall robot speed are different. As can be found from \figref{fig:mocap_plot}~(b), \textbf{L+I} fails at a similar point due to the broken stationary floor assumption in the slippery zone. Because of this, \textbf{L+I} results in higher \ac{APE}$_{xy}$ compared with other GaRLILEO variants. On vertical drift, the tendency of \ac{APE}$_z$ is similar to \texttt{MoCap-E}, but the upper direction bias of leg kinematics in the cushion zone is notably mild because of the locomotion speed difference. Due to the higher speed of the robot, the effect of the cushion less emerges and leads to a highly accurate \ac{APE}$_z$ of \textbf{L+I}. In conclusion, exploiting the biased accuracy of \textbf{R+I} and \textbf{L+I} on \ac{APE}$_{xy}$ and \ac{APE}$_z$ each, \textbf{R+L+I} presents most accurate \ac{APE}$_t$ overall, while lying between \textbf{R+I} and \textbf{L+I} on other \ac{APE} metrics. This tendency to rely more on relatively more accurate sensor modalities can be inferred as the cooperative effect of GaRLILEO. 

\begin{table}[tbp]
\small\sf\centering
\caption{Complementary effects of radar and leg kinematic on RAI sequences. \textbf{L+I} refers to the leg kinematics-only and \textbf{R+I} refers to the radar-only version of the full GaRLILEO (\textbf{R+L+I}).}
\label{tab:mocap_result}
\resizebox{\columnwidth}{!}{
\begin{tabular}{l|c|c|cc|cc}
\toprule  \hline
& & \textbf{R+L+I} & \textbf{R+I} & \textbf{L+I} & \textbf{River} & \textbf{MUSE} \\ \hline
\multirow{4}{*}{MoCap-E} & APE$_t$  & \cellcolor[HTML]{def3e6}\textbf{0.852}  & {1.010}   & 1.171  & \underline{0.916} & 1.313 \\  
& RPE$_t$  & \cellcolor[HTML]{def3e6}\textbf{0.184}  & {0.243}   & 0.318  & 0.239 & \underline{0.201} \\ 
& APE$_{z}$ & \cellcolor[HTML]{def3e6}\textbf{0.103}   & 0.563  & 0.256 & \underline{0.236} & 0.818  \\  
& APE$_{xy}$   & \underline{0.846}    & \cellcolor[HTML]{def3e6}{\textbf{0.838}}    & 1.143  & 0.885  & 1.027 \\ \hline
\multirow{4}{*}{MoCap-H} &  APE$_t$  & \cellcolor[HTML]{def3e6}{\textbf{0.880}}     & 1.038   & 0.920 & 1.145 & \underline{0.900}  \\ 
& RPE$_t$  &  \underline{0.204}   & 0.242   & 0.220   & 0.250  & \textbf{0.195} \\ 
& APE$_{z}$  & \underline{0.118}   & 0.643  & \cellcolor[HTML]{def3e6}{\textbf{0.102}} & 0.236  & 0.283  \\
& APE$_{xy}$ & {0.871}     & \cellcolor[HTML]{def3e6}\textbf{0.814}  & 0.915  & 1.120 & \underline{0.854} \\   \hline \bottomrule
\end{tabular}
 }
\end{table}

\subsection{Effect of Gravity Factors on State Estimation Accuracy}
\label{sec:result.gravity}

In this subsection, we evaluate the effect of our gravity factors $\mathbf{r_{\mathcal{S}^2}}, \mathbf{r_{post}}$ on two different sub-experiments: 1) effect of the soft $\mathcal{S}^2$-constrained gravity factor, and 2) effect of the post optimization. 

\noindent\textit{1) Effect of the Soft $\mathcal{S}^2$-Constrained Gravity Factor:}
In the first sub-experiment, we aimed to highlight the effectiveness of the soft $\mathcal{S}^2$-constrained gravity factor by comparing the accuracy of the estimated local gravity vector between the full model of GaRLILEO: \textbf{Ours} and the model without the factor: \textbf{w/o} $\mathbf{r_{\mathcal{S}^2}}$. Ground-truth local gravity $\mathbf{g}^{*}_t$ was obtained by (i) rigidly aligning the ground truth pose to the estimated trajectory with Evo Trajectory Evaluator~\citep{grupp2017evo}, (ii) B-spline interpolation of the aligned poses at the estimating timestamps, and (iii) rotating the global gravity vector into the local frame using aligned ground truth orientation as follows:
\begin{equation}
\mathbf{g}^{*}_t \;=\; \mathbf{R}_t^{\top}\,\mathbf{g}_0, ~~\text{where}~~~ \mathbf{g}_0 = \begin{bmatrix} 0 \\ 0 \\ -9.81 \end{bmatrix}\,.
\end{equation}
To evaluate roll and pitch observation accuracy, we use the mean angular error of the estimated local gravity vectors over each sequence.


\begin{table}[!b]
\small\sf\centering
\caption{Effect of $\mathbf{r_{\mathcal{S}^2}}$ Factor on Estimating Local Gravity}
\label{tab:gravity_result}
\resizebox{\columnwidth}{!}{
\begin{tabular}{c|c|c|c|c}
\toprule
\hline
\degree\;(deg) & Atrium & BridgeLoop & CorriLoop & BiCorridor \\ \hline
\textbf{Ours} & \cellcolor[HTML]{def3e6}\textbf{1.507} & \cellcolor[HTML]{def3e6}\textbf{2.000} & \cellcolor[HTML]{def3e6}\textbf{1.438} & \cellcolor[HTML]{def3e6}\textbf{1.483} \\
\textbf{w/o} $\mathbf{r_{\mathcal{S}^2}}$ & 4.815 & 4.229 & 2.735 & 4.651 \\
\hline \hline
\degree\;(deg) & Downstair & Upstair & SlopeStair & Overpass \\ \hline
\textbf{Ours} & \cellcolor[HTML]{def3e6}\textbf{1.727} & \cellcolor[HTML]{def3e6}\textbf{1.812} & \cellcolor[HTML]{def3e6}\textbf{2.147} & \cellcolor[HTML]{def3e6}\textbf{1.682}  \\
\textbf{w/o} $\mathbf{r_{\mathcal{S}^2}}$ & 4.637 & 5.092 & 5.436 & 3.703  \\
\hline \hline
\degree\;(deg) & Tunnel & Quad & MoCap-E & MoCap-H \\ \hline
\textbf{Ours} & \cellcolor[HTML]{def3e6}\textbf{1.851} & \cellcolor[HTML]{def3e6}\textbf{2.702} & \cellcolor[HTML]{def3e6}\textbf{4.244} & \cellcolor[HTML]{def3e6}\textbf{3.904} \\
\textbf{w/o} $\mathbf{r_{\mathcal{S}^2}}$ & 2.923 & 4.040 & 5.046 & 5.718 \\
\hline
\bottomrule
\end{tabular}
}
\end{table}

As detailed in \tabref{tab:gravity_result}, the ablation without the soft $\mathcal{S}^2$ gravity factor (\textbf{w/o} $\mathbf{r_{\mathcal{S}^2}}$) estimates local gravity with errors exceeding twice those of \textbf{Ours}. This behavior is closely related to gravity-norm preservation during continuous-time estimation. Although we clamp the gravity control points to have magnitude $9.81~\mathrm{m/s^2}$, gravity is parameterized with an $\mathbb{R}^3$ B-spline rather than an $\mathcal{S}^2$ spline; hence, the interpolated B-spline segment between adjacent control points is not inherently norm-preserving. As a result, without the soft $\mathcal{S}^2$ factor, $\|\mathbf{g}(t)\|$ may exhibit local norm drops between control points, which degrade the reliability of local gravity direction estimation, especially under strong contact-induced vibrations.

\begin{figure}[!t]
    \centering 
    \includegraphics[width=\columnwidth]{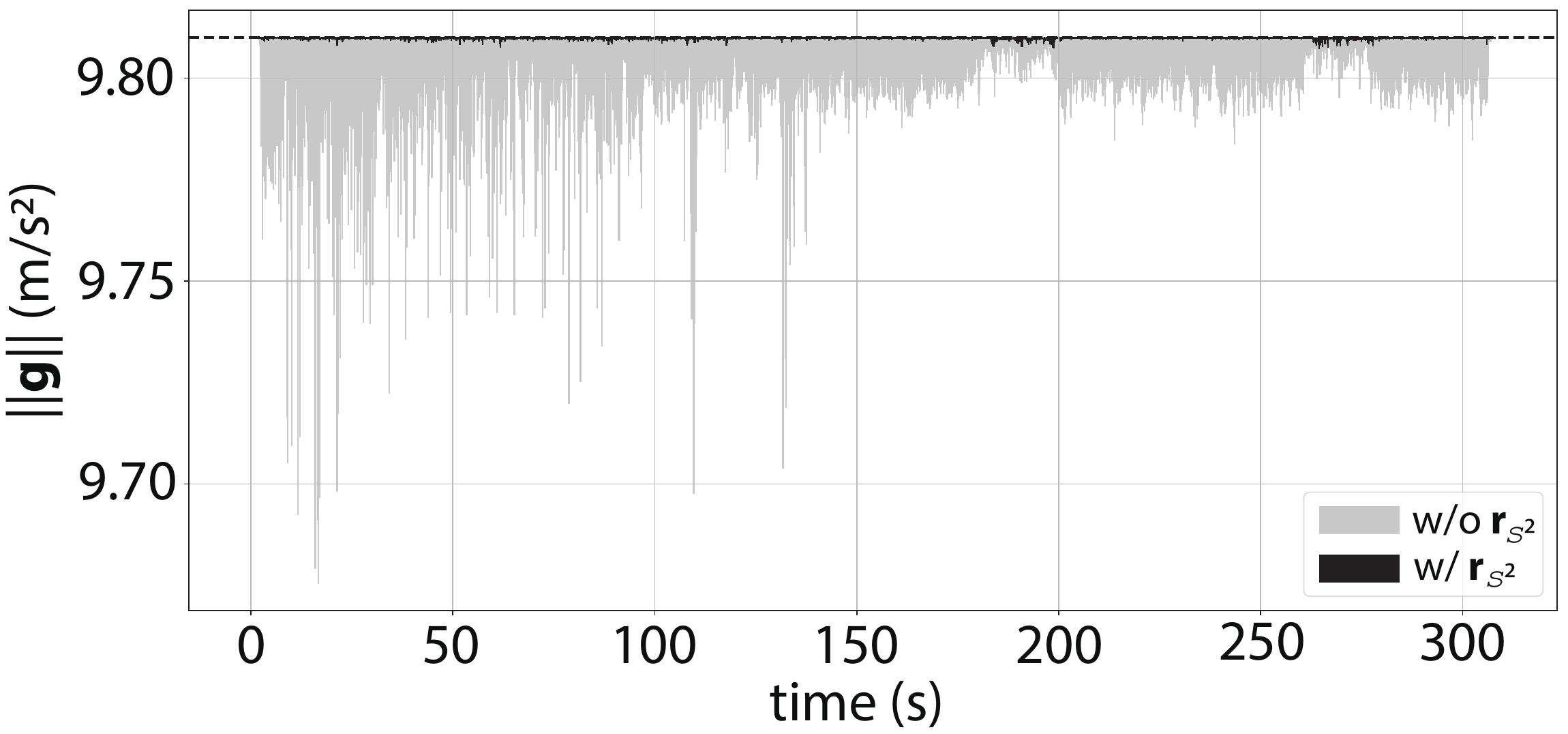}
    \caption{Gravity-norm behavior over time in \texttt{SlopeStair} sequence (w/ vs.\ w/o $\mathbf{r_{\mathcal{S}^2}}$). Enabling equation~\eqref{eq:s2_manifold} suppresses local norm dips of $\|\mathbf{g}(t)\|$ between adjacent control points, keeping $\|\mathbf{g}(t)\|$ much closer to $9.81~\mathrm{m/s^2}$. }
    \label{fig:gravity_norm_time}
\end{figure}

\begin{table}[!b]
\sf\centering
\caption{Gravity magnitude statistics (m/s$^2$) without and with the $\mathbf{r_{\mathcal{S}^2}}$ factor.}
\label{tab:gravity_stats_compact}
\setlength{\tabcolsep}{3.6pt}
\renewcommand{\arraystretch}{1.12}
\footnotesize

\begin{adjustbox}{max width=\linewidth}
\begin{tabular}{l cccc cccc}
\toprule
& \multicolumn{4}{c}{w/o $\mathbf{r_{\mathcal{S}^2}}$} & \multicolumn{4}{c}{w/ $\mathbf{r_{\mathcal{S}^2}}$} \\
\cmidrule(lr){2-5}\cmidrule(lr){6-9}
Sequence & Min & Max & Mean & Std. & Min & Max & Mean & Std. \\
\midrule
Atrium      & 9.7250 & 9.8100 & 9.8046 & 6.43$\times$10$^{-3}$ & 9.8088 & 9.8100 & 9.8099 & 7.53$\times$10$^{-5}$ \\
BridgeLoop  & 9.6489 & 9.8100 & 9.8004 & 1.11$\times$10$^{-2}$ & 9.8069 & 9.8100 & 9.8099 & 1.90$\times$10$^{-4}$ \\
CorriLoop   & 9.6888 & 9.8100 & 9.7960 & 1.43$\times$10$^{-2}$ & 9.8081 & 9.8100 & 9.8099 & 8.87$\times$10$^{-5}$ \\
BiCorridor  & 9.7335 & 9.8100 & 9.7974 & 1.08$\times$10$^{-2}$ & 9.8070 & 9.8100 & 9.8099 & 1.41$\times$10$^{-4}$ \\
Downstair   & 9.7123 & 9.8100 & 9.8044 & 9.14$\times$10$^{-3}$ & 9.8047 & 9.8100 & 9.8099 & 2.75$\times$10$^{-4}$ \\
Upstair     & 9.6334 & 9.8100 & 9.7866 & 2.27$\times$10$^{-2}$ & 9.8026 & 9.8100 & 9.8099 & 2.18$\times$10$^{-4}$ \\
SlopeStair  & 9.6722 & 9.8100 & 9.8013 & 1.03$\times$10$^{-2}$ & 9.8071 & 9.8100 & 9.8099 & 2.08$\times$10$^{-4}$ \\
Overpass    & 9.7189 & 9.8100 & 9.8036 & 8.90$\times$10$^{-3}$ & 9.8060 & 9.8100 & 9.8099 & 2.57$\times$10$^{-4}$ \\
Tunnel      & 9.6534 & 9.8100 & 9.8015 & 1.04$\times$10$^{-2}$ & 9.8073 & 9.8100 & 9.8099 & 1.10$\times$10$^{-4}$ \\
Quad        & 9.7016 & 9.8100 & 9.8053 & 8.17$\times$10$^{-3}$ & 9.8076 & 9.8100 & 9.8099 & 1.34$\times$10$^{-4}$ \\
MoCap-E     & 9.7940 & 9.8100 & 9.8087 & 1.74$\times$10$^{-3}$ & 9.8085 & 9.8100 & 9.8099 & 1.30$\times$10$^{-4}$ \\
MoCap-H     & 9.6579 & 9.8100 & 9.7988 & 1.44$\times$10$^{-2}$ & 9.7959 & 9.8100 & 9.8098 & 4.44$\times$10$^{-4}$ \\
\bottomrule
\end{tabular}
\end{adjustbox}
\end{table}

To make this explicit, we conducted additional experiments to measure and compare the gravity-norm behavior with and without the soft $\mathcal{S}^{2}$ factor $\mathbf{r_{\mathcal{S}^2}}$. Specifically, we compute the minimum, maximum, mean, and standard deviation of the gravity vector's magnitude $\|\mathbf{g}(t)\|$ over each sequence, which are summarized in \tabref{tab:gravity_stats_compact}. The minimum captures the worst-case local norm dip, while the standard deviation reflects the overall fluctuation. Additionally, we plot the gravity-norm behavior over runtime on seq.~\texttt{SlopeStair} in \figref{fig:gravity_norm_time}.

As shown in \figref{fig:gravity_norm_time}, soft $\mathcal{S}^{2}$ factor $\mathbf{r_{\mathcal{S}^2}}$ suppresses local norm dips between adjacent control points. \tabref{tab:gravity_stats_compact} further quantifies this effect across every sequence: without $\mathbf{r_{\mathcal{S}^2}}$, the minimum $\|\mathbf{g}(t)\|$ can drop to $9.633~\mathrm{m/s^2}$ and the standard deviation is on the order of $10^{-2}~\mathrm{m/s^2}$, whereas with $\mathbf{r_{\mathcal{S}^2}}$ the minimum dip improves to $9.796~\mathrm{m/s^2}$ and the standard deviation decreases to the order of $10^{-4}~\mathrm{m/s^2}$. The maximum stays at $9.81~\mathrm{m/s^2}$ because the B-spline evaluates $\mathbf{g}(t)$ as a convex weighted combination of control points with nonnegative weights that sum to one, and all control points are clamped to have magnitude $9.81~\mathrm{m/s^2}$.

\begin{figure}[!t]
    \centering
    \includegraphics[width=\columnwidth]{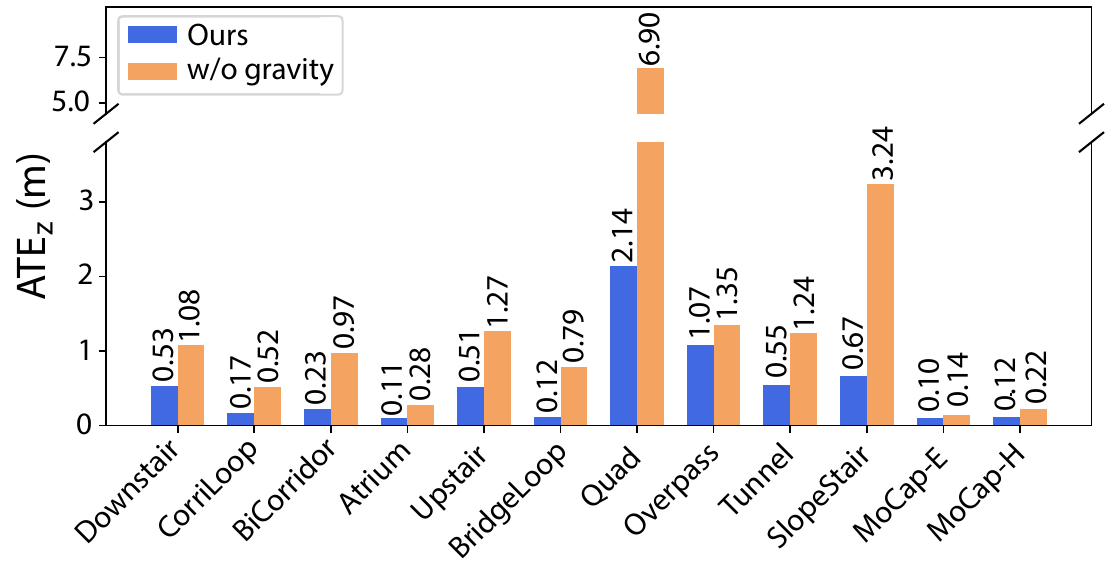}
    \caption{Effect of $\mathbf{r_{post}}$ on vertical pose estimation. Vertical odometry accuracy is notably enhanced when $\mathbf{r_{post}}$ is added to refine the rotation using the estimated local gravity in a post optimization step. This result supports the idea that precise local gravity estimation can provide roll and pitch observations, thereby mitigating vertical drift in odometry.}
    \label{fig:gravity_residual}
\end{figure}
\begin{figure}[!t]
    \centering 
    \includegraphics[width=\columnwidth]{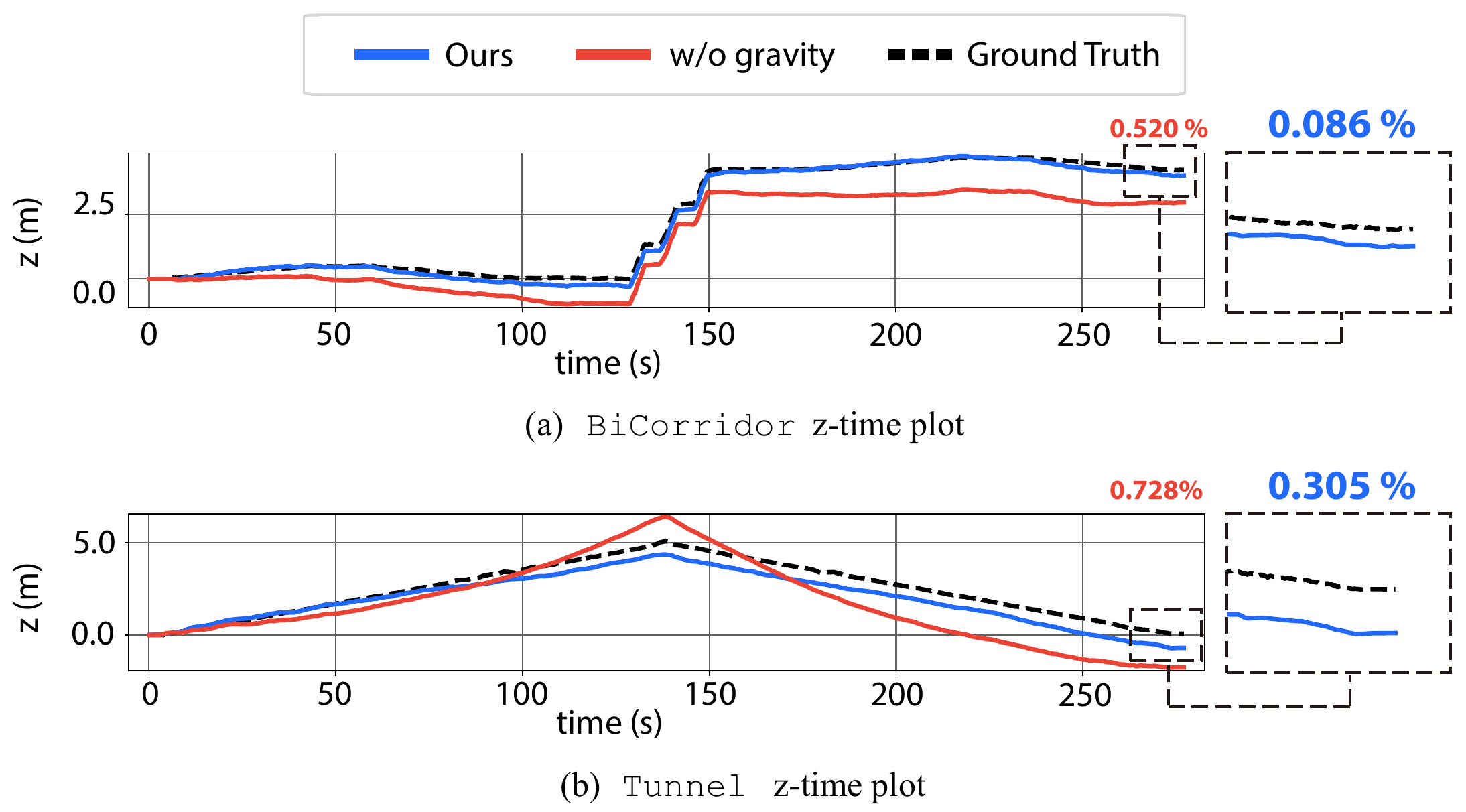}
    \caption{Z-time odometry plot including comparison between with and without the post optimization. $\mathbf{r_{post}}$ factor prevents odometry from diverging in the vertical direction. }
    \label{fig:gravity_ablation}
\end{figure}

\noindent\textit{2) Effect of the Post Optimization:}
In the second sub-experiment, we compared the direct vertical drift \ac{APE}$_z$ with and without the rotation refinement via post optimization. As mentioned in the previous sub-experiment, this level of accurate local gravity is directly translated into more precise observability of roll and pitch. As shown in \figref{fig:gravity_residual}, the addition of the $\mathbf{r_{post}}$ factor notably improves odometry accuracy in the vertical direction. Specifically, \ac{APE}$_z$ is consistently reduced on every sequence; for example, on \texttt{Quad}, the vertical drift is reduced about 5 \meter{} through the addition of post optimization process. Example graphs of \texttt{BiCorridor} and \texttt{Tunnel} sequences are presented in \figref{fig:gravity_ablation}. Notably, even without loop-closure or point cloud-based registration schemes, GaRLILEO achieves \ac{APE}$_z$ lower than 1 \meter{} on most sequences. 

Taken together, these two sub-experiments demonstrate that GaRLILEO maintains a small, single-digit level of accuracy in local gravity estimation, even under aggressive legged robot locomotion, and that this precision is sufficient to deliver robust, low-drift vertical-state estimation on stairs, slopes, and even on slippery, potentially deformable surfaces.

\subsection{Effect of Velocity Bias}
\label{sec:result.bias}

In this subsection, we evaluate the odometry performance of GaRLILEO with and without the velocity bias. The results are summarized in \tabref{tab:bias_result}, where the error metric is \ac{APE}$_t$ projected onto the $xy$ plane. This 2D metric is used because the velocity bias operates only in the horizontal directions, where foot slip occurs parallel to the ground plane, resulting in horizontal velocity discrepancies.

As shown in \tabref{tab:bias_result}, more than half of the sequences exhibit improved accuracy with the bias, while the others perform better without it. Specifically, in the \texttt{Downstair}, \texttt{Upstair}, \texttt{MoCap-E}, and \texttt{MoCap-H} sequences, excluding the velocity bias yields more accurate $xy$ plane odometry. A common feature of these datasets is the presence of long indoor stairs or challenging environments that frequently degrade leg odometry.

\begin{table}[!b]
\small\sf\centering
\caption{Effect of Velocity Bias on \ac{APE}$_{t}$ projected onto the $xy$ plane}
\label{tab:bias_result}
\resizebox{\columnwidth}{!}{
\begin{tabular}{c|c|c|c|c}
\toprule
\hline
\meter\;(meter) & Atrium & BridgeLoop & CorriLoop & BiCorridor \\ \hline
\textbf{ours}    & \cellcolor[HTML]{def3e6}\textbf{0.809} & \cellcolor[HTML]{def3e6}\textbf{1.187} & \cellcolor[HTML]{def3e6}\textbf{1.619} & \cellcolor[HTML]{def3e6}\textbf{1.207} \\
\textbf{w/o bias} & 1.375 & 1.297 & 2.329 & 1.831 \\
\hline \hline
\meter\;(meter) & Downstair & Upstair & SlopeStair & Overpass \\ \hline
\textbf{ours}    & 3.926 & 1.405 & \cellcolor[HTML]{def3e6}\textbf{2.264} & \cellcolor[HTML]{def3e6}\textbf{1.084} \\
\textbf{w/o bias} & \cellcolor[HTML]{def3e6}\textbf{3.752} & \cellcolor[HTML]{def3e6}\textbf{0.938} & 4.395 & 1.936 \\
\hline \hline
\meter\;(meter) & Tunnel & Quad & MoCap-E & MoCap-H \\ \hline
\textbf{ours}    & \cellcolor[HTML]{def3e6}\textbf{3.481} & \cellcolor[HTML]{def3e6}\textbf{7.028} & 0.846 & 0.885 \\
\textbf{w/o bias} & 4.652 & 7.114 & \cellcolor[HTML]{def3e6}\textbf{0.794} & \cellcolor[HTML]{def3e6}\textbf{0.805} \\
\hline
\bottomrule
\end{tabular}
}
\end{table}


\begin{figure*}[!t]
    \centering
    \includegraphics[width=\textwidth]{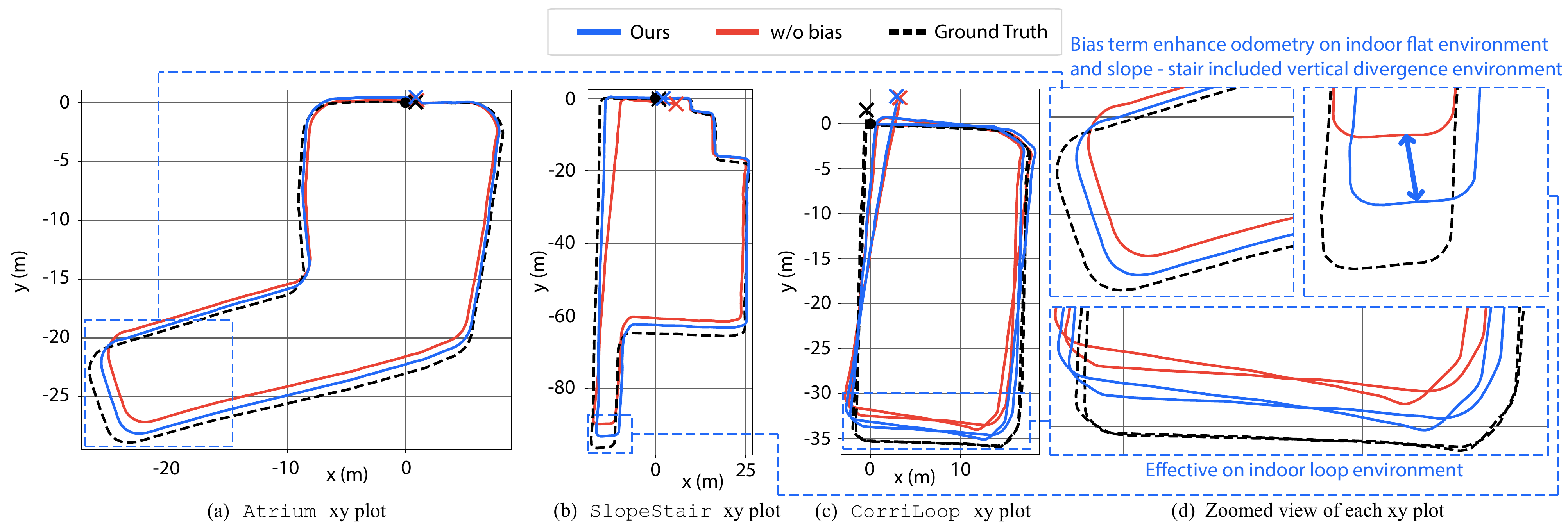}
    \caption{$xy$ plane plot of odometry estimation with and without velocity bias. On \texttt{Atrium}, \texttt{SlopeStair}, and \texttt{CorriLoop} sequences, which are presented in (a), (b), and (c), velocity bias makes odometry more robust as radial velocity information of each radar target point mitigates the error of leg velocity occurring from inaccurate contact and joint encoder measurement. A more detailed zoomed view of each subfigure, showing the effect of the velocity bias term, is included in (d).}
    \label{fig:bias_fig_xy}
\end{figure*}

By contrast, in most indoor sequences, adding the bias improves accuracy, as radar sensors acquire denser, more reliable measurements from nearby static objects. Qualitative odometry is shown in \figref{fig:bias_fig_xy}, which demonstrates enhanced odometry accuracy for both indoor and outdoor sequences. 

On indoor flat sequences, such as \texttt{Atrium} and \texttt{CorriLoop}, the velocity bias term improves odometry accuracy by accounting for potential contact slip using radar target radial velocity information. Due to the relatively rich radar targets in indoor environments, velocity bias may provide more reliable velocity information to the system's velocity factors. Similarly, on \texttt{Tunnel} and \texttt{Overpass} sequences, where the surrounding library building or overpass (including the ceiling) enables the radar to collect sufficient reflections, the result is that odometry got enhanced even if a part of the trajectory traverses open outdoor space. 

In \texttt{BridgeLoop} and \texttt{BiCorridor} sequences, where short stairs are included in the trajectory, present partially enhanced odometry estimation, but their enhancement proportion is a slight fall back compared with simple flat indoor sequences. This is due to the direction of the radar sensor when the robot is traversing staircases. As the radar sensor points upward, it may detect fewer radar targets, since fewer objects may potentially exist in that direction. This phenomenon occurs in \texttt{Downstair} and \texttt{Upstair} sequences, where the longest stair is included in both upstairs and downstairs directions, which even degrades the final odometry estimation when a velocity bias is added. Still, as shown in \tabref{tab:bias_result}, the performance degradation in those cases is less drastic than the performance enhancement on other sequences. 

One interesting point about this experiment is that even the \texttt{SlopeStair} sequence includes a long upstairs region, similar to the downstairs area of \texttt{Downstair}; however, the final odometry estimate was enhanced by almost 50\%. This indirectly demonstrates the effect of the velocity bias factor on mitigating mild leg odometry failures, such as temporal contact slip or contact impact, using radar target velocity information. Even though the possibility of odometry degradation in the stair region exists, the velocity bias factor meaningfully handles the contact failure in the slope area, where almost half of the \texttt{SlopeStair} sequence is designed. This can be qualitatively verified from \figref{fig:bias_fig_xy}~(b) and \figref{fig:bias_fig_xy}~(d), as the odometry accuracy degrades after the downstair region, particularly without a bias term, due to temporal contact slip or impact. The \texttt{Quad} sequence also includes the upstair and down-slope regions during the traverse, but the velocity bias effect is shown comparably mild, as the slope in this sequence is much milder compared with the \texttt{SlopeStair} sequence. 

In the two \texttt{MoCap} sequences, leg odometry deviates from ground truth in the $xy$ plane due to the slippery zone. Because the bias term is designed to change smoothly, it cannot compensate for abrupt or inconsistent slips, resulting in slight performance degradation.

Overall, the velocity bias can enhance odometry by reducing the discrepancy between radar and leg kinematics when leg kinematics temporally fail due to unanticipated contact issues. Still, a potential limitation of the velocity bias on drastically varying discrepancies remains. This effect is particularly evident in places where the static floor assumption fails, such as \texttt{MoCap} sequences. To address this, future work may consider alternative radar configurations (e.g., including ground reflections within the field of view) or an adaptive mechanism that enables or disables the bias term temporally depending on the environment in which the system operates.

\subsection{Real-time Capability on Edge Device}
\label{sec:result.computation}

\begin{figure}[!t]
    \centering
    \includegraphics[width=\columnwidth]{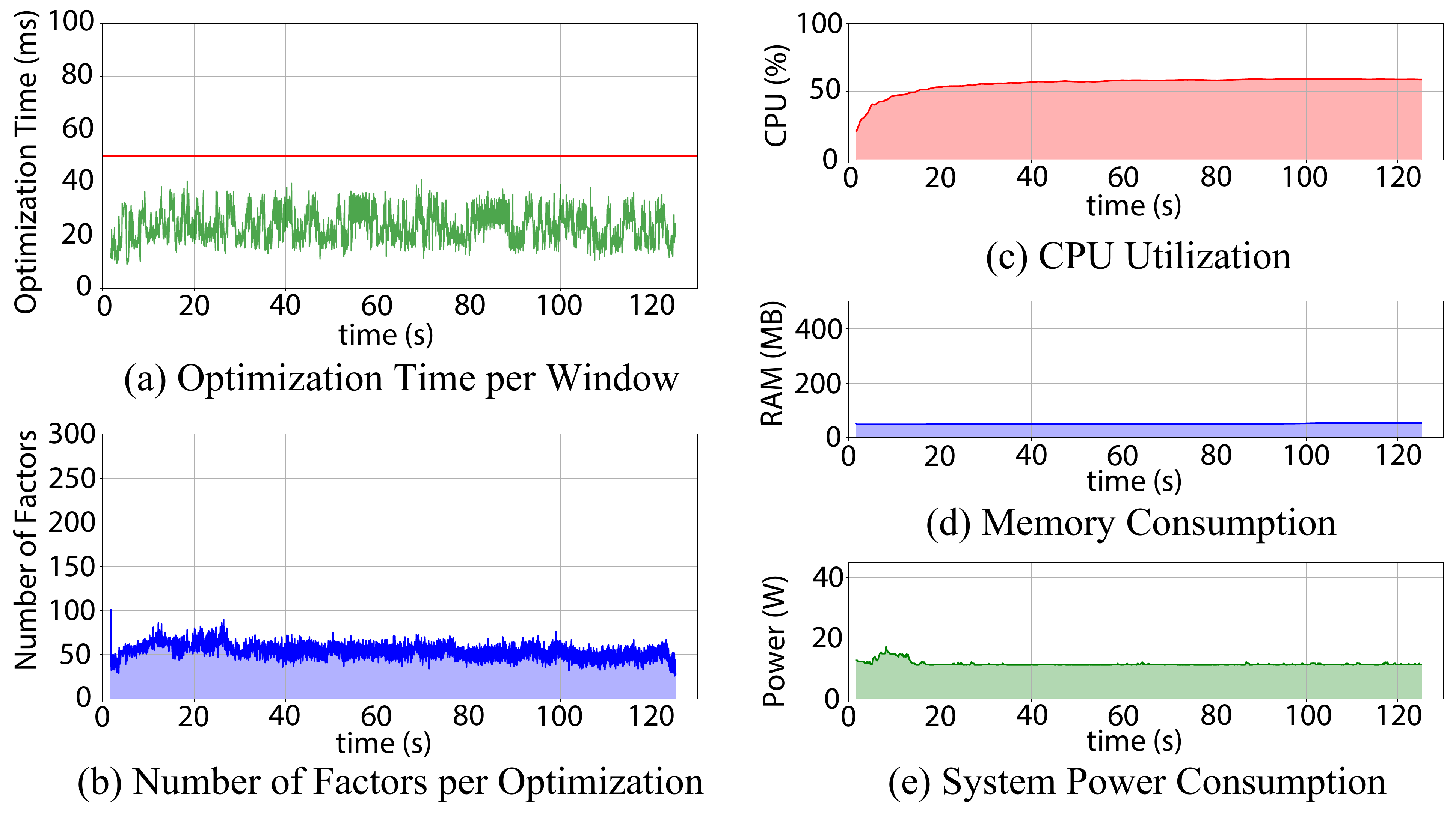}
    \caption{End-to-end edge device test of GaRLILEO on NVIDIA Jetson AGX Orin with \texttt{Atrium} sequence. (a) Optimization time per window during factor graph optimization. Red line denotes \unit{50}{\ms}, which stands for speed of SoC radar sensor input. (b) Number of factors in the factor graph per single optimization. (c) CPU Utilization of GaRLILEO. (d) Memory Consumption of GaRLILEO. (e) Power consumption of NVIDIA Jetson AGX Orin while running GaRLILEO. }
    \label{fig:metric_exp}
\end{figure}

To demonstrate the real-time capability and computational efficiency of our framework, we conducted an experiment by deploying GaRLILEO on an onboard NVIDIA Jetson AGX Orin for the \texttt{Atrium} Sequence. A key factor in achieving this lightweight performance is the sliding window marginalization scheme, detailed in Section~\secref{subsec:marginalization}. This ensures the factor graph size does not grow unbounded over time. As shown in \figref{fig:metric_exp}~(b), an average of 54 residual factors processed per single optimization step, while graph size is internally bounded to a maximum of 21 active control points. This consistency is maintained throughout the entire estimation process.

This bounded graph size directly translates to efficient computation frequencies. During the experiment, the end-to-end computation time per sliding window optimization averaged \unit{23.277}{\ms}, with a maximum recorded value of \unit{41.097}{\ms}. Because the incoming radar scans operate at \unit{20}{\Hz}, the system is required to complete its optimization within a \unit{50}{\ms} window. As illustrated by the red threshold line in \figref{fig:metric_exp}~(a), GaRLILEO's computation time consistently stays below this, guaranteeing real-time deployment capabilities on the edge device without dropping sensor frames. 

Furthermore, the overall system throughput and resource consumption demonstrates the stability of GaRLILEO during operation. On the NVIDIA Jetson AGX Orin, GaRLILEO's CPU utilization averaged \unit{55.856}{\%}, while memory consumption remained at an average of \unit{51.013}{\MB}. The average system power usage during execution was \unit{11.521}{W}, while maximum at \unit{17.198}{\W}. As depicted in \figref{fig:metric_exp}~(c), \figref{fig:metric_exp}~(d), and \figref{fig:metric_exp}~(e), the CPU, memory, and power metrics do not exhibit drastic growth over the algorithm runtime. Instead, they maintain a stable and predictable level, confirming GaRLILEO is suitable for deployment on power-constrained systems. 

\section{Conclusion}
In this work, we presented GaRLILEO, a continuous-time radar-leg-\ac{IMU} odometry framework that leverages B-spline modeling, local gravity estimation, and horizontal velocity bias correction to achieve robust and accurate state estimation in challenging environments. By tightly coupling radar-derived ego velocity with leg kinematics and inertial measurements, our method addresses the limitations of existing odometry pipelines that suffer from vertical drift, contact slip, and sensor modality degradation. 

Extensive experiments across diverse sequences, including long indoor loops, multi-story staircases, outdoor slopes, and environments with severe contact disturbances, demonstrated that GaRLILEO consistently outperforms other \ac{SOTA} baselines in odometry accuracy. Notably, its continuous local gravity estimation significantly reduces vertical drift. At the same time, the B-spline-based fusion scheme ensures robustness against modality-specific failures such as radar sparsity or leg kinematics slips. 

To evaluate the complementary effect of radar and leg kinematics sensors, we tested the performance of the radar-only and leg-only versions of GaRLILEO at \texttt{MoCap} sequences and compared their odometry accuracy. Through this experiment, we verified that both sensor modalities have their own strength points in horizontal and vertical direction accuracy. Consequently, the B-spline-based sensor fusion of GaRLILEO effectively blends the different modalities to enhance the odometry estimation result. 

Through two sub-experiments, we validated the effect of proposing local gravity factors on both the accuracy of local gravity estimation itself and odometry in the vertical direction. These experiments demonstrate how GaRLILEO can achieve sub-meter vertical odometry accuracy without relying on loop closure or point cloud registration algorithms, highlighting that reliable local gravity estimation is crucial for robust, vertically low-drift odometry estimation in stairs, slopes, and challenging terrains. 

We further analyzed the effect of the velocity bias term and verified that it provides clear benefits in structured indoor environments, while in sparse or highly dynamic outdoor scenes, it may lead to potential performance degradation. This suggests that environment-aware adaptation or improved radar hardware configurations could further enhance robustness. 

Overall, our results highlight the potential of continuous-time multimodal fusion to enable legged robots to navigate reliably in complex real-world environments. 

\subsection{Limitations and Future Extensions}
Though GaRLILEO demonstrates robust odometry estimation across diverse indoor and outdoor environments, several limitations remain. First, our framework prefers denser radar points for stable velocity spline estimation. In open outdoor areas such as the \texttt{Quad} sequence, sparse radar returns degrade velocity bias adaptation and limit accuracy. Furthermore, due to the limited observation of yaw orientation, which relies on the \ac{IMU} measurements, an inevitable odometry drift occurs in the horizontal direction as the trajectory becomes longer. 

To mitigate the above limitations, future work will explore adaptive gravity post-optimization, which dynamically enables or disables the term depending on the reliability of the optimized gravity vector. For classification, fusion with exteroceptive sensors, such as cameras and \ac{LiDAR}, or friction estimation based on recent contact tactile sensors can be exploited. Furthermore, to expand the field of view and improve odometry robustness, a multi-\ac{SoC} radar system can be leveraged, which may provide additional yaw orientation observations, as mentioned in \citet{yoon2023need}, or enable point cloud registration. 

Additionally, the current framework relies on static covariances during optimization. To address the unpredictability of legged-robot environments, future work will explore adaptive covariance estimation that dynamically adjusts factor weights based on real-world environmental context and sensor degradation, thereby further enhancing robustness.

\begin{acks}
This work was supported by the Technology Innovation Program (1415187329, 20024355, Development of autonomous driving connectivity technology based on sensor-infrastructure cooperation) funded by the Ministry of Trade, Industry \& Energy (MOTIE, Korea), and in part by the Robotics and AI (RAI) Institute.
\end{acks}

\bibliographystyle{SageH}
\bibliography{string-long,reference}

@article{harlow2024new,
  title={A new wave in robotics: Survey on recent mmwave radar applications in robotics},
  author={Harlow, Kyle and Jang, Hyesu and Barfoot, Timothy D and Kim, Ayoung and Heckman, Christoffer},
  journal=IEEE_J_RO,
  year={2024}
}

@inproceedings{doer2020ekf,
  title={An ekf based approach to radar inertial odometry},
  author={Doer, Christopher and Trommer, Gert F},
  booktitle=C-MFI,
  pages={152--159},
  year={2020}
}

@inproceedings{kellner2013instantaneous,
  title={Instantaneous ego-motion estimation using doppler radar},
  author={Kellner, Dominik and Barjenbruch, Michael and Klappstein, Jens and Dickmann, J{\"u}rgen and Dietmayer, Klaus},
  booktitle=C-ITSC,
  pages={869--874},
  year={2013}
}

@article{park20213d,
  title={3d ego-motion estimation using low-cost mmwave radars via radar velocity factor for pose-graph slam},
  author={Park, Yeong Sang and Shin, Young-Sik and Kim, Joowan and Kim, Ayoung},
  journal=IEEE_J_RAL,
  volume={6},
  number={4},
  pages={7691--7698},
  year={2021}
}

@inproceedings{michalczyk2022tightly,
  title={Tightly-coupled ekf-based radar-inertial odometry},
  author={Michalczyk, Jan and Jung, Roland and Weiss, Stephan},
  booktitle=C-IROS,
  pages={12336--12343},
  year={2022}
}

@article{zhuang20234d,
  title={4d iriom: 4d imaging radar inertial odometry and mapping},
  author={Zhuang, Yuan and Wang, Binliang and Huai, Jianzhu and Li, Miao},
  journal=IEEE_J_RAL,
  volume={8},
  number={6},
  pages={3246--3253},
  year={2023}
}

@article{chen2023drio,
  title={Drio: Robust radar-inertial odometry in dynamic environments},
  author={Chen, Hongyu and Liu, Yimin and Cheng, Yuwei},
  journal=IEEE_J_RAL,
  volume={8},
  number={9},
  pages={5918--5925},
  year={2023}
}

@inproceedings{huang2024less,
  title={Less is more: Physical-enhanced radar-inertial odometry},
  author={Huang, Qiucan and Liang, Yuchen and Qiao, Zhijian and Shen, Shaojie and Yin, Huan},
  booktitle=C-ICRA,
  pages={15966--15972},
  year={2024}
}

@inproceedings{do2024dero,
  title={Dero: Dead reckoning based on radar odometry with accelerometers aided for robot localization},
  author={Do, Hoang Viet and Kim, Yong Hun and Lee, Joo Han and Lee, Min Ho and Song, Jin Woo},
  booktitle=C-IROS,
  pages={8547--8554},
  year={2024}
}

@inproceedings{jung2024co,
  title={Co-RaL: Complementary Radar-Leg Odometry with 4-DoF Optimization and Rolling Contact},
  author={Jung, Sangwoo and Yang, Wooseong and Kim, Ayoung},
  booktitle=C-IROS,
  pages={13289--13296},
  year={2024}
}

@article{geiger2013vision,
  title={Vision meets robotics: The kitti dataset},
  author={Geiger, Andreas and Lenz, Philip and Stiller, Christoph and Urtasun, Raquel},
  journal=J-IJRR,
  volume={32},
  number={11},
  pages={1231--1237},
  year={2013}
}

@inproceedings{roston1992dead,
  title={Dead Reckoning Navigation For Walking Robots},
  author={Roston, GP and Krotkov, EP},
  booktitle=C-IROS,
  volume={1},
  pages={607--612},
  year={1992}
}

@inproceedings{bloesch2012state,
  title={State estimation for legged robots-consistent fusion of leg kinematics and IMU},
  author={Bloesch, Michael and Hutter, Marco and Hoepflinger, Mark A and Leutenegger, Stefan and Gehring, Christian and Remy, C David and Siegwart, Roland},
  booktitle=C-RSS,
  year={2012},
  pages={17--24}
}

@inproceedings{bloesch2013state,
  title={State estimation for legged robots on unstable and slippery terrain},
  author={Bloesch, Michael and Gehring, Christian and Fankhauser, P{\'e}ter and Hutter, Marco and Hoepflinger, Mark A and Siegwart, Roland},
  booktitle=C-IROS,
  pages={6058--6064},
  year={2013}
}

@inproceedings{fallon2014drift,
  title={Drift-free humanoid state estimation fusing kinematic, inertial and lidar sensing},
  author={Fallon, Maurice F and Antone, Matthew and Roy, Nicholas and Teller, Seth},
  booktitle=C-ICHR,
  pages={112--119},
  year={2014}
}

@inproceedings{nobili2017heterogeneous,
  title={Heterogeneous sensor fusion for accurate state estimation of dynamic legged robots},
  author={Nobili, Simona and Camurri, Marco and Barasuol, Victor and Focchi, Michele and Caldwell, Darwin and Semini, Claudio and Fallon, Maurice},
  booktitle=C-RSS,
  year={2017}
}

@article{bloesch2017two,
  title={The two-state implicit filter recursive estimation for mobile robots},
  author={Bloesch, Michael and Burri, Michael and Sommer, Hannes and Siegwart, Roland and Hutter, Marco},
  journal=IEEE_J_RAL,
  volume={3},
  number={1},
  pages={573--580},
  year={2017}
}

@inproceedings{hartley2018legged,
  title={Legged robot state-estimation through combined forward kinematic and preintegrated contact factors},
  author={Hartley, Ross and Mangelson, Josh and Gan, Lu and Jadidi, Maani Ghaffari and Walls, Jeffrey M and Eustice, Ryan M and Grizzle, Jessy W},
  booktitle=C-ICRA,
  pages={4422--4429},
  year={2018}
}

@inproceedings{hartley2018hybrid,
  title={Hybrid contact preintegration for visual-inertial-contact state estimation using factor graphs},
  author={Hartley, Ross and Jadidi, Maani Ghaffari and Gan, Lu and Huang, Jiunn-Kai and Grizzle, Jessy W and Eustice, Ryan M},
  booktitle=C-IROS,
  pages={3783--3790},
  year={2018}
}

@article{hartley2020contact,
  title={Contact-aided invariant extended Kalman filtering for robot state estimation},
  author={Hartley, Ross and Ghaffari, Maani and Eustice, Ryan M and Grizzle, Jessy W},
  journal=J-IJRR,
  volume={39},
  number={4},
  pages={402--430},
  year={2020}
}

@article{camurri2020pronto,
  title={Pronto: A multi-sensor state estimator for legged robots in real-world scenarios},
  author={Camurri, Marco and Ramezani, Milad and Nobili, Simona and Fallon, Maurice},
  journal={Frontiers in Robotics and AI},
  volume={7},
  pages={68},
  year={2020},
  publisher={Frontiers Media SA}
}

@inproceedings{fink2020proprioceptive,
  title={Proprioceptive sensor fusion for quadruped robot state estimation},
  author={Fink, Geoff and Semini, Claudio},
  booktitle=C-IROS,
  pages={10914--10920},
  year={2020}
}

@article{kim2021legged,
  title={Legged robot state estimation with dynamic contact event information},
  author={Kim, Joon-Ha and Hong, Seungwoo and Ji, Gwanghyeon and Jeon, Seunghun and Hwangbo, Jemin and Oh, Jun-Ho and Park, Hae-Won},
  journal=IEEE_J_RAL,
  volume={6},
  number={4},
  pages={6733--6740},
  year={2021},
  publisher={IEEE}
}

@article{kim2022step,
  title={STEP: State estimator for legged robots using a preintegrated foot velocity factor},
  author={Kim, Yeeun and Yu, Byeongho and Lee, Eungchang Mason and Kim, Joon-ha and Park, Hae-won and Myung, Hyun},
  journal=IEEE_J_RAL,
  volume={7},
  number={2},
  pages={4456--4463},
  year={2022}
}

@inproceedings{yang2023multi,
  title={Multi-IMU Proprioceptive Odometry for Legged Robots},
  author={Yang, Shuo and Zhang, Zixin and Bokser, Benjamin and Manchester, Zachary},
  booktitle=C-IROS,
  pages={774--779},
  year={2023}
}

@article{ou2024leg,
  title={Leg-KILO: Robust kinematic-inertial-lidar odometry for dynamic legged robots},
  author={Ou, Guangjun and Li, Dong and Li, Hanmin},
  journal=IEEE_J_RAL,
  year={2024}
}

@inproceedings{dhedin2023visual,
  title={Visual-Inertial and Leg Odometry Fusion for Dynamic Locomotion},
  author={Dh{\'e}din, Victor and Li, Haolong and Khorshidi, Shahram and Mack, Lukas and Ravi, Adithya Kumar Chinnakkonda and Meduri, Avadesh and Shah, Paarth and Grimminger, Felix and Righetti, Ludovic and Khadiv, Majid and others},
  booktitle=C-ICRA,
  pages={9966--9972},
  year={2023}
}

@article{wisth2022vilens,
  title={VILENS: Visual, inertial, lidar, and leg odometry for all-terrain legged robots},
  author={Wisth, David and Camurri, Marco and Fallon, Maurice},
  journal=IEEE_J_RO,
  volume={39},
  number={1},
  pages={309--326},
  year={2022}
}

@inproceedings{yang2023cerberus,
  title={Cerberus: Low-drift visual-inertial-leg odometry for agile locomotion},
  author={Yang, Shuo and Zhang, Zixin and Fu, Zhengyu and Manchester, Zachary},
  booktitle=C-ICRA,
  pages={4193--4199},
  year={2023}
}

@article{lin2023proprioceptive,
  title={Proprioceptive invariant robot state estimation},
  author={Lin, Tzu-Yuan and Li, Tingjun and Tong, Wenzhe and Ghaffari, Maani},
  journal={arXiv preprint arXiv:2311.04320},
  year={2023}
}

@article{nistico2025muse,
  title={MUSE: A real-time multi-sensor state estimator for quadruped robots},
  author={Nistic{\`o}, Ylenia and Soares, Jo{\~a}o Carlos Virgolino and Amatucci, Lorenzo and Fink, Geoff and Semini, Claudio},
  journal=IEEE_J_RAL,
  year={2025}
}

@article{nubert2025holistic,
  title={Holistic fusion: Task-and setup-agnostic robot localization and state estimation with factor graphs},
  author={Nubert, Julian and Tuna, Turcan and Frey, Jonas and Cadena, Cesar and Kuchenbecker, Katherine J and Khattak, Shehryar and Hutter, Marco},
  journal={arXiv preprint arXiv:2504.06479},
  year={2025}
}

@article{qin2018vins,
  title={Vins-mono: A robust and versatile monocular visual-inertial state estimator},
  author={Qin, Tong and Li, Peiliang and Shen, Shaojie},
  journal=IEEE_J_RO,
  volume={34},
  number={4},
  pages={1004--1020},
  year={2018}
}

@inproceedings{kubelka2022gravity,
  title={Gravity-constrained point cloud registration},
  author={Kubelka, Vladim{\'\i}r and Vaidis, Maxime and Pomerleau, Fran{\c{c}}ois},
  booktitle=C-IROS,
  pages={4873--4879},
  year={2022}
}

@article{xu2022fast,
  title={Fast-lio2: Fast direct lidar-inertial odometry},
  author={Xu, Wei and Cai, Yixi and He, Dongjiao and Lin, Jiarong and Zhang, Fu},
  journal=IEEE_J_RO,
  volume={38},
  number={4},
  pages={2053--2073},
  year={2022}
}

@inproceedings{kim2025hercules,
  author={Kim, Hanjun and Jung, Minwoo and Noh, Chiyun and Jung, Sangwoo and Song, Hyunho and Yang, Wooseong and Jang, Hyesu and Kim, Ayoung},
  booktitle=C-ICRA, 
  title={HeRCULES: Heterogeneous Radar Dataset in Complex Urban Environment for Multi-Session Radar SLAM}, 
  year={2025},
  pages={4649-4656}
}

@misc{grupp2017evo,
  title={evo: Python package for the evaluation of odometry and SLAM.},
  author={Grupp, Michael},
  howpublished={\url{https://github.com/MichaelGrupp/evo}},
  year={2017}
}

@ARTICLE{hu2024paloc,
  author={Hu, Xiangcheng and Zheng, Linwei and Wu, Jin and Geng, Ruoyu and Yu, Yang and Wei, Hexiang and Tang, Xiaoyu and Wang, Lujia and Jiao, Jianhao and Liu, Ming},
  journal=IEEE_J_MECH, 
  title={PALoc: Advancing SLAM Benchmarking With Prior-Assisted 6-DoF Trajectory Generation and Uncertainty Estimation}, 
  year={2024},
  volume={29},
  number={6},
  pages={4297-4308},
  doi={10.1109/TMECH.2024.3362902}
}

@inproceedings{zhu2022robust,
  title={Robust real-time lidar-inertial initialization},
  author={Zhu, Fangcheng and Ren, Yunfan and Zhang, Fu},
  booktitle=C-IROS,
  pages={3948--3955},
  year={2022}
}

@inproceedings{doer2021yaw,
  title={Yaw aided radar inertial odometry using manhattan world assumptions},
  author={Doer, Christopher and Trommer, Gert F},
  booktitle=C-ICINS,
  pages={1--9},
  year={2021}
}

@inproceedings{barnes2020oxford,
  title={The oxford radar robotcar dataset: A radar extension to the oxford robotcar dataset},
  author={Barnes, Dan and Gadd, Matthew and Murcutt, Paul and Newman, Paul and Posner, Ingmar},
  booktitle=C-ICRA,
  pages={6433--6438},
  year={2020}
}

@ARTICLE{D-LIOM,
  author={Wang, Zhong and Zhang, Lin and Shen, Ying and Zhou, Yicong},
  journal=IEEE_J_MM, 
  title={{D-LIOM}: Tightly-Coupled Direct LiDAR-Inertial Odometry and Mapping}, 
  year={2023},
  volume={25},
  number={},
  pages={3905-3920},
  doi={10.1109/TMM.2022.3168423}}

@inproceedings{nemiroff2023joint,
  title={Joint On-Manifold Gravity and Accelerometer Intrinsics Estimation for Inertially Aligned Mapping},
  author={Nemiroff, Ryan and Chen, Kenny and Lopez, Brett T},
  booktitle=C-IROS,
  pages={1388--1394},
  year={2023}
}

@article{wildcat,
  title={Wildcat: Online continuous-time 3d lidar-inertial slam},
  author={Ramezani, Milad and Khosoussi, Kasra and Catt, Gavin and Moghadam, Peyman and Williams, Jason and Borges, Paulo and Pauling, Fred and Kottege, Navinda},
  journal={arXiv preprint arXiv:2205.12595},
  year={2022}
}

@article{agha2021nebula,
  title={Nebula: Quest for robotic autonomy in challenging environments; team costar at the darpa subterranean challenge},
  author={Agha, Ali and Otsu, Kyohei and Morrell, Benjamin and Fan, David D and Thakker, Rohan and Santamaria-Navarro, Angel and Kim, Sung-Kyun and Bouman, Amanda and Lei, Xianmei and Edlund, Jeffrey and others},
  journal={arXiv preprint arXiv:2103.11470},
  year={2021}
}

@inproceedings{noh2025garlio,
  author={Noh, Chiyun and Yang, Wooseong and Jung, Minwoo and Jung, Sangwoo and Kim, Ayoung},
  booktitle=C-ICRA, 
  title={GaRLIO: Gravity Enhanced Radar-LiDAR-Inertial Odometry}, 
  year={2025},
  pages={9869-9875}
}

@article{burnett2025tro,
  author={Burnett, Keenan and Schoellig, Angela P. and Barfoot, Timothy D.},
  journal=IEEE_J_RO, 
  title={Continuous-Time Radar-Inertial and Lidar-Inertial Odometry Using a Gaussian Process Motion Prior}, 
  year={2025},
  volume={41},
  number={},
  pages={1059-1076},
}

@inproceedings{furgale2012continuous,
  title={Continuous-time batch estimation using temporal basis functions},
  author={Furgale, Paul and Barfoot, Timothy D and Sibley, Gabe},
  booktitle=C-ICRA,
  pages={2088--2095},
  year={2012}
}

@article{furgale2015continuous,
  title={Continuous-time batch trajectory estimation using temporal basis functions},
  author={Furgale, Paul and Tong, Chi Hay and Barfoot, Timothy D and Sibley, Gabe},
  journal=J-IJRR,
  volume={34},
  number={14},
  pages={1688--1710},
  year={2015}
}

@article{ovren2019trajectory,
  title={Trajectory representation and landmark projection for continuous-time structure from motion},
  author={Ovr{\'e}n, Hannes and Forss{\'e}n, Per-Erik},
  journal=J-IJRR,
  volume={38},
  number={6},
  pages={686--701},
  year={2019}
}

@inproceedings{hug2020hyperslam,
  title={HyperSLAM: A generic and modular approach to sensor fusion and simultaneous localization and mapping in continuous-time},
  author={Hug, David and Chli, Margarita},
  booktitle=C-3DV,
  pages={978--986},
  year={2020}
}

@article{lang2022ctrl,
  title={Ctrl-VIO: Continuous-time visual-inertial odometry for rolling shutter cameras},
  author={Lang, Xiaolei and Lv, Jiajun and Huang, Jianxin and Ma, Yukai and Liu, Yong and Zuo, Xingxing},
  journal=IEEE_J_RAL,
  volume={7},
  number={4},
  pages={11537--11544},
  year={2022}
}

@article{hug2022continuous,
  title={Continuous-time stereo-inertial odometry},
  author={Hug, David and B{\"a}nninger, Philipp and Alzugaray, Ignacio and Chli, Margarita},
  journal=IEEE_J_RAL,
  volume={7},
  number={3},
  pages={6455--6462},
  year={2022}
}

@inproceedings{droeschel2018efficient,
  title={Efficient continuous-time SLAM for 3D lidar-based online mapping},
  author={Droeschel, David and Behnke, Sven},
  booktitle=C-ICRA,
  pages={5000--5007},
  year={2018}
}

@article{lv2023continuous,
  title={Continuous-time fixed-lag smoothing for LiDAR-inertial-camera SLAM},
  author={Lv, Jiajun and Lang, Xiaolei and Xu, Jinhong and Wang, Mengmeng and Liu, Yong and Zuo, Xingxing},
  journal=IEEE_J_MECH,
  volume={28},
  number={4},
  pages={2259--2270},
  year={2023}
}

@article{lang2023coco,
  title={Coco-lic: continuous-time tightly-coupled lidar-inertial-camera odometry using non-uniform b-spline},
  author={Lang, Xiaolei and Chen, Chao and Tang, Kai and Ma, Yukai and Lv, Jiajun and Liu, Yong and Zuo, Xingxing},
  journal=IEEE_J_RAL,
  volume={8},
  number={11},
  pages={7074--7081},
  year={2023}
}

@article{lang2024gaussian,
  title={Gaussian-LIC: Real-time photo-realistic SLAM with Gaussian splatting and LiDAR-inertial-camera fusion},
  author={Lang, Xiaolei and Li, Laijian and Wu, Chenming and Zhao, Chen and Liu, Lina and Liu, Yong and Lv, Jiajun and Zuo, Xingxing},
  journal={arXiv preprint arXiv:2404.06926},
  year={2024}
}

@article{jung2023asynchronous,
  title={Asynchronous multiple lidar-inertial odometry using point-wise inter-lidar uncertainty propagation},
  author={Jung, Minwoo and Jung, Sangwoo and Kim, Ayoung},
  journal=IEEE_J_RAL,
  volume={8},
  number={7},
  pages={4211--4218},
  year={2023},
}

@article{talbot2024continuous,
  title={Continuous-time state estimation methods in robotics: A survey},
  author={Talbot, William and Nubert, Julian and Tuna, Turcan and Cadena, Cesar and D{\"u}mbgen, Frederike and Tordesillas, Jesus and Barfoot, Timothy D and Hutter, Marco},
  journal={arXiv preprint arXiv:2411.03951},
  year={2024}
}

@article{chen2024river,
  title={River: A tightly-coupled radar-inertial velocity estimator based on continuous-time optimization},
  author={Chen, Shuolong and Li, Xingxing and Li, Shengyu and Zhou, Yuxuan and Wang, Shiwen},
  journal=IEEE_J_RAL,
  volume={9},
  number={7},
  pages={6107--6114},
  year={2024}
}

@article{tranzatto2022cerberus,
  title={Cerberus in the darpa subterranean challenge},
  author={Tranzatto, Marco and Miki, Takahiro and Dharmadhikari, Mihir and Bernreiter, Lukas and Kulkarni, Mihir and Mascarich, Frank and Andersson, Olov and Khattak, Shehryar and Hutter, Marco and Siegwart, Roland and others},
  journal={Science Robotics},
  volume={7},
  number={66},
  pages={eabp9742},
  year={2022},
  publisher={American Association for the Advancement of Science}
}

@article{kim2025sherloc,
  title={SHeRLoc: Synchronized Heterogeneous Radar Place Recognition for Cross-Modal Localization},
  author={Kim, Hanjun and Jung, Minwoo and Yang, Wooseong and Kim, Ayoung},
  journal={arXiv preprint arXiv:2506.15175},
  year={2025}
}

@article{sibley2010sliding,
  title={Sliding window filter with application to planetary landing},
  author={Sibley, Gabe and Matthies, Larry and Sukhatme, Gaurav},
  journal={Journal of field robotics},
  volume={27},
  number={5},
  pages={587--608},
  year={2010},
  publisher={Wiley Online Library}
}

@book{patrikalakis2002shape,
  title={Shape interrogation for computer aided design and manufacturing},
  author={Patrikalakis, Nicholas M and Maekawa, Takashi},
  volume={15},
  year={2002},
  publisher={Springer}
}

@inproceedings{sommer2020efficient,
  title={Efficient derivative computation for cumulative b-splines on lie groups},
  author={Sommer, Christiane and Usenko, Vladyslav and Schubert, David and Demmel, Nikolaus and Cremers, Daniel},
  booktitle={Proceedings of the IEEE/CVF conference on computer vision and pattern recognition},
  pages={11148--11156},
  year={2020}
}

@inproceedings{shan2018lego,
  title={Lego-loam: Lightweight and ground-optimized lidar odometry and mapping on variable terrain},
  author={Shan, Tixiao and Englot, Brendan},
  booktitle=C-IROS,
  pages={4758--4765},
  year={2018}
}

@article{wei2021ground,
  title={Ground-SLAM: Ground constrained LiDAR SLAM for structured multi-floor environments},
  author={Wei, Xin and Lv, Jixin and Sun, Jie and Pu, Shiliang},
  journal={arXiv preprint arXiv:2103.03713},
  year={2021}
}

@inproceedings{seo2022pago,
  title={PaGO-LOAM: Robust ground-optimized LiDAR odometry},
  author={Seo, Dong-Uk and Lim, Hyungtae and Lee, Seungjae and Myung, Hyun},
  booktitle=C-UR,
  pages={1--7},
  year={2022},
  organization={IEEE}
}

@article{wang2024robust,
  title={Robust Ground Constrained SLAM for Mobile Robot With Sparse-Channel LiDAR},
  author={Wang, Shaocong and Cao, Fengkui and Wang, Ting and Shao, Shiliang and Liu, Lianqing},
  journal=IEEE_J_IV,
  year={2024}
}

@inproceedings{shu2021visual,
  title={Visual slam with graph-cut optimized multi-plane reconstruction},
  author={Shu, Fangwen and Xie, Yaxu and Rambach, Jason and Pagani, Alain and Stricker, Didier},
  booktitle=C-ISMAR,
  pages={165--170},
  year={2021}
}

@article{li2020structure,
  title={Structure-slam: Low-drift monocular slam in indoor environments},
  author={Li, Yanyan and Brasch, Nikolas and Wang, Yida and Navab, Nassir and Tombari, Federico},
  journal=IEEE_J_RAL,
  volume={5},
  number={4},
  pages={6583--6590},
  year={2020}
}

@inproceedings{arndt2023planar,
  title={Do planar constraints improve camera pose estimation in monocular slam?},
  author={Arndt, Charlotte and Sabzevari, Reza and Civera, Javier},
  booktitle=C-ICCV,
  pages={2221--2230},
  year={2023}
}

@inproceedings{yoon2023need,
  title={Need for speed: Fast correspondence-free lidar-inertial odometry using doppler velocity},
  author={Yoon, David J and Burnett, Keenan and Laconte, Johann and Chen, Yi and Vhavle, Heethesh and Kammel, Soeren and Reuther, James and Barfoot, Timothy D},
  booktitle=C-IROS,
  pages={5304--5310},
  year={2023}
}

@misc{tranzatto2022teamcerberus,
      title={Team CERBERUS Wins the DARPA Subterranean Challenge: Technical Overview and Lessons Learned}, 
      author={Marco Tranzatto and Mihir Dharmadhikari and Lukas Bernreiter and Marco Camurri and Shehryar Khattak and Frank Mascarich and Patrick Pfreundschuh and David Wisth and Samuel Zimmermann and Mihir Kulkarni and Victor Reijgwart and Benoit Casseau and Timon Homberger and Paolo De Petris and Lionel Ott and Wayne Tubby and Gabriel Waibel and Huan Nguyen and Cesar Cadena and Russell Buchanan and Lorenz Wellhausen and Nikhil Khedekar and Olov Andersson and Lintong Zhang and Takahiro Miki and Tung Dang and Matias Mattamala and Markus Montenegro and Konrad Meyer and Xiangyu Wu and Adrien Briod and Mark Mueller and Maurice Fallon and Roland Siegwart and Marco Hutter and Kostas Alexis},
      year={2022},
      eprint={2207.04914},
      archivePrefix={arXiv},
      primaryClass={cs.RO},
      url={https://arxiv.org/abs/2207.04914}, 
}

@STRING{C-MFI = {Proceedings of the {IEEE} International Conference on Multisensor Fusion and Integration for Intelligence System}}

@STRING{C-ICCV = {Proceedings of the {IEEE} International Conference on Computer Vision}}

@STRING{C-ICINS = {Proceedings of the {IEEE} Saint Petersburg International Conference on Integrated Navigation System}}

@STRING{C-ICRA = {Proceedings of the {IEEE} International Conference on Robotics and Automation}}

@STRING{C-ICHR = {Proceedings of the {IEEE} International Conference on Humanoid Robots}}

@STRING{C-IROS = {Proceedings of the {IEEE}/{RSJ} International Conference on Intelligent Robots and Systems}}

@STRING{C-3DV = {Proceedings of the {IEEE} International Conference on 3D Vision}}

@STRING{C-ISMAR = {Proceedings of the {IEEE} International Symposium on Mixed and Augmented Reality}}

@STRING{C-ITSC ={Proceedings of the {IEEE} Intelligent Transportation Systems Conference}}

@STRING{C-RSS = {Proceedings of the Robotics: Science \& Systems Conference}}

@STRING{C-UR = {Proceedings of the International Conference on Ubiquitous Robots}}

@STRING{IEEE_J_IV = {{IEEE} Transactions on Intelligent Vehicles}}

@STRING{IEEE_J_MM = {{IEEE} Transactions on Multimedia}}

@STRING{IEEE_J_RAL = {{IEEE} Robotics and Automation Letters}}

@STRING{IEEE_J_RO = {{IEEE} Transactions on Robotics}}

@STRING{IEEE_J_MECH = {{IEEE/ASME} Transactions on Mechatronics}}

@STRING{J-IJRR = {International Journal of Robotics Research}}

\end{document}